\documentclass{article} 
\usepackage{iclr2027_conference,times}

\usepackage{amsmath,amsfonts,bm}

\def\eqref#1{equation~\ref{#1}}

\def\1{\bm{1}}

\DeclareMathAlphabet{\mathsfit}{\encodingdefault}{\sfdefault}{m}{sl}
\SetMathAlphabet{\mathsfit}{bold}{\encodingdefault}{\sfdefault}{bx}{n}

\usepackage{hyperref}
\hypersetup{hypertexnames=false,hidelinks}
\usepackage{url}

\usepackage{multirow}
\usepackage{makecell}
\usepackage{graphicx}
\usepackage{amsfonts}
\usepackage{amsmath}
\usepackage{amssymb}
\usepackage{algorithm}
\usepackage{algpseudocode}
\usepackage{booktabs}
\usepackage{xcolor}
\usepackage{enumitem}
\usepackage{inconsolata}
\usepackage{listings}
\usepackage{fancyvrb}
\usepackage{fvextra}
\usepackage{CJKutf8}
\usepackage{nicefrac}
\usepackage{siunitx}
\usepackage{xspace}
\usepackage{wrapfig}
\usepackage{longtable}
\usepackage{array}

\setlist[itemize]{leftmargin=1.2em,itemsep=2pt,topsep=3pt,parsep=0pt,partopsep=0pt}

\title{
    RealCADBench: Benchmarking Parametric CAD Modelimng via Industrial Design Intents
}

\author{\parbox[t]{\dimexpr\textwidth-2\tabcolsep\relax}{%
\centering\normalfont\small
\textbf{Guanlin Li, Zhichao Huang, Huimu Yu, Yichen Long, Hongsen Liu\thanks{Corresponding to Hongsen Liu (\href{mailto:liuhongsen3@jd.com}{\texttt{liuhongsen3@jd.com}}), and Yuchen Wang (\href{mailto:wangyuchen.101@jd.com}{\texttt{wangyuchen.101@jd.com}}).}, Linxin Cai, Qiuhe Hong}\\
\textbf{Zongzhen Li, Wei Wang, Xianwen Zhong, Yuchen Wang\textsuperscript{*}}\\[0.5em]
JD Industrial
}}

\iclrfinalcopy
\begin{document}

\maketitle

\begin{abstract}
Parametric computer-aided design (CAD) modeling is difficult to evaluate with a single metric. Existing CAD benchmarks often emphasize synthetic or CAD-native settings, limited input modalities, or executability and IoUs alone. We introduce \textbf{RealCADBench}, a benchmark for intent-to-program CAD modeling from real industrial design intents. It contains 12,632 tasks from 19 factory-automation categories and spans \emph{text descriptions}, \emph{2D engineering drawings}, \emph{real product pictures}, and \emph{rendered images} for both \textit{Part} and \textit{Assembly} modeling. We report results on a 1,770-task evaluation slice: 1,745 Part tasks across four input regimes and RCB-Assm25, a 25-task assembly study used in every reported assembly comparison. Each method generates FreeCAD API Python, which a shared runtime executes to export the 3D model. We evaluate the exported model with executability, Solid IoU, Surface IoU, and a rubric-based visual-semantic identity Judge. Among the nine standalone frontier large models we evaluate, no model leads all four metrics. Across the six frontier-scale large models, executability ranges from $0.565$ to $0.812$, Solid IoU from $0.2841$ to $0.5379$, and Surface IoU from $0.112$ to $0.217$ across the four Part regimes. The highest regime-balanced composite comes from a different model than the leaders on the four component metrics. On RCB-Assm25, Codex with GPT-5.5 improves executability and both IoU metrics over standalone GPT-5.5, but lowers the Judge by $6.98$ pp, leaving GPT-5.5 as the Judge leader. We also observe recurring failure modes, most notably missing fine structures, loss of part identity, and incorrect assembly placement. These results show that execution alone is insufficient to characterize realistic CAD modeling and that frontier models and agents differ substantially across executability, IoUs, and visual-semantic identity.
\end{abstract}

\section{Introduction}
\label{sec:introduction}

For parametric CAD modeling, a runnable program is only a partial indicator of success. A syntactically valid program may still omit mounting holes from a bracket. A plausible render may still place the wrong components in an assembly. Executability, IoUs with a stated target, and visual-semantic identity therefore do not necessarily move together. Evaluating systems by executability alone, or by a single IoU score, misses these failure modes.

The problem becomes harder on real industrial design intents. Inputs may come from text specifications, 2D engineering drawings, real product pictures, or rendered images. Each source reveals a different and often incomplete view of the intended geometry. Interface details and repeated structures can be decisive even when they are only partially visible. Evaluation in this setting must therefore distinguish whether a system can produce a valid artifact, recover the underlying shape, and preserve the visible identity of the target part or assembly.

ABC~\citep{abc}, Fusion 360 Gallery~\citep{fusion360}, DeepCAD~\citep{deepcad}, and SketchGraphs~\citep{sketchgraphs} provide large geometric corpora and construction histories. Text2CAD~\citep{text2cad} and CAD-Recode~\citep{cadrecode} evaluate generation conditioned on language or observations. More recent benchmarks expand the scope further by introducing mixed inputs~\citep{text2cadbench,unicad,cadbench2026}, editable or executable outputs~\citep{parametriccadbench}, and semantic, assembly, or engineering criteria~\citep{benchcad2026,p3dbench2026,muse2026,cadengbench2026}. Most existing benchmarks still emphasize synthetic or CAD-native settings, limited input modalities, or executability and geometric IoUs alone. As a result, they often fail to separate executability, IoUs, and product identity in realistic industrial settings. They also usually evaluate standalone models rather than agents.

\begin{figure*}[htbp]
    \centering
    {\includegraphics[width=0.9\textwidth]{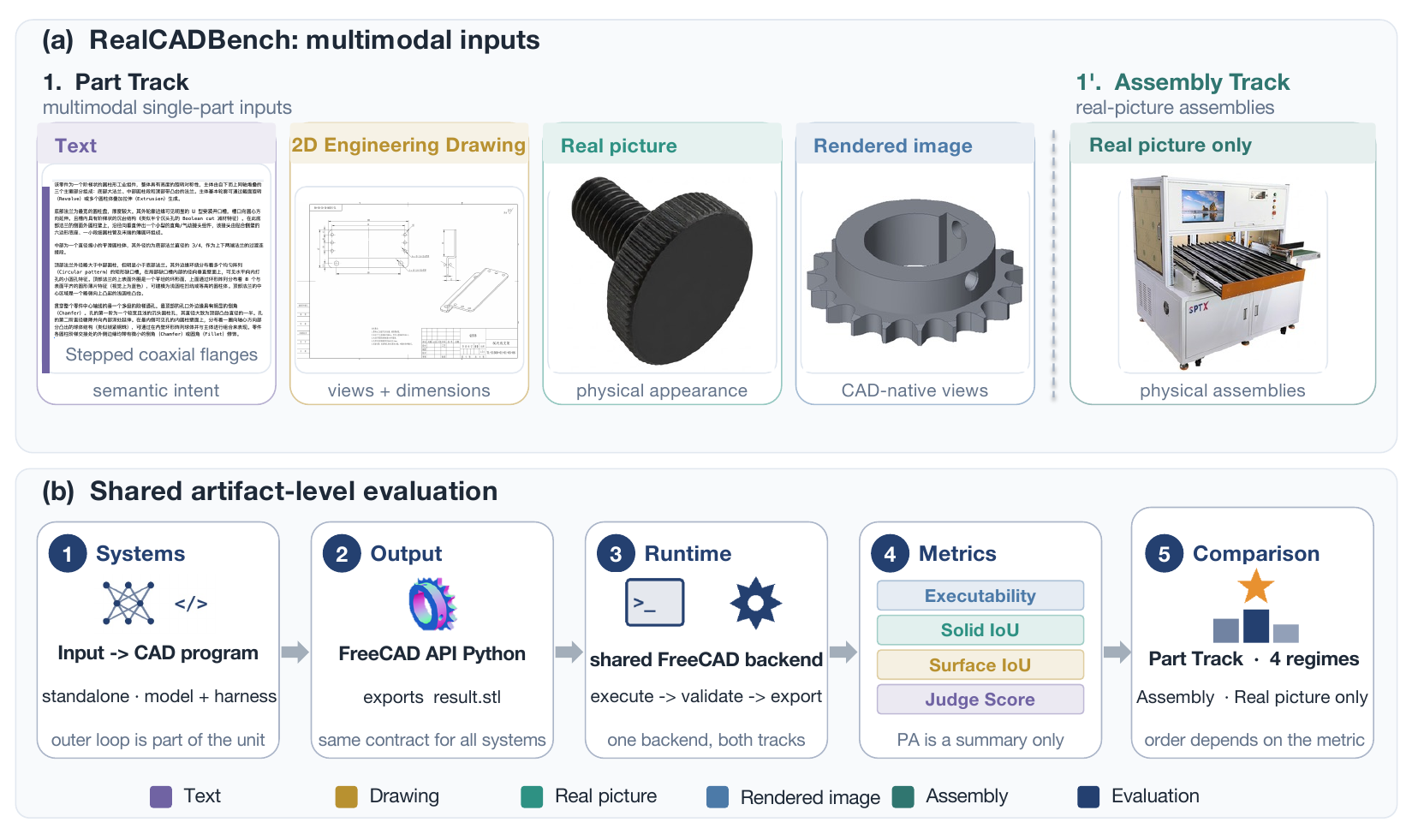}}
    \caption{Overview of RealCADBench. (a)~Four Part input regimes and the real picture Assembly track. (b)~Shared evaluation: methods generate FreeCAD API Python, the common runtime exports the final 3D model, and scoring reports four metrics.}
    \label{fig:overview}
\end{figure*}

We introduce \textbf{RealCADBench} (Figure~\ref{fig:overview}), a benchmark for intent-to-program CAD modeling from real industrial design intents. It contains 12,632 tasks from 19 factory-automation categories. These tasks span text descriptions, 2D engineering drawings, real product pictures, and rendered images, and cover both Part and Assembly modeling. We report results on a 1,770-task evaluation slice consisting of 1,745 Part tasks across four input regimes and RCB-Assm25, a stratified 25-task assembly study used in every reported assembly comparison. Each method generates FreeCAD API Python, which a shared runtime executes to export \texttt{result.stl} as the final 3D model. We evaluate the exported model using executability, Solid IoU, Surface IoU, and a rubric-based visual-semantic identity Judge that compares rendered views of the exported model with the original inputs. 
In this paper, we use ``agent'' to refer to a harness configuration rather than a standalone model.

Our experiments show three main patterns. First, executability, IoUs, and visual-semantic identity separate in practice. Among the nine standalone frontier large models we evaluate, no model leads all four metrics. The model with the highest regime-balanced composite is also different from the leaders on the four component metrics. Second, difficulty does not follow a single order across input regimes. Across the six frontier-scale large models, executability ranges from $0.565$ to $0.812$, Solid IoU from $0.2841$ to $0.5379$, and Surface IoU from $0.112$ to $0.217$ across the four Part regimes. Third, frontier models and agents differ substantially once evaluation goes beyond execution alone. On RCB-Assm25, Codex with GPT-5.5 improves executability and both IoU metrics over standalone GPT-5.5, but lowers the Judge by $6.98$ pp, leaving standalone GPT-5.5 as the Judge leader. We also observe recurring failure modes, most notably missing fine structures, loss of part identity, and incorrect assembly placement.


The paper makes three contributions:
\begin{itemize}
    \item \textbf{A comprehensive benchmark for real industrial design intents}: RealCADBench curates 12,632 tasks across text, 2D engineering drawing, real picture, and rendered image inputs, spanning both Part and Assembly modeling under one shared FreeCAD artifact contract. It covers 19 factory-automation categories and is designed to evaluate frontier multimodal large models and state-of-the-art agents in the intent-to-program setting.

    \item \textbf{A systematic evaluation protocol beyond executability and IoUs}: RealCADBench reports executability, Solid IoU, Surface IoU, and a rubric-based visual-semantic identity Judge for both parts and assemblies. The Judge complements the IoUs by assessing product identity, salient features, and assembly relations. The component scores and their denominators remain the basis for interpretation when Profile Average is reported (Equation~\ref{eq:profile_average}).

    \item \textbf{A comparative study of frontier models and agents on realistic CAD modeling}: RealCADBench evaluates nine standalone frontier large models on the Part Track and two state-of-the-art agents, together with their paired standalone models, on RCB-Assm25. The resulting quantitative and qualitative analyses show that different metrics favor different systems and reveal recurring failure modes in intent-to-program CAD modeling.
\end{itemize}

\section{Related Work}
\label{sec:related_work}

\paragraph{Geometry and construction-history corpora.}
ABC~\citep{abc}, Fusion 360 Gallery~\citep{fusion360}, DeepCAD~\citep{deepcad}, and SketchGraphs~\citep{sketchgraphs} provide large corpora of CAD geometry, design sequences, command histories, and constrained sketches. Text2CAD~\citep{text2cad} extends this line of work to language-conditioned generation, while CAD-Recode~\citep{cadrecode} recovers executable code from point clouds. These resources broaden the range of CAD modeling tasks that can be studied, but they do not evaluate executability, IoUs, and visual-semantic identity as distinct outcomes for real industrial design intents.

\paragraph{CAD benchmarks beyond a single geometry score.}
Recent CAD benchmarks expand evaluation beyond a single geometry score in several ways. Text2CAD-Bench~\citep{text2cadbench}, UniCAD~\citep{unicad}, and CADBench~\cite{cadbench2026} study mixed-input settings. Parametric CAD Bench~\citep{parametriccadbench} and CADGenBench~\citep{cadgenbench2026} focus on executable or editable outputs. BenchCAD~\citep{benchcad2026}, P3D-Bench~\citep{p3dbench2026}, MUSE~\citep{muse2026}, and CADEngBench~\citep{cadengbench2026} incorporate semantic, assembly, or engineering criteria. Despite these advances, most CAD benchmarks still focus on synthetic or CAD-native tasks, whereas RealCADBench evaluates real industrial design intents and reports four distinct measures.

\paragraph{Execution and inference-time repair.}
Zero-to-CAD~\citep{zerotocad} and the reference agent in CADGenBench~\citep{cadgenbench2026} execute and revise programs at inference time. In such settings, runtime feedback, tool use, and compute budget become part of the evaluated method. RealCADBench therefore compares each agent with its paired standalone model on the same assembly tasks. Table~\ref{tab:cad_benchmark_summary} places this evaluation choice in the context of prior CAD benchmarks.

\begin{table*}[htbp]
\centering
\caption{Selected CAD benchmarks along input, output, and system-level evaluation axes.}
\label{tab:cad_benchmark_summary}
\scriptsize
\setlength{\tabcolsep}{3.5pt}
\begin{tabular}{@{}llllcc@{}}
\toprule
\textbf{Work} & \textbf{Input} & \textbf{Track} & \textbf{Evaluated output} & \makecell{\textbf{Beyond}\\\textbf{geometry}} & \textbf{Agent} \\
\midrule
DeepCAD~(\citealp{deepcad}) & CAD sequence & Part & Parametric sequence & $\times$ & $\times$ \\
Text2CAD~(\citealp{text2cad}) & Text & Part & Parametric sequence & $\times$ & $\times$ \\
CAD-Recode~(\citealp{cadrecode}) & Point cloud & Part & Executable code & $\times$ & $\times$ \\
CADBench~(\citealp{cadbench2026}) & Mesh / image / render & Part & Executable program & $\times$ & $\times$ \\
BenchCAD~(\citealp{benchcad2026}) & Text / image / code & Part & CadQuery program & $\checkmark$ & $\times$ \\
P3D-Bench~(\citealp{p3dbench2026}) & Text / image & Part + assembly & Parametric program & $\checkmark$ & $\times$ \\
CADGenBench~(\citealp{cadgenbench2026}) & 2D Engineering Drawing / CAD edit & Part & STEP/BREP & $\checkmark$ & $\checkmark$ \\
MUSE~(\citealp{muse2026}) & Text specification & Assembly & Executable B-Rep & $\checkmark$ & $\times$ \\
CADEngBench~(\citealp{cadengbench2026}) & Specification / CAD & Part + assembly & CAD / behavior & $\checkmark$ & $\times$ \\
\textbf{RealCADBench} & \makecell[l]{Text / 2D Engineering Drawing \\real picture / rendered image} & Part + assembly & FreeCAD Python API & $\checkmark$ & $\checkmark$ \\
\bottomrule
\end{tabular}
\end{table*}

\section{RealCADBench: Tasks and Protocol}
\label{sec:benchmark_construction}

\subsection{Tasks and Tracks}
In RealCADBench, a method takes real industrial design intents as input and produces a parametric CAD program whose export matches a reference 3D model. Each task is defined by an input $x_i$, a reference 3D model $C_i^{\star}$, and metadata for the input regime, category, and views. The method outputs a Python program $p_i$ that calls the FreeCAD API~\citep{freecad_software}. The shared runtime executes $p_i$ and exports \texttt{result.stl} as the final 3D model. Scoring depends only on that exported model, not on the source code or construction path.

\begin{figure*}[htbp]
    \centering
    {\includegraphics[width=1\textwidth]{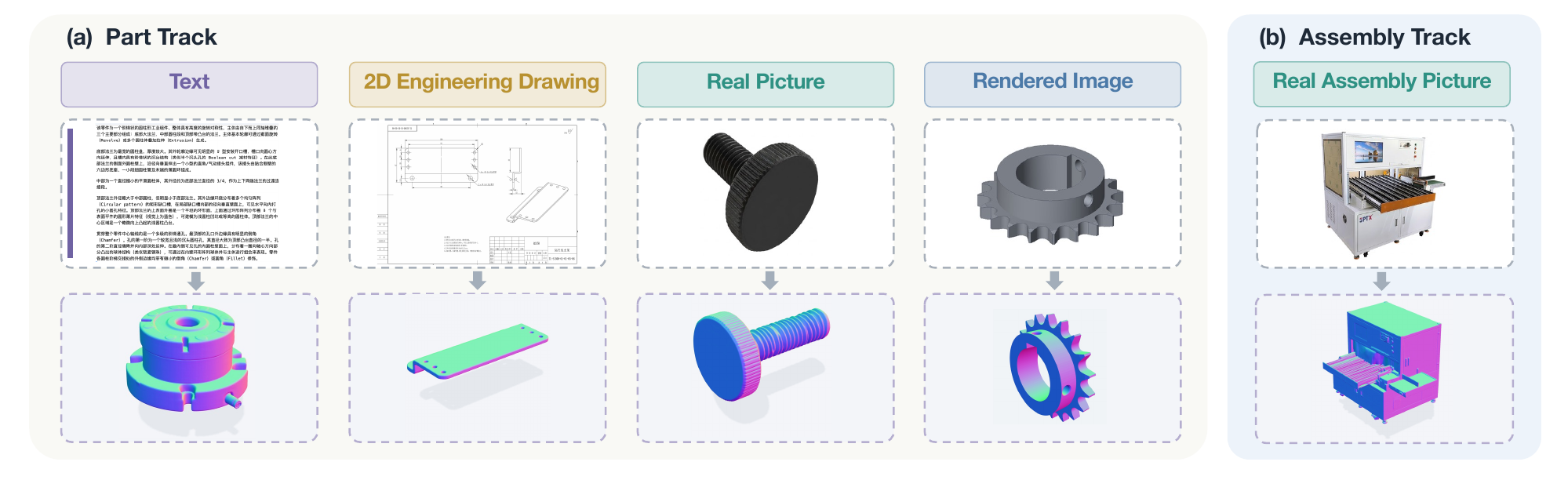}}
    \caption{Representative inputs and reference 3D models.}
    \label{fig:modality_examples}
\end{figure*}

The benchmark contains two tracks: the \emph{Multimodal Part CAD Modeling Track} and the \emph{Image-based Assembly CAD Modeling Track}. The Part Track contains 12,465 tasks, including 568 text descriptions, 236 2D engineering drawings, 11,288 real product pictures, and 373 rendered images. The Assembly Track contains 167 real picture tasks. In this paper, we report a 1,770-task evaluation slice consisting of 1,745 Part tasks across four input regimes, namely 568 text descriptions, 236 2D engineering drawings, 568 real product pictures, and 373 rendered images, together with \textit{RCB-Assm25}, a 25-task assembly study used in all reported assembly comparisons. Table~\ref{tab:benchmark_regimes} summarizes the full benchmark and the reported slice. Figure~\ref{fig:modality_examples} shows representative inputs and their benchmark references.

\begin{table}[htbp]
    \centering
    \caption{Benchmark totals and reported evaluation sets. All regimes use verified benchmark references. More details appear in next paragraphs and Appendix~\ref{sec:appendix_construction}.}
    \label{tab:benchmark_regimes}
    \scriptsize
    \begin{tabular}{@{}llrr@{}}
        \toprule
        \textbf{Track} & \textbf{Input regime} & \textbf{Benchmark tasks} & \textbf{Reported tasks} \\
        \midrule
        Part & Text & 568 & 568 \\
        Part & 2D Engineering Drawing & 236 & 236 \\
        Part & Real picture & 11,288 & 568 \\
        Part & Rendered image & 373 & 373 \\
        Assembly & Real picture & 167 & 25 \\
        \midrule
        \multicolumn{2}{l}{Total tasks} & 12,632 & 1,770 \\
        \bottomrule
    \end{tabular}
\end{table}

\subsection{Data and References}

\paragraph{Reference construction.}
All benchmark references are verified before evaluation. For rendered image tasks, the reference 3D model is exported from the same verified STEP asset used to generate the input render. For real product pictures, 2D engineering drawings, and text, paired vendor CAD is usually unavailable, so the benchmark reference is verified by CAD experts. Text tasks are derived from corresponding real picture cases through Gemini 3.0 Pro generation and validation, and the Judge for the text regime therefore uses the corresponding real picture input. Geometric scores are computed against the verified benchmark reference for each task. Appendix~\ref{sec:appendix_construction} provides the construction details.

\paragraph{Data curation.}
Candidates come from an industrial marketplace and a purchased repository of CAD assets, including STEP files, 2D drawings, and metadata. We remove corrupted, duplicate, ambiguous, or otherwise unusable records. For visual inputs, this screening removes corrupted files, exact or near duplicates, records with ambiguous product identity, and samples that cannot be associated with a consistent product version. For CAD assets, we require successful parsing, non-empty geometry, and successful rendering or 3D model generation when the asset is used to construct benchmark tasks. Before partitioning, we group records that share a SKU, product family, CAD file, or near-duplicate image cluster to reduce overlap across benchmark splits.
\paragraph{Coverage.}
The final tasks cover 19 factory-automation categories (Appendix Figure~\ref{fig:source_taxonomy_main}), including mounting interfaces and repeated structures. Table~\ref{tab:source_pools} summarizes the scale of the candidate pools, and Appendix~\ref{sec:appendix_construction} provides additional construction details.

\begin{table}[htbp]
\centering
\caption{Source pools before final curation.}
\label{tab:source_pools}
\small
\begin{tabular}{@{}lrrl@{}}
\toprule
\textbf{Source} & \textbf{Raw scale} & \textbf{Filtered} & \textbf{Primary assets} \\
\midrule
Public industrial marketplace & $\sim 680$k & 366,487 & Product images \\
Purchased industrial data & $\sim 200$k & 125,489 & STEP files, 2D drawings, metadata \\
\bottomrule
\end{tabular}
\end{table}

\paragraph{Task construction.}
Task construction separates proposal, screening, reference verification, and evaluation. For each non-rendered instance, multiple candidate methods first propose reference 3D models, which are then screened for reconstruction quality and input--geometry consistency. Candidate 3D models are scored by CAD experts, and only about 17\% with scores of at least 8 are retained as benchmark references. Accepted references are frozen only after practitioner cross-checks within the curation workflow. Proposal methods, screening models, evaluated frontier large models and agents, and the Judge remain distinct throughout construction and evaluation (Appendix~\ref{sec:appendix_construction}). The same grouping rules also keep Judge-development cases separate from reported evaluation. RCB-Assm25 preserves the category, component-count, visibility, and geometry-complexity strata of all 167 assemblies.
\subsection{Evaluation Protocol}

\paragraph{Required output.}
Every submission must create a valid FreeCAD document and export \texttt{result.stl} in millimetres. A part must contain at least one non-empty solid, while an assembly may contain multiple solids. Geometric metrics are computed on the exported union. The shared runtime keeps frontier large models and agents comparable. Feature histories, constraints, tolerances, assembly trees, and manufacturing readiness matter only when they affect the exported 3D model. The visual-semantic identity Judge can assess visible structure and assembly relations, but it cannot recover parametric metadata that is not present in the export.

\paragraph{Scoring.}
\emph{Executability} requires a valid exported artifact, so syntax errors, runtime failures, missing outputs, empty geometry, and invalid models all count as failures. After deterministic signed-PCA alignment removes pose and uniform-scale differences, \emph{Solid IoU} measures occupied volume, while \emph{Surface IoU} emphasizes boundaries and thin details. The rubric-based \emph{visual-semantic identity Judge} compares rendered views of the exported model with the original inputs and never sees the reference model. For text tasks, the Judge uses the corresponding real picture evidence from which the text description was generated and validated, so its score remains independent of the geometric reference 3D model. The Part rubric assesses identity and salient features, while the Assembly rubric also scores components, placement, and system structure. To develop the Judge, we sampled benchmark input evidence and reference pairs and evaluated candidate prompts. We then iteratively refined the protocol through CAD-expert review, preference screening, and post-hoc validation before freezing the Kimi K2.6 protocol~\citep{kimi26_2026}. Appendix~\ref{sec:appendix_judge_reliability} reports a six-system consistency check on BenchCAD, and Appendix~\ref{sec:appendix_judge_prompts} provides the full prompt details.

For a reported setting \(c\), let \(E_c\), \(G_c^{\mathrm{solid}}\), \(G_c^{\mathrm{surface}}\), and \(J_c\) denote the four aggregate scores. Their displayed summary is
\begin{equation}
P_c=\tfrac{1}{4}\left(E_c+G_c^{\mathrm{solid}}+G_c^{\mathrm{surface}}+J_c/100\right).
\label{eq:profile_average}
\end{equation}
We refer to \(P_c\) as the \emph{Profile Average} (PA). Here \(E_c\) uses every assigned task, whereas each quality term uses only artifacts available to its evaluator. Section~\ref{sec:evaluations} interprets the component columns together with their denominators. Appendix~\ref{sec:appendix_metrics} gives the alignment, voxelization, and aggregation details.

\section{Experiments}
\label{sec:evaluations}

We study three questions on the reported evaluation slice: \textbf{RQ1.} \emph{Do executability, geometric IoUs, and the visual-semantic identity Judge favor the same frontier large models?} \textbf{RQ2.} \emph{Does difficulty follow a single order across the four Part regimes, or does that ordering depend on the metric and the input type?} \textbf{RQ3.} \emph{What recurring failure modes remain when frontier large models and agents are applied to realistic CAD modeling tasks?}

\subsection{Setup}
\label{sec:experimental_setup}

\paragraph{Models.}
We evaluate nine standalone frontier large models on the reported Part slice: Qwen3-VL-8B~\cite{qwen3vl2025}, Qwen3-VL-32B~\cite{qwen3vl2025}, Qwen3.8-27B~\cite{qwen38_2026}, Claude Opus 4.8~\cite{claudeopus48_2026}, Gemini 3.1 Pro Preview~\cite{gemini31pro_2026}, GPT-5.4~\cite{gpt54_2026}, GPT-5.5~\cite{gpt55_2026}, Kimi K3~\cite{kimik3_2026}, and Doubao Seed 2.0 Pro~\cite{seed20_2026}. Qwen3-VL-8B, Qwen3-VL-32B, and Kimi K3 are open-weight, while Claude Opus 4.8, Gemini 3.1 Pro Preview, GPT-5.4, GPT-5.5, and Doubao Seed 2.0 Pro are proprietary. All request settings use the provider default. The three Qwen baselines are limited to the Part evaluation, whereas the remaining six frontier-scale large models also enter the assembly comparison. Each run record preserves the evaluated model version, provider settings, and platform metadata.

\paragraph{Agent comparisons.}
On RCB-Assm25, we additionally evaluate Codex with GPT-5.5~\cite{codex_2026} and Claude Code with Claude Opus 4.8~\cite{claudecode_2026}. These are general-purpose coding agents with access to files, execution, error observation, and revision, but no CAD-specific tools. We compare each agent with its standalone model on the same tasks under the recorded budget, so the evaluated object is the complete method rather than the base model alone.

\paragraph{Aggregation.}
Standalone large models share the task instruction, the FreeCAD backend, and the maximum context, while request settings and agent limits follow the recorded provider or tool defaults. PA follows Equation~\ref{eq:profile_average}. For the Part Track, regime-balanced PA macro-averages the four regime-specific PAs, and the comparisons below are interpreted through the component columns. Appendix~\ref{sec:appendix_metrics} gives the alignment, voxelization, and aggregation details.

\subsection{RQ1: Different metrics favor different frontier large models}
\label{sec:main_results}

\paragraph{Executability and surface recovery diverge.}
Table~\ref{tab:part_main} reports the regime-balanced Part Track results, while Appendix~\ref{sec:appendix_part_full} provides the full regime-specific breakdown. GPT-5.5 executes successfully on $93.11\%$ of tasks in the regime-balanced average, but its conditional Surface IoU is only $0.1393$. Kimi K3 attains the highest conditional Solid IoU and Surface IoU, yet it executes on only $34.22\%$ of tasks and reaches a Surface IoU of just $0.1657$. The main finding is that surface recovery remains limited even among successful executions.

\paragraph{Different frontier large models lead different metrics.}
Gemini 3.1 Pro Preview has the highest regime-balanced PA ($0.5507$). GPT-5.5 leads executability and the visual-semantic identity Judge, while Kimi K3 leads both IoU metrics. No frontier large model leads all four metrics.

Among open models, Qwen3.8-27B executes on $79.61\%$ of tasks, close to GPT-5.4. However, its conditional Surface IoU is only $0.1126$ and Judge score is $56.30$. From Qwen3-VL-8B to Qwen3-VL-32B, executability rises from $0.2100$ to $0.3322$, and Surface IoU from $0.0885$ to $0.0995$. For the 8B and 32B Qwen3-VL models, limited execution remains the dominant constraint.

\begin{table*}[htbp]
\centering
\scriptsize
\caption{Regime-balanced Part Track results. Each component is macro-averaged across the four input regimes, and regime-balanced PA macro-averages the four regime-specific PAs. The three rightmost columns are conditional on evaluator-available artifacts and should be read together with Exec. Appendix Tables~\ref{tab:part_full_text}--\ref{tab:part_full_rendered} report the full regime-specific results. Best and second-best values are \textbf{bold} and \underline{underlined} only for Exec. and regime-balanced PA.}
\label{tab:part_main}
\setlength{\tabcolsep}{4pt}
\begin{tabular}{lccccc}
\toprule
Model & Exec. & \makecell{Solid IoU\\(cond.)} & \makecell{Surface IoU\\(cond.)} & \makecell{Judge\\(cond.)} & \makecell{Regime-\\balanced PA} \\
\midrule
Claude Opus 4.8~(\citealp{claudeopus48_2026})        & \underline{0.8922} & 0.4080 & 0.1433 & 68.31 & 0.5316 \\
Gemini 3.1 Pro Preview~(\citealp{gemini31pro_2026}) & 0.8825 & 0.4291 & 0.1578 & 73.33 & \textbf{0.5507} \\
GPT-5.4~(\citealp{gpt54_2026})                & 0.8005 & 0.3795 & 0.1337 & 61.19 & 0.4814 \\
GPT-5.5~(\citealp{gpt55_2026})                & \textbf{0.9311} & 0.3829 & 0.1393 & 73.79 & \underline{0.5478} \\
Kimi K3~(\citealp{kimik3_2026})                & 0.3422 & 0.4481 & 0.1657 & 72.23 & 0.4195 \\
Doubao Seed 2.0 Pro~(\citealp{seed20_2026})    & 0.4315 & 0.3767 & 0.1328 & 59.57 & 0.3842 \\
\midrule
Qwen3-VL-8B~(\citealp{qwen3vl2025})            & 0.2100 & 0.2833 & 0.0885 & 34.19 & 0.2309 \\
Qwen3-VL-32B~(\citealp{qwen3vl2025})           & 0.3322 & 0.2885 & 0.0995 & 44.07 & 0.2902 \\
Qwen3.8-27B~(\citealp{qwen38_2026})            & 0.7961 & 0.3585 & 0.1126 & 56.30 & 0.4575 \\
\bottomrule
\end{tabular}
\end{table*}

\subsection{RQ2: Different input regimes change the difficulty order}
\label{sec:modality_track_results}

\begin{figure*}[!t]
\centering
{\includegraphics[width=0.9\textwidth]{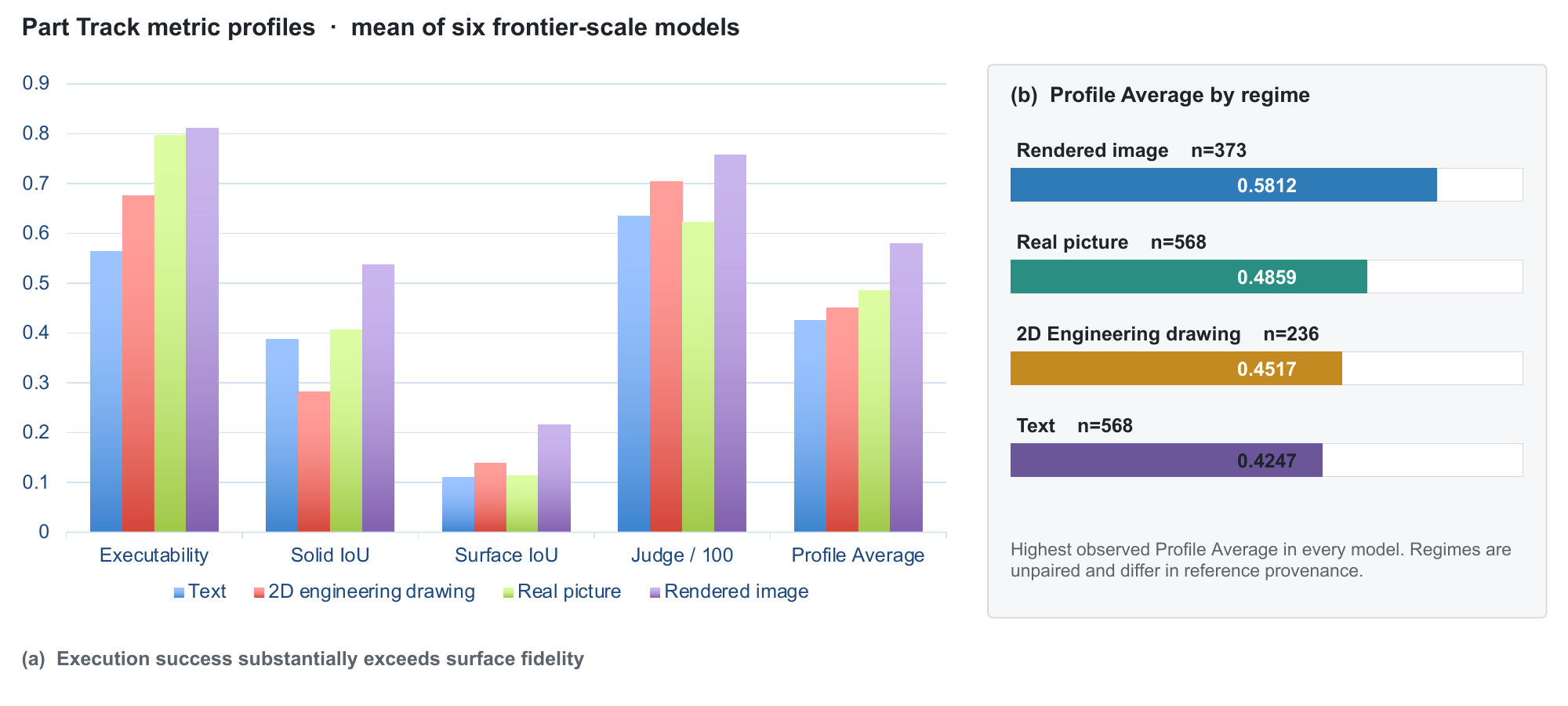}}
\caption{Part Track profiles averaged over six frontier-scale large models. Judge is divided by 100 for display. The three non-rendered regimes change order across metrics.}
\label{fig:modality_profile_main}
\end{figure*}

\paragraph{Rendered-image inputs lead every component.}
Across six frontier-scale large models, the rendered image regime leads every component: $0.8123$ executability, $0.5379$ Solid IoU, $0.2165$ Surface IoU, and $75.80$ Judge (Figure~\ref{fig:modality_profile_main}). The other three Part regimes also change order across metrics, so any average over the four regimes combines results from different input evidence types.

\paragraph{The ordering does not stay fixed beyond the rendered image regime.}
Among the remaining three Part regimes, no single condition leads every metric. As Figure~\ref{fig:modality_profile_main} shows, the regime ordering changes once evaluation moves from executability to IoUs and then to the Judge. This confirms that input type affects different aspects of CAD modeling in different ways.

\paragraph{Assembly Solid IoU is lower than on real picture parts.}
Across the six shared standalone frontier large models, assembly Solid IoU is $0.2290$, compared with $0.4071$ on real picture parts. By contrast, executability ($0.8133$ versus $0.7987$) and Judge ($64.95$ versus $62.39$) are similar. Because the two studies are unmatched, the $0.1781$ difference reflects two different populations rather than a paired comparison. This pattern suggests that, once multiple components must fit together, an exported model can look plausible in a render while still getting the overall solid composition wrong.

\subsection{RQ3: Frontier models and agents reveal recurring failure modes}
\label{sec:agent_failure_modes}

\paragraph{Agent comparisons expose recurring failure modes.}
The same separation appears in the assembly setting. GPT-5.4 and Codex + GPT-5.5 both execute successfully on all 25 tasks in RCB-Assm25 (Table~\ref{tab:assembly_main}), yet their Surface IoU values are only $0.1046$ and $0.1161$, respectively. Codex leads the two IoU metrics, while standalone GPT-5.5 leads the visual-semantic identity Judge. Relative to standalone GPT-5.5, Codex + GPT-5.5 improves executability by $0.1600$, Solid IoU by $0.0714$, and Surface IoU by $0.0190$, but lowers the Judge by $6.98$ pp. Relative to standalone Claude Opus 4.8, Claude Code changes PA by only $+0.0006$, while Surface IoU decreases from $0.1058$ to $0.0864$. These comparisons treat agents as complete evaluated methods under the recorded budgets. Appendix~\ref{sec:appendix_harness_assm} traces recurring failure modes in which execution is repaired while product identity, fine structures, or assembly placement remain weak.

\begin{table*}[htbp]
\centering
\scriptsize
\caption{Results on RCB-Assm25, the stratified study used for all reported assembly comparisons. Executability uses all 25 tasks. The three rightmost columns are conditional on evaluator-available artifacts and should be read together with Exec. Judge is reported on a $0$--$100$ scale. Best and second-best values are \textbf{bold} and \underline{underlined} only for Exec. and PA.}
\label{tab:assembly_main}
\setlength{\tabcolsep}{4.5pt}
\begin{tabular}{lcccccc}
\toprule
Model & PA & Exec. & \makecell{Solid IoU\\(cond.)} & \makecell{Surface IoU\\(cond.)} & \makecell{Judge\\(cond.)} \\
\midrule
Codex + GPT-5.5~(\citealp{codex_2026})              & \textbf{0.5228} & \textbf{1.0000} & 0.2817 & 0.1161 & 69.35 \\
Claude Code + Claude Opus 4.8~(\citealp{claudecode_2026}) & 0.4644 & 0.8800 & \underline{0.2732} & 0.0864 & 61.80 \\
\midrule
Claude Opus 4.8~(\citealp{claudeopus48_2026})        & 0.4638 & 0.8800 & 0.2670 & 0.1058 & 60.24 \\
Gemini 3.1 Pro Preview~(\citealp{gemini31pro_2026}) & 0.4724 & 0.9200 & 0.2306 & 0.0979 & 64.10 \\
GPT-5.4~(\citealp{gpt54_2026})                & \underline{0.5028} & \textbf{1.0000} & 0.2366 & 0.1046 & 66.98 \\
GPT-5.5~(\citealp{gpt55_2026})                & 0.4777 & 0.8400 & 0.2103 & 0.0971 & 76.33 \\
Kimi K3~(\citealp{kimik3_2026})                & 0.4338 & 0.7200 & 0.2665 & 0.1039 & 64.48 \\
Doubao Seed 2.0 Pro~(\citealp{seed20_2026})    & 0.3389 & 0.5200 & 0.1632 & 0.0969 & 57.56 \\
\bottomrule
\end{tabular}
\end{table*}

\section{Discussion}
\label{sec:analysis_discussion}

\begin{figure}[htbp]
\centering
\vspace{-12pt}
\includegraphics[width=0.9\columnwidth]{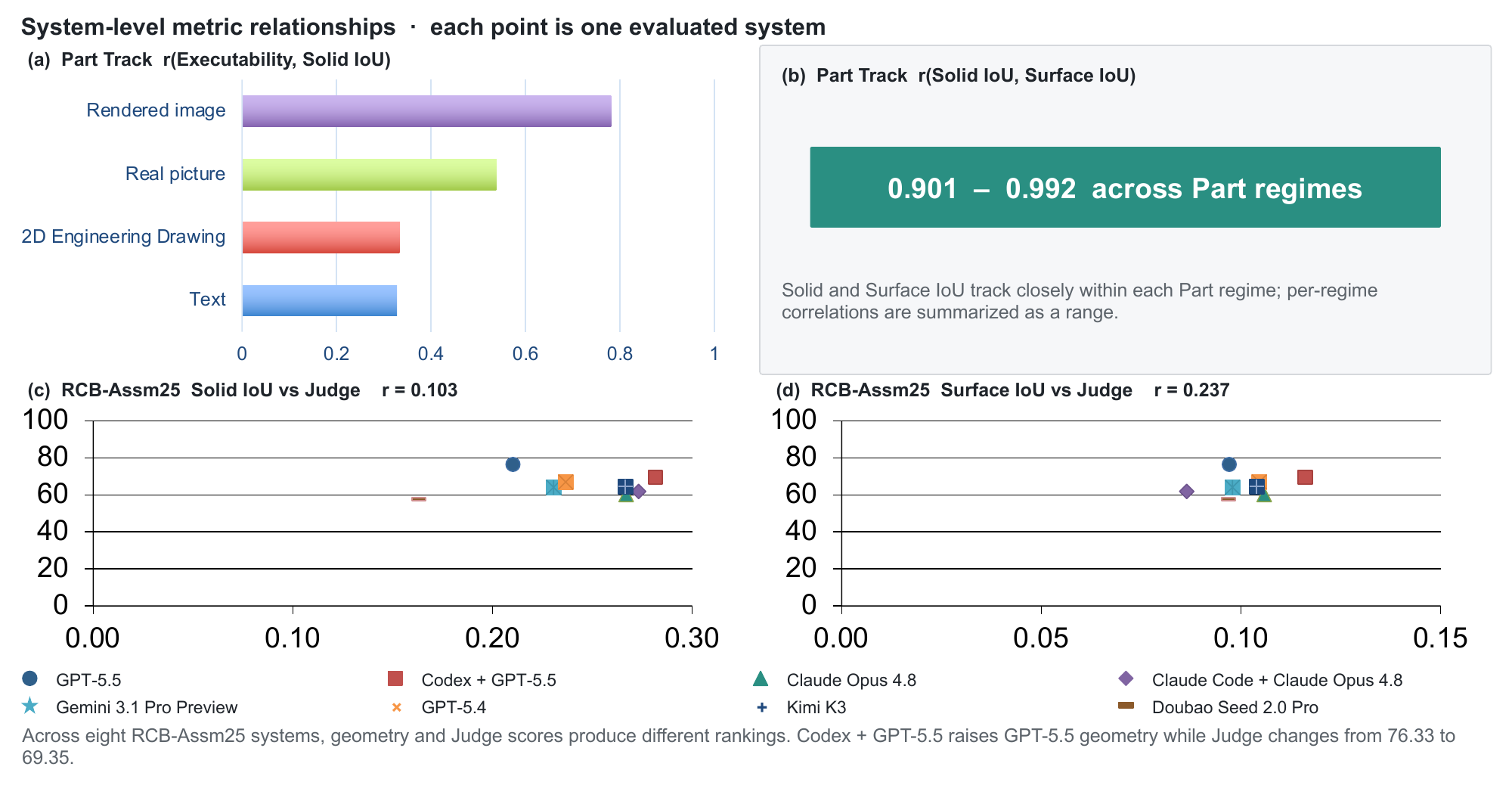}
\caption{System relationships vary by metric. Solid and Surface IoU are tightly coupled within Part regimes, whereas geometric IoUs and Judge scores are only weakly correlated across RCB-Assm25 configurations ($n = 8$).}
\label{fig:metric_correlations}
\vspace{-4pt}
\end{figure}

The results indicate that RealCADBench measures complementary aspects of system performance, including executability, geometric IoUs, and visual-semantic identity Judge. This pattern is already evident in Tables~\ref{tab:part_main} and~\ref{tab:assembly_main}. Figure~\ref{fig:metric_correlations} further shows that systems with similar geometric IoUs can nonetheless differ substantially in Judge scores. Appendix~\ref{sec:appendix_analysis} provides the aggregate values underlying the modality-profile and track-comparison analyses.

\paragraph{Different metrics favor different frontier large models.}
RQ1 shows that the reported metrics do not rank systems in the same way. The model that leads on executability is not the one that leads on IoUs, and the highest regime-balanced PA comes from a different model again. This difference matters because the metrics answer different questions. Executability asks whether a system can produce a valid artifact at all. The IoUs ask how much of the target geometry is recovered once such an artifact exists. The quality columns are computed only on tasks with valid executions, so they should be read together with the number of tasks that actually enter each average. Otherwise, a model that succeeds on few tasks can still look strong because its IoUs or Judge is averaged over a much smaller subset.

\paragraph{Different input regimes shift the difficulty order.}
RQ2 shows that the four Part regimes do not follow a single difficulty order. The rendered image regime is consistently the easiest under the reported metrics, but the other three regimes change order as evaluation moves from executability to geometric IoUs and then to the Judge. This pattern suggests that the benchmark is not measuring one generic notion of CAD difficulty. Instead, different input types make different parts of the CAD modeling problem difficult. Some regimes mainly test visible shape recovery, whereas others put more weight on hidden geometry inference or identity preservation from incomplete evidence.

\paragraph{Agents alter the trade-offs among metrics.}
RQ3 shows that agents should be treated as distinct evaluated methods rather than as simple wrappers around the same base model. In the assembly setting, tool use and iterative repair can improve execution success and sometimes improve geometry, but they do not necessarily improve visual-semantic identity. An agent loop can fix syntax errors, execution failures, and some construction mistakes without necessarily correcting the core design decision. The comparison is therefore between complete evaluated methods, not base models alone. Each method includes its harness, tools, and budget.

\paragraph{The Judge captures what IoUs miss.}
The visual-semantic identity Judge matters because IoUs do not fully capture product identity, salient features, or assembly relations. It compares the exported model with the input evidence rather than with the reference model. It therefore complements the IoUs instead of duplicating them. In practice, it asks whether the model still looks like the intended object, not just how much geometry overlaps after alignment. RealCADBench therefore reports the Judge as a separate metric rather than treating IoUs alone as sufficient. The consistency check in Appendix~\ref{sec:appendix_judge_reliability} supports this choice, and expert post-checks remain part of validation on evaluated outputs.

\paragraph{Recurring failure modes explain the metric separation.}
The qualitative examples help explain why the reported metrics diverge. Some outputs match the overall shape yet omit thin structures or small interfaces that matter for use. Others obtain non-trivial IoUs while still missing the visual characteristics that determine product identity, particularly when hidden geometry is not specified by the input. For assemblies, rendered views may make local component placements appear reasonable even when the resulting solid composition is incorrect. These failure modes reinforce that executability, IoUs, and visual-semantic identity measure different aspects of quality.


\section{Conclusion}
\label{sec:conclusion}

We introduced \texttt{RealCADBench}, a benchmark for intent-to-program CAD modeling from real industrial design intents, including text, 2D engineering drawings, real product pictures, rendered images, and multi-part assemblies. It places these tasks under a shared FreeCAD artifact contract and evaluates outputs with four measures, including executability, Solid IoU, Surface IoU, and a rubric-based visual-semantic identity Judge.

The results show that no single metric is sufficient for realistic CAD modeling. Frequent successful execution does not imply strong IoUs, and stronger IoUs do not guarantee recovery of product identity or assembly structure. Across the reported Part and Assembly studies, different systems lead different metrics, and the difficulty order shifts across input regimes. Agent configurations should also be treated as complete evaluated methods rather than as wrappers around the same base model. The qualitative analyses further show recurring failure modes in fine structures, part identity, and assembly placement.

These findings matter for how realistic CAD modeling is evaluated. Executability, IoUs, and visual-semantic identity should be reported separately rather than collapsed into a single score or treated as if execution alone were sufficient. RealCADBench is designed to make these trade-offs visible on realistic industrial design intents.

In future, we see three immediate directions. We will publicly release the full RealCADBench benchmark, report results on the full benchmark rather than only on the current evaluation slice, and compare a broader set of specialized AI-for-CAD algorithms alongside frontier large models and general-purpose agents.

\clearpage

\bibliography{iclr2027_conference}

@inproceedings{abc,
    title={{ABC}: A Big {CAD} Model Dataset for Geometric Deep Learning},
    author={Koch, Sebastian and Matveev, Albert and Jiang, Zhongshi and
  Williams, Francis and Artemov, Alexey and Burnaev, Evgeny and Alexa, Marc and
  Zorin, Denis and Panozzo, Daniele},
    booktitle={Proceedings of the IEEE/CVF Conference on Computer Vision and
  Pattern Recognition (CVPR)},
    pages={9601--9611},
    year={2019}
  }

@article{fusion360,
  title={Fusion 360 gallery: A dataset and environment for programmatic cad construction from human design sequences},
  author={Willis, Karl DD and Pu, Yewen and Luo, Jieliang and Chu, Hang and Du, Tao and Lambourne, Joseph G and Solar-Lezama, Armando and Matusik, Wojciech},
  journal={ACM Transactions on Graphics (TOG)},
  volume={40},
  number={4},
  pages={1--24},
  year={2021},
  publisher={ACM New York, NY, USA}
}

@inproceedings{deepcad,
  title={Deepcad: A deep generative network for computer-aided design models},
  author={Wu, Rundi and Xiao, Chang and Zheng, Changxi},
  booktitle={2021 IEEE/CVF International Conference on Computer Vision (ICCV)},
  pages={6752--6762},
  year={2021},
  organization={IEEE}
}

@inproceedings{sketchgraphs,
    title={{SketchGraphs}: A Large-Scale Dataset for Modeling Relational
  Geometry in Computer-Aided Design},
    author={Seff, Ari and Ovadia, Yaniv and Zhou, Wenda and Adams, Ryan P},
    booktitle={ICML 2020 Workshop on Object-Oriented Learning},
    pages={8614--8624},
    year={2020}
  }

@inproceedings{text2cad,
    author    = {Khan, Mohammad Sadil and Sinha, Sankalp and Sheikh, Talha Uddin
  and Stricker, Didier and Ali, Sk Aziz and Afzal, Muhammad Zeshan},
    title     = {{Text2CAD}: Generating Sequential {CAD} Designs from
  Beginner-to-Expert Level Text Prompts},
    booktitle = {Advances in Neural Information Processing Systems},
    volume    = {37},
    pages     = {7552--7579},
    year      = {2024},
    url       = {https://proceedings.neurips.cc/paper_files/paper/2024/file/0e5b96f97c1813bb75f6c28532c2ecc7-Paper-Conference.pdf}
  }

@article{text2cadbench,
    title={{Text2CAD-Bench}: A Benchmark for {LLM}-based Text-to-Parametric
  {CAD} Generation},
    author={Wang, Liang and Meng, Heng and Xiang, Zekai and Liu, Jin and
  Zhou, Pingyi and Chen, Litao and Tang, Yongqiang},
    journal={arXiv preprint arXiv:2605.18430},
    year={2026}
  }

@inproceedings{cadrecode,
    title={{CAD-Recode}: Reverse Engineering {CAD} Code from Point Clouds},
    author={Rukhovich, Danila and Dupont, Elona and Mallis, Dimitrios and
  Cherenkova, Kseniya and Kacem, Anis and Aouada, Djamila},
    booktitle={Proceedings of the IEEE/CVF International Conference on Computer
  Vision (ICCV)},
    year={2025}
  }

@article{zerotocad,
    title={{Zero-to-CAD}: Agentic Synthesis of Interpretable {CAD} Programs at
  Million-Scale Without Real Data},
    author={Ataei, Mohammadmehdi and Askari, Farzaneh and Malekshan, Kamal
  Rahimi and Jayaraman, Pradeep Kumar},
    journal={arXiv preprint arXiv:2604.24479},
    year={2026}
  }

@article{unicad,
    title={{UniCAD}: A Unified Benchmark and Universal Model for Multi-Modal
  Multi-Task {CAD}},
    author={Chen, Jingyuan and Jin, Sheng and Sun, Haopeng and Liu, Wentao and
  Qian, Chen},
    journal={arXiv preprint arXiv:2606.05058},
    year={2026}
  }

@misc{parametriccadbench,
    title={Parametric CAD Bench},
    author={cadbench.ai},
    year={2026},
    howpublished={\url{https://cadbench.ai/zh}},
    note={Accessed August 13, 2026}
  }

@article{cadbench2026,
    author  = {Doris, Anna C. and Sony, Jacob Thomas and Nehme, Ghadi and Syla,
  Era and Nobari, Amin Heyrani and Ahmed, Faez},
    title   = {{CADBench}: A Multimodal Benchmark for AI-Assisted {CAD} Program
  Generation},
    journal = {arXiv preprint arXiv:2605.10873},
    year    = {2026},
    url     = {https://arxiv.org/abs/2605.10873}
  }

@article{benchcad2026,
    author  = {Zhang, Haozhe and Liu, Kaichen and Chen, Miaomiao and Li, Lei and
  Yang, Shaojie and Peng, Cheng and Chen, Hanjie},
    title   = {{BenchCAD}: A Comprehensive, Industry-Standard Benchmark for
  Programmatic {CAD}},
    journal = {arXiv preprint arXiv:2605.10865},
    year    = {2026},
    url     = {https://arxiv.org/abs/2605.10865}
  }

@article{p3dbench2026,
    author  = {Yang, Yikang and Hu, Zhanpeng and Lin, Youtian and Zhou, Mengqi
  and Xu, Jingxi and Zhang, Feihu and Liu, Jiaheng and Yao, Yao},
    title   = {{P3D-Bench}: Benchmarking {MLLMs} for Parametric 3D Generation
  and Structural Reasoning},
    journal = {arXiv preprint arXiv:2606.11152},
    year    = {2026},
    url     = {https://arxiv.org/abs/2606.11152}
  }

@misc{cadgenbench2026,
    author       = {{Hugging Face}},
    title        = {{CADGenBench}: A Benchmark for AI-Driven {CAD} Generation
  and Editing},
    year         = {2026},
    howpublished = {\url{https://github.com/huggingface/cadgenbench}},
    note         = {Accessed: 2026-08-13}
  }

@article{muse2026,
    author  = {Dong, Xiaoyu and Li, Zhi and Wu, Xiao-Ming},
    title   = {{MUSE}: Benchmarking Manufacturable, Functional, and Assemblable
  Text-to-{CAD} Generation},
    journal = {arXiv preprint arXiv:2605.28579},
    year    = {2026},
    url     = {https://arxiv.org/abs/2605.28579}
  }

@article{cadengbench2026,
    title={{CADEngBench}: It Looks Like {CAD}, but Does It Work? Evaluating
  Parametric Design, Assembly Reasoning, and Physics Simulation},
    author={Singh, Harmanjot and Dubey, Abhra and Herrera, Jorge Alejandro
  Amador},
    journal={arXiv preprint arXiv:2608.09296},
    year={2026},
    url={https://arxiv.org/abs/2608.09296}
  }

@article{qwen3vl2025,
  author  = {Bai, Shuai and others},
  title   = {{Qwen3-VL} Technical Report},
  journal = {arXiv preprint arXiv:2511.21631},
  year    = {2025},
  url     = {https://arxiv.org/abs/2511.21631}
}

@misc{qwen38_2026,
    title = {{Qwen3.8-Max}: A New Bar for Coding and Cowork},
    url = {https://qwen.ai/blog?id=qwen3.8},
    author = {{Qwen Team}},
    month = {August},
    year = {2026}
}

@misc{gpt54_2026,
  author       = {{OpenAI}},
  title        = {Introducing {GPT-5.4}},
  year         = {2026},
  howpublished = {\url{https://openai.com/index/introducing-gpt-5-4/}},
  note         = {Published March 5, 2026; accessed August 13, 2026}
}

@misc{gpt55_2026,
  author       = {{OpenAI}},
  title        = {{GPT-5.5} System Card},
  year         = {2026},
  howpublished = {\url{https://openai.com/index/gpt-5-5-system-card/}},
  note         = {Published April 23, 2026; accessed August 13, 2026}
}

@misc{claudeopus48_2026,
  author       = {{Anthropic}},
  title        = {Introducing {Claude Opus 4.8}},
  year         = {2026},
  howpublished = {\url{https://www.anthropic.com/news/claude-opus-4-8}},
  note         = {Published May 28, 2026; accessed August 13, 2026}
}

@misc{gemini31pro_2026,
  author       = {{Google DeepMind}},
  title        = {{Gemini 3.1 Pro} Model Card},
  year         = {2026},
  howpublished = {\url{https://deepmind.google/models/model-cards/gemini-3-1-pro/}},
  note         = {Published February 19, 2026; accessed August 13, 2026}
}

@misc{gemini3pro_2026,
  author       = {{Google DeepMind}},
  title        = {{Gemini 3 Pro} Model Card},
  year         = {2026},
  howpublished = {\url{https://deepmind.google/models/model-cards/gemini-3-pro/}},
  note         = {Model release November 2025; last updated May 2026; accessed September 1, 2026}
}

@article{kimik3_2026,
  author  = {{Moonshot AI}},
  title   = {{Kimi K3}: Open Frontier Intelligence},
  journal = {arXiv preprint arXiv:2607.24653},
  year    = {2026},
  url     = {https://arxiv.org/abs/2607.24653}
}

@misc{kimi26_2026,
  author       = {{Moonshot AI}},
  title        = {{Meet Kimi K2.6: Advancing Open-Source Coding}},
  year         = {2026},
  howpublished = {\url{https://forum.moonshot.ai/t/meet-kimi-k2-6-advancing-open-source-coding/369}},
  note         = {Announcement, published April 21, 2026; accessed September 1, 2026}
}

@misc{seed20_2026,
  author       = {{ByteDance Seed}},
  title        = {{Seed2.0} Model Card: Towards Intelligence Frontier for Real-World Complexity},
  year         = {2026},
  howpublished = {\url{https://research.doubao.com/en/seed2}},
  note         = {Accessed August 13, 2026}
}

@misc{codex_2026,
  author       = {{OpenAI}},
  title        = {{Codex} Manual},
  year         = {2026},
  howpublished = {\url{https://developers.openai.com/codex/codex-manual.md}},
  note         = {Accessed August 13, 2026}
}

@misc{claudecode_2026,
  author       = {{Anthropic}},
  title        = {{Claude Code} Overview},
  year         = {2026},
  howpublished = {\url{https://docs.anthropic.com/en/docs/claude-code/overview}},
  note         = {Accessed August 13, 2026}
}

@misc{freecad_software,
  author       = {Riegel, J{\"u}rgen and Mayer, Werner and van Havre, Yorik and {FreeCAD Community}},
  title        = {{FreeCAD}: Your Own 3D Parametric Modeler},
  year         = {2026},
  howpublished = {\url{https://www.freecad.org/}},
  note         = {Software; accessed August 13, 2026}
}
\bibliographystyle{iclr2027_conference}

\clearpage
\appendix
\renewcommand{\topfraction}{0.9}
\renewcommand{\bottomfraction}{0.8}
\renewcommand{\textfraction}{0.07}
\renewcommand{\floatpagefraction}{0.8}
\setlength{\textfloatsep}{10pt plus 2pt minus 2pt}
\setlength{\floatsep}{8pt plus 2pt minus 2pt}
\setlength{\intextsep}{8pt plus 2pt minus 2pt}
\section{Appendix}
\label{sec:appendix}
\suppressfloats[t]




\subsection{Additional Benchmark Construction Details}
\label{sec:appendix_construction}

\begin{figure}[tbp]
    \centering
    {\includegraphics[width=\textwidth]{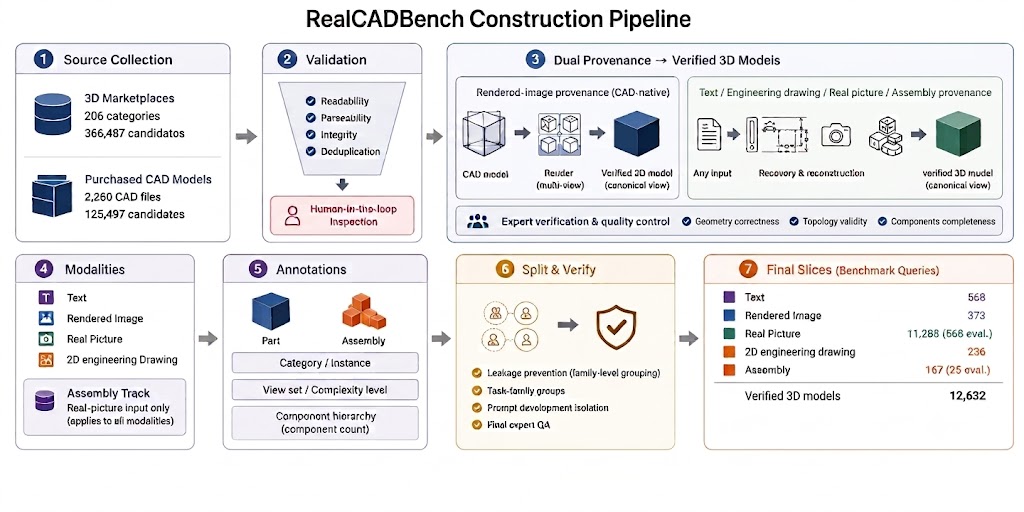}}
    \caption{Construction pipeline from industrial source pools to final benchmark tasks. All reported tasks use verified benchmark references. When original vendor CAD is unavailable, the benchmark reference is constructed through multi-system proposals, Gemini editing, Gemini~3 Pro screening~\citep{gemini3pro_2026}, and practitioner verification. The Assembly Track contains 167 tasks, and RCB-Assm25 is the  stratified study used for the reported assembly comparisons.}
    \label{fig:appendix_construction_pipeline}
\end{figure}

This section adds construction details for the protocol summarized in Section~\ref{sec:benchmark_construction}. We describe how benchmark references are constructed and verified across regimes, why we report RCB-Assm25 in the main paper, and how release scope is defined. Figure~\ref{fig:appendix_construction_pipeline} summarizes the path from source validation to final task formation.

\begin{figure}[tbp]
    \centering
    {\includegraphics[width=0.88\textwidth]{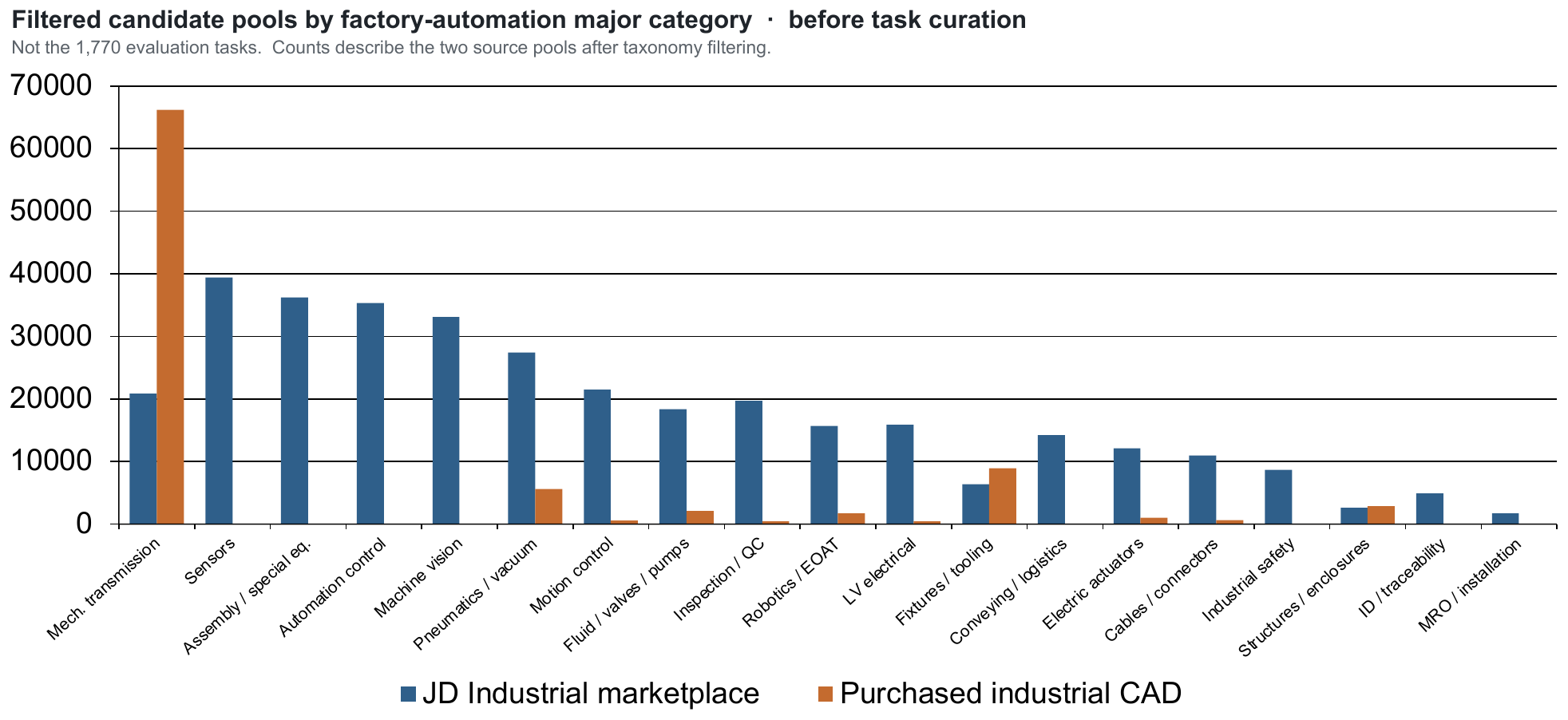}}
    \caption{Category composition of the two filtered candidate pools across 19 factory-automation categories before final task curation.}
    \label{fig:source_taxonomy_main}
\end{figure}

\paragraph{Reference construction for non-rendered tasks.}
For text, 2D engineering drawing, real picture, and assembly tasks, paired vendor CAD is usually unavailable. These tasks therefore follow a staged proposal-and-verification workflow to produce verified benchmark references. Multiple candidate methods first propose candidate 3D models. Gemini-based editing then refines surviving candidates, and Gemini~3 Pro screens reconstruction quality, input quality, and image--geometry consistency~\citep{gemini3pro_2026}. Five professional CAD practitioners participate across the four Part regimes and the Assembly Track. They review the surviving candidates inside the curation workflow, score the candidate 3D models, and accept a reference only after cross-checks. Consistent with Section~\ref{sec:benchmark_construction}, only about 17\% of candidates with scores of at least 8 are retained as benchmark references. Text tasks are derived from corresponding real picture cases through Gemini 3.0 Pro generation and validation, and the Judge for the text regime therefore uses the corresponding real picture evidence. Gemini~3 Pro used in curation is distinct from the evaluated Gemini~3.1 Pro Preview, and the frozen Judge is Kimi K2.6. Proposal methods, screening models, evaluated frontier large models and agents, and the Judge remain separated throughout construction and evaluation.

\paragraph{Reference construction for rendered image tasks.}
Rendered-image tasks follow the same verified-reference principle, but their references come directly from the underlying CAD asset. For each rendered image task, the source STEP asset must parse, tessellate, render, and export to a non-empty STL. The standardized input render and the benchmark reference 3D model are then generated from that same verified asset.

\paragraph{Why RCB-Assm25.}
The Assembly Track contains 167 accepted tasks and forms the benchmark pool. In this paper, we report RCB-Assm25, a stratified study used for all assembly comparisons. Stratification covers category, component count, visibility, and geometry-complexity metadata already present in the track. The study is configuration-complete: six standalone models and two agents are scored on the same 25 tasks. Mean wall-clock time on the diagnostic sheet is $12.1$\,min for Codex and $31.6$\,min for Claude Code (Appendix~\ref{sec:appendix_harness_assm}). Future evaluations can extend the same protocol to the full 167-task track.

\paragraph{Selection and quality control.}
We apply source-specific screening followed by common eligibility checks. For visual inputs, we remove corrupted files, exact or near duplicates, records with ambiguous product identity, and samples that cannot be associated with a consistent product version. For CAD files, we require successful parsing, non-empty geometry, successful tessellation or rendering, and successful STEP-to-STL export when the asset is used in the rendered image regime. Assembly candidates must additionally contain multiple components with recoverable relative placement.

\paragraph{Release boundary.}
The final public release includes the full benchmark, together with the evaluation code and frozen prompts. It also specifies the scoring environment, including FreeCAD 1.1.1, Python 3.11, and dependency requirements. The release follows the same access conditions and source-license constraints. Per-instance release follows those same conditions. The benchmark tasks and evaluation protocol remain defined at full benchmark scope rather than through a small illustrative subset.

\subsection{Metric Implementation and Aggregation Details}
\label{sec:appendix_metrics}

Section~\ref{sec:benchmark_construction} introduces the four reported scores and Profile Average. This section makes the pairwise definitions, implementation choices, and aggregation sets explicit.

\paragraph{Pairwise geometric metrics.}
Let \(\tilde C_i\) denote the aligned prediction for task \(i\), and \(C_i^\star\) its reference. Solid and Surface IoU reuse one transform and a shared grid of resolution \(R=96\) with \(2\%\) padding:
\begin{equation}
\begin{aligned}
g_i^{\mathrm{solid}} &=
\frac{|V_{\mathrm{filled}}(\tilde C_i)\cap V_{\mathrm{filled}}(C_i^\star)|}
{|V_{\mathrm{filled}}(\tilde C_i)\cup V_{\mathrm{filled}}(C_i^\star)|},\\
g_i^{\mathrm{surface}} &=
\frac{|V_{\mathrm{boundary}}(\tilde C_i)\cap V_{\mathrm{boundary}}(C_i^\star)|}
{|V_{\mathrm{boundary}}(\tilde C_i)\cup V_{\mathrm{boundary}}(C_i^\star)|}.
\end{aligned}
\label{eq:geometry_metrics}
\end{equation}
Filled occupancy captures volume. Boundary occupancy is more sensitive to thin structures and local detail. Because alignment includes uniform scaling, these IoUs measure geometric recovery under the stated aligned reference rather than absolute millimetre accuracy.

\paragraph{Evaluation sets and denominators.}
Let \(c\) denote a reported setting, \(\mathcal{A}_c\) its assigned tasks, \(\mathcal{G}_c^q\) the tasks with a value from geometry evaluator \(q\in\{\mathrm{solid},\mathrm{surface}\}\), and \(\mathcal{J}_c\) the tasks with a Judge result. For task \(i\), executability is defined as
\begin{equation}
e_i=\mathbb{I}\!\left[p_i\text{ executes and produces a valid }\hat C_i\right].
\end{equation}
The reported aggregates are
\begin{equation}
\begin{aligned}
E_c &= \frac{1}{|\mathcal{A}_c|}\sum_{i\in\mathcal{A}_c} e_i,\\
G_c^q &= \frac{1}{|\mathcal{G}_c^q|}\sum_{i\in\mathcal{G}_c^q} g_i^q,
\qquad q\in\{\mathrm{solid},\mathrm{surface}\},\\
J_c &= \frac{1}{|\mathcal{J}_c|}\sum_{i\in\mathcal{J}_c} j_i,
\qquad j_i\in[0,100].
\end{aligned}
\label{eq:aggregate_metrics}
\end{equation}
Executability uses every assigned task. Each quality column averages the tasks that provide a value for the corresponding evaluator. Missing quality values are neither zero-filled nor inferred from \(e_i\). The released aggregation specification records the corresponding counts, and the component metrics are reported separately alongside PA.

\paragraph{PCA sign resolution.}
The right-handed PCA construction in Section~\ref{sec:benchmark_construction} leaves two principal-axis signs ambiguous. We enumerate
\(S(s_0,s_1)=\mathrm{diag}(s_0,s_1,s_0s_1)\)
for \(s_0,s_1\in\{-1,+1\}\), with candidate rotation
\(R=A_\star^{\mathsf T}S A_{\mathrm{pred}}\).
Here \(A_{\mathrm{pred}}\) and \(A_\star\) are the prediction and reference PCA bases. The selected candidate minimizes one-way mean nearest-neighbor distance from transformed prediction samples to reference samples. The subsequent uniform scale and center translation are applied once, and the resulting \(\tilde C_i\) is reused for both metrics in Equation~\ref{eq:geometry_metrics}.

\paragraph{Voxelization and fallback.}
The shared grid covers the joint bounding box with padding equal to \(2\%\) of its diagonal. Its pitch is the longest padded extent divided by \(R=96\). \(V_{\mathrm{filled}}\) fills mesh interiors, and \(V_{\mathrm{boundary}}\) keeps only cells that intersect the mesh surface, with no dilation. When direct surface voxelization is unavailable, boundary occupancy uses a histogram of deterministically sampled surface points on the same grid.

\paragraph{Judge implementation.}
The frozen Judge model is Kimi K2.6~\citep{kimi26_2026}. It is distinct from Gemini~3 Pro used during reference screening and from the evaluated Gemini~3.1 Pro Preview. Part Track artifacts use the Part Judge, whereas RCB-Assm25 artifacts use the Assembly Judge. For text tasks, the Judge uses the corresponding real picture evidence from which the text description was generated and validated. Judge development sampled benchmark input evidence and verified reference pairs across the four Part regimes and the Assembly Track, ran candidate prompts on those cases, and asked CAD practitioners to inspect the resulting judgments. Human review focused on rubric clarity, use of visual evidence, JSON validity, separation of part geometry from assembly relations, and reliable parser behavior. Candidate prompts were then revised and re-evaluated through repeated expert review and preference screening until practitioners reached consensus, after which the prompts were frozen for reported scoring. Expert review also includes follow-up checks on evaluated outputs. Each artifact receives one query under the frozen rubric. A separate visibility pass records \texttt{full}, \texttt{partial}, or \texttt{invisible} metadata for assembly components. These labels serve only as evidence metadata and do not change the scoring dimensions or weights. Verbatim prompts appear in Appendix~\ref{sec:appendix_judge_prompts}.

\paragraph{Regime-balanced PA.}
Equation~\ref{eq:profile_average} defines \(P_c\) for a reported setting \(c\). For the Part Track, let \(\mathcal{M}\) be the four input regimes. The reported regime-balanced score is
\begin{equation}
P_{\mathrm{part}}=\frac{1}{|\mathcal{M}|}\sum_{m\in\mathcal{M}}P_m.
\label{eq:part_overall}
\end{equation}
This macro-average weights the four regimes equally. Judge columns remain on the original \(0\)--\(100\) scale and are divided by 100 only in Equation~\ref{eq:profile_average}. Profile Average equally weights executability, Solid IoU, Surface IoU, and the normalized Judge. It provides a compact summary of the reported dimensions, while the component columns provide the corresponding breakdown.

\paragraph{Why geometry and Judge.}
Solid IoU and Surface IoU measure overlap with the benchmark reference 3D model after pose and uniform-scale alignment. The Judge instead compares renders of the exported mesh with the original input evidence and never sees that reference. For text tasks, it uses the corresponding real picture evidence from which the text description was generated and validated. The two families therefore capture complementary information, which is why both are reported.

\subsection{Additional Aggregate Results}
\label{sec:appendix_analysis}

Table~\ref{tab:appendix_modality_means} reports the exact values visualized in Figure~\ref{fig:modality_profile_main}.

\begin{table}[tbp]
\centering
\scriptsize
\caption{Mean Part Track profiles of the six frontier-scale models by input regime. Solid IoU, Surface IoU, and Judge are reported on the tasks that provide the corresponding evaluator outputs.}
\label{tab:appendix_modality_means}
\setlength{\tabcolsep}{6pt}
\begin{tabular}{lccccc}
\toprule
Input & Exec. & \makecell{Solid\\(cond.)} & \makecell{Surface\\(cond.)} & \makecell{Judge\\(cond.)} & \makecell{PA} \\
\midrule
Text                & 0.5648 & 0.3871 & 0.1117 & 63.53 & 0.4247 \\
2D Engineering Drawing & 0.6775 & 0.2841 & 0.1397 & 70.55 & 0.4517 \\
Real Picture          & 0.7987 & 0.4071 & 0.1138 & 62.39 & 0.4859 \\
Rendered Image      & \textbf{0.8123} & \textbf{0.5379} & \textbf{0.2165} & \textbf{75.80} & \textbf{0.5812} \\
\bottomrule
\end{tabular}
\end{table}

\begin{figure}[t]
\centering
{\includegraphics[width=\textwidth]{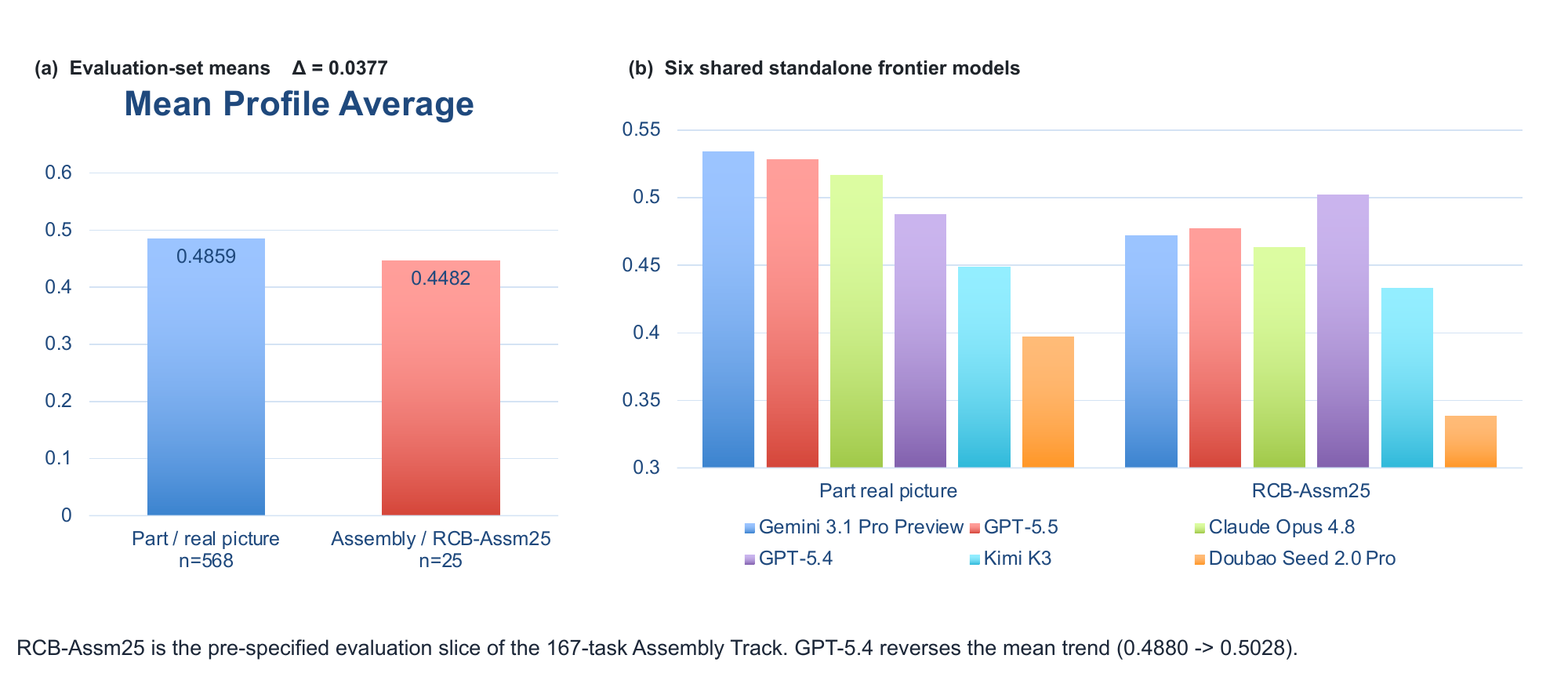}}
\caption{Profiles for the real picture Part regime and RCB-Assm25 across the six shared standalone frontier models.}
\label{fig:appendix_track_comparison}
\end{figure}

Figure~\ref{fig:metric_correlations} in the main text shows the system-level relationships among metrics. Figure~\ref{fig:appendix_harness_delta} below shows the paired agent differences on RCB-Assm25. Instance-level diagnostic traces for both outer loops appear in Appendix~\ref{sec:appendix_harness_assm}.

\begin{figure}[t]
\centering
\includegraphics[width=0.76\textwidth]{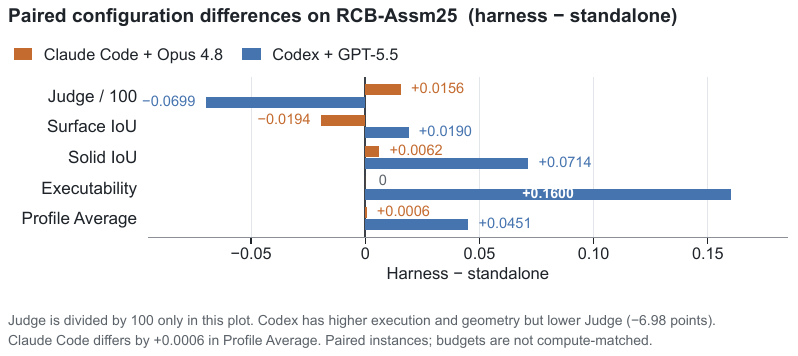}
\caption{Paired aggregate differences on RCB-Assm25 (agent minus standalone). Judge is divided by 100 for display.}
\label{fig:appendix_harness_delta}
\end{figure}

\subsection{Complete Part Track Component Results}
\label{sec:appendix_part_full}

Tables~\ref{tab:part_full_text}--\ref{tab:part_full_rendered} expand the compact summary in Table~\ref{tab:part_main}. Executability uses every assigned task. Solid IoU, Surface IoU, and Judge are reported on the tasks that provide the corresponding evaluator outputs, consistent with the reporting protocol in Section~\ref{sec:benchmark_construction}. Best and second-best values are highlighted only in the Exec. column.

\begin{table}[tbp]
\centering
\scriptsize
\caption{Complete Part Track component results for the text regime ($n=568$). Solid IoU, Surface IoU, and Judge are reported on the tasks that provide the corresponding evaluator outputs.}
\label{tab:part_full_text}
\setlength{\tabcolsep}{8pt}
\begin{tabular}{llrrrr}
\toprule
Model & Group & Exec. & \makecell{Solid IoU\\(cond.)} & \makecell{Surface IoU\\(cond.)} & \makecell{Judge\\(cond.)} \\
\midrule
Claude Opus 4.8~(\citealp{claudeopus48_2026})        & Frontier & 0.6356 & 0.4041 & 0.1066 & 63.49 \\
Gemini 3.1 Pro Preview~(\citealp{gemini31pro_2026}) & Frontier & \underline{0.8451} & 0.3995 & 0.1106 & 65.45 \\
GPT-5.4~(\citealp{gpt54_2026})                & Frontier & 0.6984 & 0.3643 & 0.0982 & 59.25 \\
GPT-5.5~(\citealp{gpt55_2026})                & Frontier & \textbf{0.9067} & 0.3509 & 0.0943 & 63.16 \\
Kimi K3~(\citealp{kimik3_2026})                & Frontier & 0.0581 & 0.4584 & 0.1685 & 73.26 \\
Doubao Seed 2.0 Pro~(\citealp{seed20_2026})    & Frontier & 0.2447 & 0.3451 & 0.0922 & 56.60 \\
\midrule
Qwen3-VL-8B~(\citealp{qwen3vl2025})            & Open 8B  & 0.0616 & 0.2575 & 0.0682 & 42.33 \\
Qwen3-VL-32B~(\citealp{qwen3vl2025})           & Open 32B & 0.1391 & 0.2507 & 0.0730 & 48.33 \\
Qwen3.8-27B~(\citealp{qwen38_2026})            & Open 27B & 0.7359 & 0.3600 & 0.0921 & 57.48 \\
\bottomrule
\end{tabular}
\end{table}

\begin{table}[tbp]
\centering
\scriptsize
\caption{Complete Part Track component results for the 2D engineering drawing regime ($n=236$). Solid IoU, Surface IoU, and Judge are reported on the tasks that provide the corresponding evaluator outputs.}
\label{tab:part_full_drawing}
\setlength{\tabcolsep}{8pt}
\begin{tabular}{llrrrr}
\toprule
Model & Group & Exec. & \makecell{Solid IoU\\(cond.)} & \makecell{Surface IoU\\(cond.)} & \makecell{Judge\\(cond.)} \\
\midrule
Claude Opus 4.8~(\citealp{claudeopus48_2026})        & Frontier & \textbf{0.9661} & 0.2811 & 0.1390 & 76.38 \\
Gemini 3.1 Pro Preview~(\citealp{gemini31pro_2026}) & Frontier & 0.8729 & 0.3378 & 0.1729 & 82.10 \\
GPT-5.4~(\citealp{gpt54_2026})                & Frontier & 0.7839 & 0.2044 & 0.0959 & 55.53 \\
GPT-5.5~(\citealp{gpt55_2026})                & Frontier & \underline{0.9237} & 0.2817 & 0.1400 & 79.75 \\
Kimi K3~(\citealp{kimik3_2026})                & Frontier & 0.1271 & 0.3325 & 0.1632 & 70.79 \\
Doubao Seed 2.0 Pro~(\citealp{seed20_2026})    & Frontier & 0.3915 & 0.2672 & 0.1274 & 58.78 \\
\midrule
Qwen3-VL-8B~(\citealp{qwen3vl2025})            & Open 8B  & 0.1314 & 0.1653 & 0.0739 & 23.99 \\
Qwen3-VL-32B~(\citealp{qwen3vl2025})           & Open 32B & 0.3220 & 0.1824 & 0.0956 & 37.23 \\
Qwen3.8-27B~(\citealp{qwen38_2026})            & Open 27B & 0.7627 & 0.1973 & 0.1052 & 56.21 \\
\bottomrule
\end{tabular}
\end{table}

\begin{table}[tbp]
\centering
\scriptsize
\caption{Complete Part Track component results for the real picture regime ($n=568$). Solid IoU, Surface IoU, and Judge are reported on the tasks that provide the corresponding evaluator outputs.}
\label{tab:part_full_real}
\setlength{\tabcolsep}{8pt}
\begin{tabular}{llrrrr}
\toprule
Model & Group & Exec. & \makecell{Solid IoU\\(cond.)} & \makecell{Surface IoU\\(cond.)} & \makecell{Judge\\(cond.)} \\
\midrule
Claude Opus 4.8~(\citealp{claudeopus48_2026})        & Frontier & \textbf{0.9806} & 0.3945 & 0.1026 & 59.16 \\
Gemini 3.1 Pro Preview~(\citealp{gemini31pro_2026}) & Frontier & 0.9085 & 0.4410 & 0.1238 & 66.41 \\
GPT-5.4~(\citealp{gpt54_2026})                & Frontier & 0.8539 & 0.3988 & 0.1155 & 58.38 \\
GPT-5.5~(\citealp{gpt55_2026})                & Frontier & \underline{0.9208} & 0.3754 & 0.1089 & 70.92 \\
Kimi K3~(\citealp{kimik3_2026})                & Frontier & 0.5722 & 0.4501 & 0.1218 & 65.40 \\
Doubao Seed 2.0 Pro~(\citealp{seed20_2026})    & Frontier & 0.5563 & 0.3826 & 0.1101 & 54.04 \\
\midrule
Qwen3-VL-8B~(\citealp{qwen3vl2025})            & Open 8B  & 0.2394 & 0.3311 & 0.0906 & 30.56 \\
Qwen3-VL-32B~(\citealp{qwen3vl2025})           & Open 32B & 0.4736 & 0.3120 & 0.0841 & 37.36 \\
Qwen3.8-27B~(\citealp{qwen38_2026})            & Open 27B & 0.8468 & 0.3802 & 0.0889 & 45.77 \\
\bottomrule
\end{tabular}
\end{table}

\begin{table}[tbp]
\centering
\scriptsize
\caption{Complete Part Track component results for the rendered image regime ($n=373$). Solid IoU, Surface IoU, and Judge are reported on the tasks that provide the corresponding evaluator outputs.}
\label{tab:part_full_rendered}
\setlength{\tabcolsep}{8pt}
\begin{tabular}{llrrrr}
\toprule
Model & Group & Exec. & \makecell{Solid IoU\\(cond.)} & \makecell{Surface IoU\\(cond.)} & \makecell{Judge\\(cond.)} \\
\midrule
Claude Opus 4.8~(\citealp{claudeopus48_2026})        & Frontier & \textbf{0.9864} & 0.5521 & 0.2250 & 74.23 \\
Gemini 3.1 Pro Preview~(\citealp{gemini31pro_2026}) & Frontier & 0.9035 & 0.5379 & 0.2240 & 79.37 \\
GPT-5.4~(\citealp{gpt54_2026})                & Frontier & 0.8660 & 0.5506 & 0.2254 & 71.60 \\
GPT-5.5~(\citealp{gpt55_2026})                & Frontier & \underline{0.9732} & 0.5234 & 0.2141 & 81.32 \\
Kimi K3~(\citealp{kimik3_2026})                & Frontier & 0.6113 & 0.5515 & 0.2091 & 79.45 \\
Doubao Seed 2.0 Pro~(\citealp{seed20_2026})    & Frontier & 0.5335 & 0.5118 & 0.2014 & 68.85 \\
\midrule
Qwen3-VL-8B~(\citealp{qwen3vl2025})            & Open 8B  & 0.4075 & 0.3794 & 0.1212 & 39.88 \\
Qwen3-VL-32B~(\citealp{qwen3vl2025})           & Open 32B & 0.3941 & 0.4090 & 0.1455 & 53.35 \\
Qwen3.8-27B~(\citealp{qwen38_2026})            & Open 27B & 0.8391 & 0.4964 & 0.1643 & 65.72 \\
\bottomrule
\end{tabular}
\end{table}

\subsection{Cross-Dataset Consistency Check for the Frozen Judge}
\label{sec:appendix_judge_reliability}

This section reports a consistency check for the frozen Kimi K2.6 Judge. On the public BenchCAD benchmark~\citep{benchcad2026}, which is separate from RealCADBench construction and Judge-prompt development, six fixed system outputs are scored by GPT-5.5 and by two independent runs each of Kimi K2.6 and Doubao. Every variant receives the same evidence and frozen prompt. Only the Judge identity or repeated query changes. This check does not use human ratings.

\begin{table}[tbp]
\centering
\caption{BenchCAD system-level Judge consistency check ($n=6$ systems). Panel (a) reports aggregate Judge scores with ranks in parentheses. Panel (b) reports cross-run and cross-model agreement. ``Pair'' is the number of concordant pairwise system orderings out of 15.}
\label{tab:benchcad_judge_reliability}
\scriptsize
\textbf{(a) Aggregate score and rank}\\[3pt]
\setlength{\tabcolsep}{3.5pt}
\begin{tabular}{@{}lccc@{}}
\toprule
System & GPT-5.5 & Kimi R1 & Kimi R2 \\
\midrule
Claude Opus 4.8~(\citealp{claudeopus48_2026})        & 71.26 (4) & 75.93 (4) & 75.45 (4) \\
Gemini 3.1 Pro Preview~(\citealp{gemini31pro_2026}) & 78.18 (2) & 77.25 (2) & 75.68 (3) \\
GPT-5.4~(\citealp{gpt54_2026})                & 70.42 (5) & 68.91 (6) & 70.93 (5) \\
GPT-5.5~(\citealp{gpt55_2026})                & 79.26 (1) & 78.12 (1) & 77.91 (1) \\
Kimi K3~(\citealp{kimik3_2026})                & 71.91 (3) & 76.78 (3) & 77.14 (2) \\
Doubao Seed 2.0 Pro~(\citealp{seed20_2026})    & 68.58 (6) & 68.98 (5) & 69.74 (6) \\
\bottomrule
\end{tabular}

\vspace{8pt}
\textbf{(b) Score and ranking agreement}\\[3pt]
\setlength{\tabcolsep}{3.0pt}
\begin{tabular}{@{}lccccc@{}}
\toprule
Comparison & Pearson & Spearman & Kendall & Pair & MAE \\
\midrule
Kimi R1 vs.\ R2    & .975 & .886 & .733 & 13/15 & 0.90 \\
GPT vs.\ Kimi R1   & .767 & .943 & .867 & 14/15 & 2.25 \\
GPT vs.\ Kimi R2   & .724 & .943 & .867 & 14/15 & 2.49 \\
Doubao R1 vs.\ R2  & .987 & 1.000 & 1.000 & 15/15 & 0.76 \\
\bottomrule
\end{tabular}
\end{table}

The two Kimi K2.6 runs differ by $0.90$ points on average (maximum $2.02$) and retain 13 of 15 pairwise orderings. Each also agrees with GPT-5.5 on 14 of 15 pairwise comparisons and selects the same top-ranked system. In this check, rank agreement is stronger than score agreement, and system order is more stable than absolute point values.

Repeatability alone does not determine the Judge choice. The nearly identical Doubao runs have only $\rho=0.60$ and $\tau=0.47$ rank agreement with GPT-5.5. In this six-system check, Kimi K2.6 combines repeatability with stronger cross-model convergence. We considered this evidence together with the sampled prompt-development loop and repeated CAD-expert review. We do not observe an obvious rank-level same-family preference: Kimi K2.6 ranks GPT-5.5 first in both runs and Kimi K3 third and second.

This six-system analysis tests the cross-dataset aggregate consistency of the chosen Judge identity. Together with the weak geometry--Judge correlation on RCB-Assm25, it supports reporting automated semantics as a separate column. The final rubric and prompts came from the sampled prompt-development loop with repeated CAD-expert review and preference screening. Expert post-checks provide an additional validation layer on evaluated outputs.

\subsection{Judge Prompts}
\label{sec:appendix_judge_prompts}

\subsubsection{Batch Visibility Pass}
\label{sec:appendix_batch_visibility}

This pass labels each BoM part as \texttt{full}, \texttt{partial}, or \texttt{invisible}.
The labels are evidence metadata. They do not change the subsequent scoring dimensions or weights.
The frozen evaluation adapter maps these labels to the Part prompt's legacy aliases \texttt{observed}, \texttt{mixed}, and \texttt{inferred}, respectively.

\begin{CJK*}{UTF8}{gbsn}
\begin{Verbatim}
你是 CAD 重建任务的 Batch Visibility Judge。一次性判断完整 BoM 中每个零件从
“原始全局参考图”获得了多少可直接约束三维建模的视觉证据。这里的可见性不是“能否认出或定位该零件”，而是“原
图是否充分显示其定义性几何，使不同建模者不会仅靠猜测得到明显不同的三维零件”。

你必须直接观察完整原图；BoM 只用于零件身份消歧，不得把 shape_desc/function 
当成图像证据。不要要求 crop、高亮、mask、grounding，也不要评价候选 CAD 
质量或装配质量。

严格区分以下概念：
- projected_visibility_fraction：目标在当前图像中的预计二维投影区域有多少未被
遮挡，范围 0..1。它只描述遮挡，不决定最终标签。
- reconstruction_observability：原图直接显示了多少定义性建模信息，范围 0..
1。综合考虑遮挡、拍摄角度、透视、分辨率、暗部、与相邻零件边界，以及主轮廓、厚度/截面线索、关键孔/凸台/接
口面的可见程度。最终 visibility 必须由它决定。
- localization_confidence：你是否找对了零件。定位置信度高不等于可观测性高。
- reference_evidence_strength：已可见区域的像素证据是否清晰可靠；它描述清晰度/
可信度，不描述覆盖面。

visibility 的统一定义（它只描述参考证据来源，不改变后续评分维度或权重）：
- full / observed：reconstruction_observability >= 0.65
 且 defining_features_sufficient=true。主包络、主要轴线/弯折、主要截面线
索及关键接口的位置与类型已经足以支持主体建模；剩余未知主要是背面、小孔、倒角等次要隐藏细节。不要因为单视图天
然看不到全部背面就降为 partial。
- partial / mixed：0.10 <= reconstruction_observability
 < 0.65，或 defining_features_sufficient=false。可以可靠定位并比较
一部分结构，但至少一个会明显改变主体建模结果的主要形状、截面、规格或关键接口仍有歧义。
- invisible：reconstruction_observability < 
0.10。无法可靠定位，或没有足够直接像素支持任何有意义的几何比较。

硬性判定规则：
1. “知道这是什么零件”或“看见完整外接轮廓”不能单独推出 full。
2. 
只看到单一圆盘侧影、单一外轮廓或通用类别形状，并且端面/厚度/主体规格的未知会改变主要建模决策时，最高为 
partial；仅有次要背面孔位未知不阻止 full。
3. 对法兰、关节盖、连接器等接口主导零件，仅当不可见的接口信息会改变其主要功能拓扑或主体规格时才判 
partial；不要求生产图级的完整孔阵列证据。
4. 与相邻零件边界混叠、主要部分处于暗部或分辨率不足以区分定义性特征时，必须降低 
reconstruction_observability；不得用工业常识、BoM 描述或标准件先验补全。
5. 如果你能列出一个会显著改变建模结果的关键未知项，通常 
defining_features_sufficient=false。

ambiguity_flags 只使用必要的短标签，可选值包括：occlusion、single_view_
depth、interface_face_unobservable、low_resolution、dark_
region、part_boundary_ambiguous。evidence 只写一句“直接可见依据 + 
最关键未知项”，不要把原图压缩成长篇文字描述。

必须覆盖 BoM 的每个 part_id，恰好一次，不得遗漏或添加。只输出 JSON，不要 
Markdown：
{
  "parts": [
    {
      "part_id": 1,
      "visibility": "full|partial|invisible",
      "projected_visibility_fraction": 0.0,
      "reconstruction_observability": 0.0,
      "defining_features_sufficient": false,
      "localization_confidence": "high|medium|low",
      "reference_evidence_strength": 0.0,
      "ambiguity_flags": 
["interface_face_unobservable"],
      "evidence": "一句简短的直接可见依据和关键未知项"
    }
  ]
}
\end{Verbatim}
\end{CJK*}

\subsubsection{Part Judge}
\label{sec:appendix_part_judge}

The Part Judge scores only the target part.
Pose, contact, collision, mating relations, and global scale are out of scope.
Mesh integrity (\(P_3\)) is produced by a programmatic audit as an internal Part-Judge rubric component. It stays inside that rubric.

\begin{CJK*}{UTF8}{gbsn}
\begin{Verbatim}
你是 CAD 重建质量的 Part Judge。你必须直接观察“原始全局参考图”定位 
target_part；BoM 只用于消歧和理解零件身份，不得用 BoM 
文字替代图像证据。候选零件六视图是同一候选网格的衍生证据（面着色 z-buffer + 
超采样；无三角描边/轮廓描边伪纹）。

边界：只评价目标零件自身；严禁评价它在装配体中的位姿、接触距离、碰撞、装配关系或全局比例。不要要求裁剪、高亮
、mask 或 grounding。原图不可见的隐藏面不能被判为“不还原”。

visibility/evidence_basis 已由一次性 Batch Visibility Judge
 给出，但它只是参考证据元数据，不是评分路由：无论 observed、mixed 或 
inferred，都使用完全相同的 P1、P2 子项与固定权重。你仍须直接查看完整原图；不得把 batch 
元数据改写成一段替代图像观察的描述。若标签明显矛盾，可报告 
visibility_dispute，但不要自行改变评分任务。

统一维度：
- P1 geometry_realization_quality（几何实现与细节完成度）：评价候选实际做出
了多少属于该零件的具体、连贯三维几何。observed 时以原图为准；mixed 
时可见按原图、不可见按实现完整性；inferred 
时不声称忠于真值。简单零件必要几何完整即可高分，禁止无意义复杂度换分。
- P2 functional_design_quality（功能设计质量）：不做外观还原比较，只评价功能拓
扑、接口与受力/运动逻辑。不得把 P1 的“像不像”重复计入 P2。
- P3 artifact_integrity：由 mesh audit 处理；你不输出 P3。

你不直接给总分。每个子项输出连续 score（0–100，可用一位小数），代码只做固定加权。禁止习惯性取整到
 0/5/10 的倍数。量尺按“对本维语义的可观测偏差有多大”连续内插（非特征清单）：
- 95–100：本维无实质可见偏差，或仅有不可辨噪声。
- 80–90：有清楚但局部的偏差，本维主目标仍成立。
- 60–75：本维主目标有一眼可见偏离，但仍可辨为同类实现。
- 35–55：本维主目标严重偏离或关键缺失。
- 0–25：本维基本未实现或无关几何。

维内裁决（强制；保持泛化，禁止特征/失败模式清单化）：
1. 子项只定义问题类型；不要用预置零件特征菜单或固定失败模式清单。
2. 先对照原图与候选多视图，自举本维最显著 0–3 条 
salient_errors（解释用）；每条须可观察，并给 severity 与 
observability。
3. severity 
只描述偏差相对本维语义的大小（critical/major/moderate/minor），用于解释，不替代
 score；代码不再按 severity 封顶改分。
4. observability：high=直接可见；medium=需结合两处证据；low=推断较多。低可观
测差异只能温和影响 score，不得仅凭 low 证据把该维拉到“主目标严重偏离”档。
5. 先写 salient_errors，再给 score；score 
应与你所描述的偏差幅度成比例，允许段内连续取值。
6. 不同零件关键差距不同；凡属本维语义且显著均可进入 errors。禁止凑分枚举无关项。

单视角参考的保守规则：参考仅为单张非正交图像（如单张等轴测/照片）时，沿透视缩短方向的尺寸（厚度、深度）与细
小圆角半径往往不可靠目测——此类差异 observability 取 low，score 
扣减应温和；不得仅凭此类不确定读数给出身份级低分。

P1 子项（问题类型，非特征清单）：
- primary_form（35%）：主包络、主要比例、轴线/弯折、截面、主体拓扑相对参考的一致性。
- defining_features（30%）：确立身份、区别于通用图元的关键可见特征是否正确可信；何为身
份关键由参考图决定。
- detail_fidelity_and_completeness（25%）：次级/局部几何的具体准确完整
程度；空壳低分。细节账本仅佐证。
- surface_refinement（10%）：连续表面质量。以形体/台阶/孔洞边界为准；材质、棋盘背景
与离散化观感应忽略，除非多视图一致且与参考明显矛盾。

detail：observed 以图像为准；mixed 分可见/不可见；inferred 
评实现是否充分克制。简单零件可 intrinsically_simple=true。

P2 子项（不做外观重复计分）：
- functional_topology（40%）：主要功能路径是否成立。
- interface_design（35%）：自身连接/安装/工作界面是否明确可用。
- load_motion_logic（25%）：基本受力/运动逻辑是否合理。

critical_findings 记录实质问题；severity 用 
major|moderate|minor。每条只属 P1 或 P2。

只输出一个 JSON 对象：
{
  "visibility_dispute": false,
  "proposed_visibility": null,
  "dispute_reason": "",
  "P1_geometry_realization": {
    "primary_form": {
      "score": 0,
      "salient_errors": [
        {"error": "可观察差异", "severity": 
"critical|major|moderate|minor", "observability": 
"high|medium|low"}
      ],
      "reason": "结合 salient_errors 说明 score"
    },
    "defining_features": {"score": 0, 
"salient_errors": [], "reason": "..."},
    "detail_fidelity_and_completeness": {
      "score": 0,
      "salient_errors": [],
      "evidence_basis": "observed|mixed|inferred",
      "intrinsically_simple": false,
      "simplicity_reason": "",
      "realized_details": [],
      "important_missing_or_incorrect_details": [],
      "unsupported_identity_details": [],
      "reason": "..."
    },
    "surface_refinement": {"score": 0, 
"salient_errors": [], "reason": "..."}
  },
  "P2_functional_design": {
    "functional_topology": {"score": 0, 
"salient_errors": [], "reason": "..."},
    "interface_design": {"score": 0, "salient_errors":
 [], "reason": "..."},
    "load_motion_logic": {"score": 0, 
"salient_errors": [], "reason": "..."}
  },
  "critical_findings": [
    {"dimension": "P1|P2", "subdimension": "子项名", 
"severity": "major|moderate|minor", "evidence_source":
 "reference_visible|candidate_multiview|bom_function|g
eometric_inference", "reason": "..."}
  ],
  "summary": "简洁、可核验",
  "strengths": ["..."],
  "issues": ["..."],
  "visual_evidence": ["..."]
}
\end{Verbatim}
\end{CJK*}

\subsubsection{Assembly Judge}
\label{sec:appendix_assembly_judge}

The Assembly Judge scores the delivered assembly independently of Part-Judge outputs along component geometry quality \(Q\), assembly accuracy \(F\), and system design quality \(D\).
The prompt requests \texttt{score} on a continuous $0$--$100$ scale and also accepts the legacy field \texttt{level} on a $0$--$10$ scale. The frozen parser normalizes either representation to $0$--$100$ before applying the released fixed aggregation.

\begin{CJK*}{UTF8}{gbsn}
\begin{Verbatim}
你是 CAD 重建质量的 Assembly Judge。你只做装配体全局级评价：直接观察完整原始参考图、完整
 BoM、候选装配体六视图，并结合程序提供的资产与几何关系事实。你绝不读取、推断、汇总或复述任何 Part 
Judge 分数；本请求也不会提供 Part Judge 结果。

你只标注三个维度 Q/F/D：
- Q component_geometry_quality：全部 BoM 
零件作为一个集合的几何实现与还原质量。它在全局层级承担与 Part 几何评分相同的功能，可在业务中与 
Part 聚合结果二选一，但本次必须直接从全局输入独立判断。
- F assembly_accuracy：候选零件是否以正确的姿态、比例、相对位置、配合关系和空间关系组成
了参考装配体。
- D system_design_quality：当前实际交付的装配体是否形成合理、完整、可工作的系统级功
能与工程逻辑。

资产可用性、配合距离和碰撞代理只是审计事实，不是额外评分维度或门控。你必须把事实的实际后果归入 
Q/F/D，代码只做 Q/F/D 的固定加权，不再施加 C/V 惩罚。

统一量尺：所有子项输出连续 score（0–100；也兼容旧字段 level 
0..10），代码只做固定加权。按对本维语义的可观测偏差连续内插：
- 95–100：无实质可见偏差。
- 80–90：局部清楚偏差，主目标仍成立。
- 60–75：主目标一眼可见偏离，仍可辨。
- 35–55：主目标严重偏离或关键缺失。
- 0–25：基本未实现或无关。

Q 的强制边界（35/30/25/10）：
1. Q 覆盖全部 BoM 零件，不排除 missing/fallback。fallback 
是没有实现目标零件几何的替代物；missing 是零实现。它们必须依据零件角色、数量和显著性拉低 
Q，而不是被平均范围排除。
2. Q 只评价零件自身几何。忽略当前装配位姿、零件间 
gap/碰撞/悬浮、主链连通和系统功能链；这些分别属于 F 或 D。
3. 原图可见内容必须直接比较，不得先压缩成文字特征再判断。不可见部分只判断实现是否具体、完整、符合 BoM
 身份与合理工程理解，不声称知道不可见真值。
4. primary_form（35%）：全体零件主包络、比例、轴线/弯折、截面和主体拓扑（问题类型，非特征
清单）。
5. defining_features（30%）：确立各零件身份的关键可见特征是否正确；由原图决定何为身份
关键，不预设特征类别。
6. detail_fidelity_and_completeness（25%）：次级/局部几何相对原图或合
理理解的实现程度。维内先自举显著误差再定级；“简单但自洽”不等于细节优秀，空壳粗模应低档。细节账本仅作佐证。
7. 
surface_refinement（10%）：连续表面质量与过渡；忽略颜色、材质、棋盘背景与三角伪纹。
8. 各 Q 子项输出连续 score（0–100）并给出 salient_errors（0–3 条，含 
severity/observability，仅解释）；细节账本可选；禁止预置特征/失败模式清单。代码不按 
severity 封顶。

F 的强制边界（35/25/30/10）：
1. F 评价“这些候选零件是否被准确装成参考对象”，不评价零件内部细节或系统功能价值。
2. global_pose_and_silhouette（35%）：整体轮廓、主链姿态、弯折走势和关键方向
相对原图的准确度。
3. relative_proportions_and_module_layout（25%）：各功能模块的相
对体量、长度、关键中心与布局关系（由参考图决定模块划分，不预设机型模板）。
4. mating_and_connectivity（30%）：应相配的零件是否实际贴合、对轴、连续并形成预
期装配链。预期配合面的 gap、错轴、悬浮和断链只在此项评价。程序关系事实中的 unavailable 
边表示该关系未实现，必须计入；距离是近似证据，应与候选视图共同判断。
5. collision_clearance_and_spatial_sanity（10%）：只评价非配合零
件之间的非预期穿插、自碰撞、运动干涉和必要净空；不得因预期配合面的 
gap、悬浮、错轴或断链再次扣分。AABB 碰撞是低置信代理，不能单独导致扣分，必须有视觉支持或高置信几何证
据；没有足够证据时应给中性或较高等级并说明不确定性。
6. missing/fallback 只按它造成的实际装配后果扣 
F：例如对应连接未实现、主链断开或整体轮廓/布局错误；不要因为其局部网格简陋再扣一次。

D 的强制边界（35/25/25/15）：
1. D 
评价当前实际交付物的系统实现，不使用“假设缺件都存在、接口都贴合”的反事实，也不比较原图外观相似度。
2. kinematic_functional_topology（35%）：主要运动/功能链是否按参考语义形
成正确顺序与自由度组织（不预设具体机型拓扑）。
3. functional_module_completeness（25%）：完成目标系统功能所需的关键模块
是否实际存在并承担其角色。关键模块 missing/fallback 会直接降低该项；非关键装饰件影响应轻。
4. load_support_logic（25%）：实际承力路径、支撑层级与运动净空是否成立。
5. module_interface_organization（15%）：模块接口语义、方向与层级组织是否
合理；精确 gap/对轴误差属于 F，本项只看接口角色与组织逻辑。
6. 零件表面粗糙、孔槽等局部细节只属于 Q；单纯“不像原图”只属于 Q/F；不要用这些理由扣 D。

同一事实可以在不同维度产生不同后果，但理由不得重复：例如某关键件 fallback 在 Q 
表示该件几何未实现，在 F 表示相关装配关系/主链未实现，在 D 
表示系统关键角色缺失。三项必须分别描述对应语义，不能把同一句“有 fallback”复制三遍。

critical_findings 只记录会实质拉低 Q 子项的可核验问题。major 
改变一个关键零件或多个零件的主要身份/形体；moderate 为重要但不改变整体身份；minor 
为局部问题。dimension 固定写 Q。

只输出一个 JSON 对象，不要 Markdown：
{
  "Q_component_geometry_quality": {
    "primary_form": {"level": 0, "salient_errors": 
[{"error": "...", "severity": "moderate", 
"observability": "high"}], "reason": "结合 
salient_errors"},
    "defining_features": {"level": 0, 
"salient_errors": [], "reason": "结合 salient_errors"},
    "detail_fidelity_and_completeness": {
      "level": 0,
      "salient_errors": [],
      "evidence_basis": "global_mixed",
      "intrinsically_simple": false,
      "simplicity_reason": "",
      "realized_details": ["可选佐证"],
      "important_missing_or_incorrect_details": 
["可选佐证"],
      "unsupported_identity_details": ["可选佐证"],
      "reason": "结合 salient_errors 与可选账本"
    },
    "surface_refinement": {"level": 0, 
"salient_errors": [], "reason": "结合 salient_errors"},
    "critical_findings": [
      {"dimension": "Q", "subdimension": "子项名", 
"severity": "major|moderate|minor", "reason": "可核验问题"}
    ],
    "reason": "Q 的全局简要理由"
  },
  "F_assembly_accuracy": {
    "global_pose_and_silhouette": {"level": 0, 
"reason": "原图与候选的直接比较"},
    "relative_proportions_and_module_layout": 
{"level": 0, "reason": "原图与候选的直接比较"},
    "mating_and_connectivity": {"level": 0, "reason": 
"关系事实与候选视图证据"},
    "collision_clearance_and_spatial_sanity": 
{"level": 0, "reason": "空间证据及其置信度"},
    "reason": "F 的全局简要理由"
  },
  "D_system_design_quality": {
    "kinematic_functional_topology": {"level": 0, 
"reason": "系统设计证据"},
    "functional_module_completeness": {"level": 0, 
"reason": "实际功能模块证据"},
    "load_support_logic": {"level": 0, "reason": 
"系统设计证据"},
    "module_interface_organization": {"level": 0, 
"reason": "系统设计证据"},
    "reason": "D 的全局简要理由"
  },
  "issue_effects": [
    {"fact": "一个关键事实", "Q_effect": "仅几何后果或不适用", 
"F_effect": "仅装配后果或不适用", "D_effect": "仅系统功能后果或不适用"}
  ],
  "summary": "同时概括 Q/F/D，明确区分几何、装配准确性和系统实现",
  "issues": ["按 Q/F/D 标注归属的问题"],
  "visual_evidence": ["原图和候选全局多视图中的可核验依据"]
}
\end{Verbatim}
\end{CJK*}

\subsection{Instance-Level Agent Traces on RCB-Assm25}
\label{sec:appendix_harness_assm}
\label{sec:appendix_codex_assm}

Table~\ref{tab:codex_assm_instances} reproduces a diagnostic sheet from the frozen RCB-Assm25 review workbook. It is separate from the official Judge column in Table~\ref{tab:assembly_main}.
Each row shows the real picture input together with ISO renders from Codex + GPT-5.5 and Claude Code + Claude Opus 4.8, the workbook's Assembly-Judge dimension traces \((Q,F,D)\) (part geometry, layout, and system design), wall-clock time, and tokens.
Each column follows the corresponding agent configuration.

The workbook total is $0.40Q+0.35F+0.25D$, computed from unrounded values. The table displays each dimension to one decimal. The sheet-level mean totals are $67.2$ for Codex + GPT-5.5 and $56.1$ for Claude Code + Claude Opus 4.8 when averaged over the $24$ exported assemblies in each column. Codex exports 24 assemblies, with \texttt{rcb\_000300010} as the non-export case, and Claude Code also exports 24 assemblies, with \texttt{rcb\_000633471} as the non-export case. Among the $23$ assemblies scored by both, Codex has the higher total on $18$ and Claude Code on $5$, including reversals larger than $30$ points: \texttt{rcb\_000575344} ($20.1$ vs.\ $75.1$), \texttt{rcb\_000398996} ($45.2$ vs.\ $85.9$), and \texttt{rcb\_000052132} ($30.6$ vs.\ $65.0$). Mean wall-clock time is $12.1$\,min for Codex and $31.6$\,min for Claude Code, and mean token use is $24.5$k and $47.1$k, respectively. For Claude Code, the two rows without recorded time or token values (\texttt{rcb\_000352539}, \texttt{rcb\_000514100}) are excluded from those means. Layout \(F\) exceeds part geometry \(Q\) for both systems, with mean gaps of $7.9$ and $13.8$ points.

The traces provide instance-level detail for the patterns summarized in Table~\ref{tab:assembly_main}. They also show that \(Q\), \(F\), and \(D\) need not move together.

\setlength{\tabcolsep}{4pt}
\setlength{\LTcapwidth}{\textwidth}
\renewcommand{\arraystretch}{0.95}
\begin{longtable}{@{}>{\raggedright\arraybackslash}p{2.05cm} >{\centering\arraybackslash}p{2.35cm} >{\centering\arraybackslash}p{3.35cm} >{\centering\arraybackslash}p{3.35cm}@{}}
\caption{Per-instance diagnostic traces on RCB-Assm25 for Codex + GPT-5.5 and Claude Code + Claude Opus 4.8. Input is the real picture assembly task. Each agent column shows the ISO render of the executed FreeCAD program, when an artifact was exported, together with the auxiliary \((Q,F,D)\) tuple, time (min), and tokens (k). Missing time or token records are marked ``---''. These tuples are diagnostic traces and are separate from the official Judge scores in Table~\ref{tab:assembly_main}.}
\label{tab:codex_assm_instances}
\\
\toprule
Case & Input & Codex + GPT-5.5 & Claude Code + Opus 4.8 \\
\midrule
\endfirsthead
\multicolumn{4}{c}{\tablename\ \thetable\ (continued)}\\
\toprule
Case & Input & Codex + GPT-5.5 & Claude Code + Opus 4.8 \\
\midrule
\endhead
\midrule
\multicolumn{4}{r}{\footnotesize Continued on next page.}\\
\endfoot
\bottomrule
\endlastfoot
\texttt{\scriptsize rcb\_000052132} &
  \includegraphics[width=0.95\linewidth,height=1.48cm,keepaspectratio]{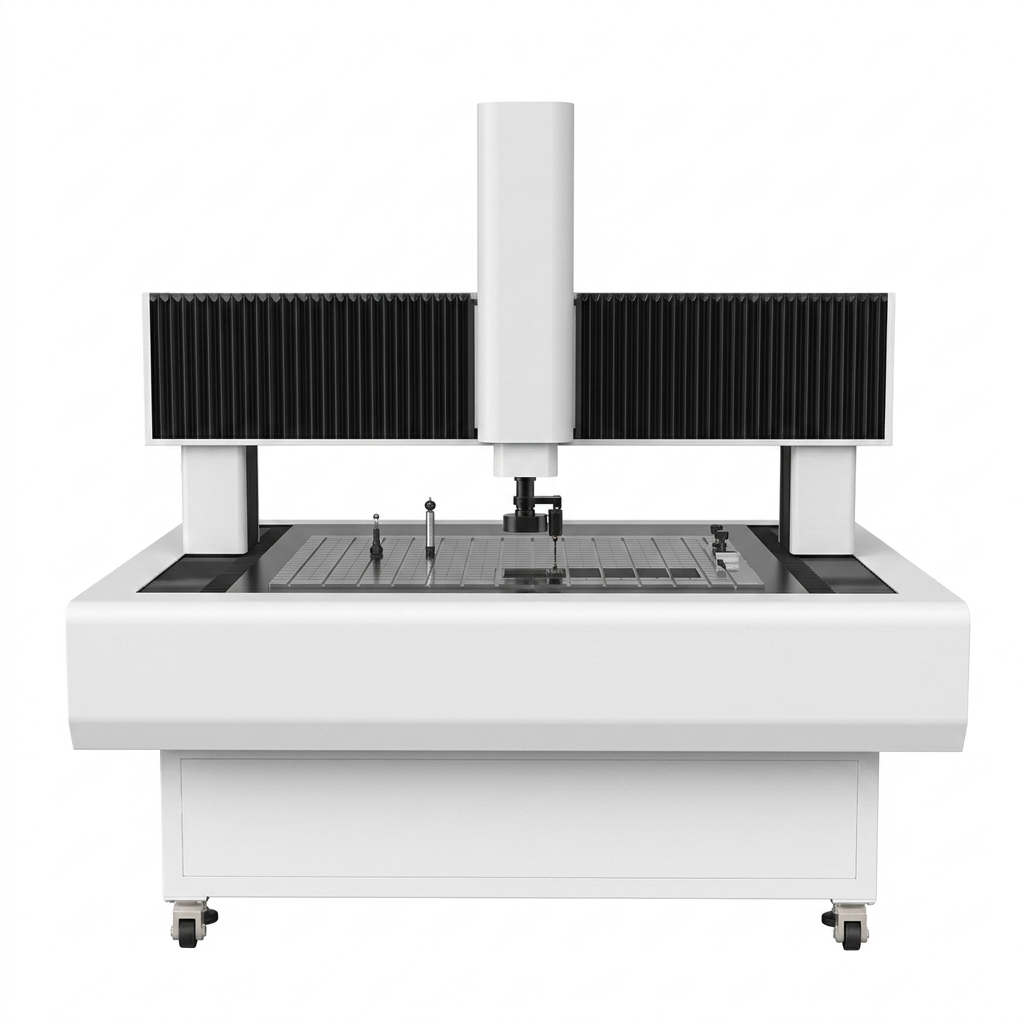} &
  \begin{tabular}[t]{@{}c@{}}
    \includegraphics[width=0.95\linewidth,height=1.48cm,keepaspectratio]{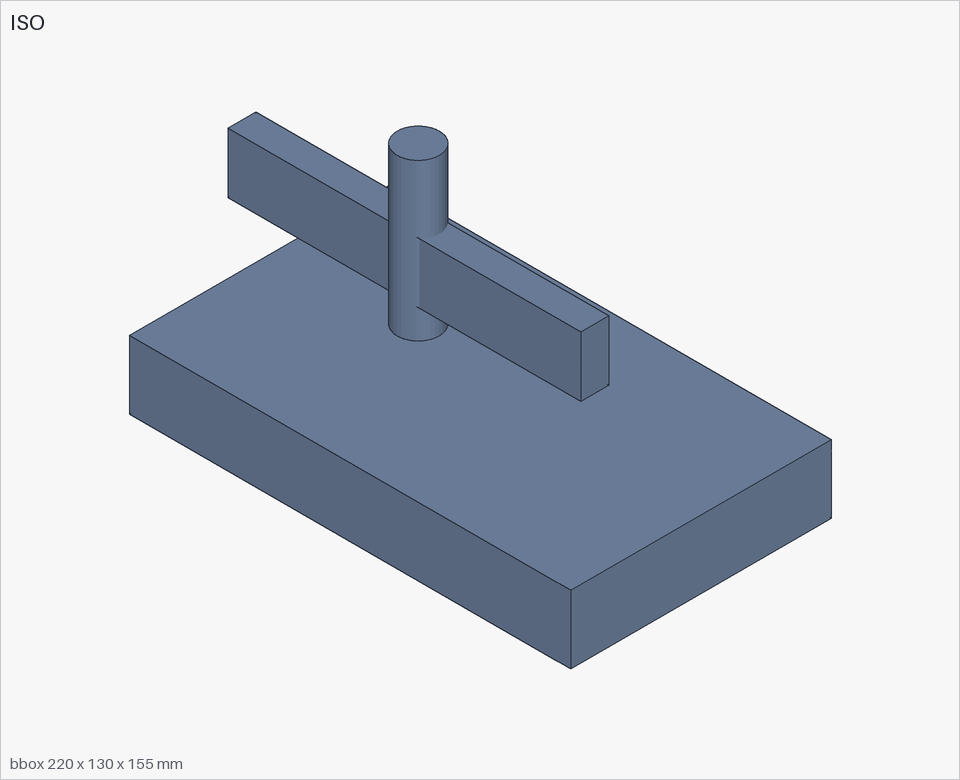} \\
    {\tiny\ttfamily\makecell[c]{Q 28.6 · F 34.6 · D 28.0 \\ 13.8 min · 23.9k}}
  \end{tabular} &
  \begin{tabular}[t]{@{}c@{}}
    \includegraphics[width=0.95\linewidth,height=1.48cm,keepaspectratio]{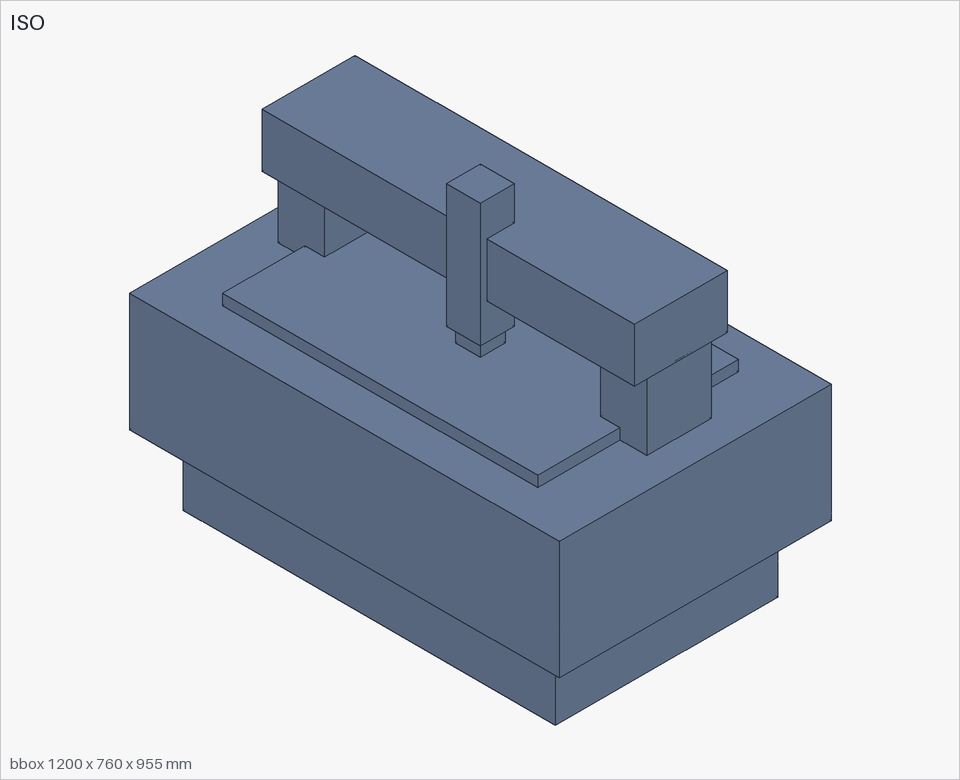} \\
    {\tiny\ttfamily\makecell[c]{Q 49.4 · F 78.0 · D 71.8 \\ 59.4 min · 57.8k}}
  \end{tabular} \\
\addlinespace[1pt]
\texttt{\scriptsize rcb\_000112974} &
  \includegraphics[width=0.95\linewidth,height=1.48cm,keepaspectratio]{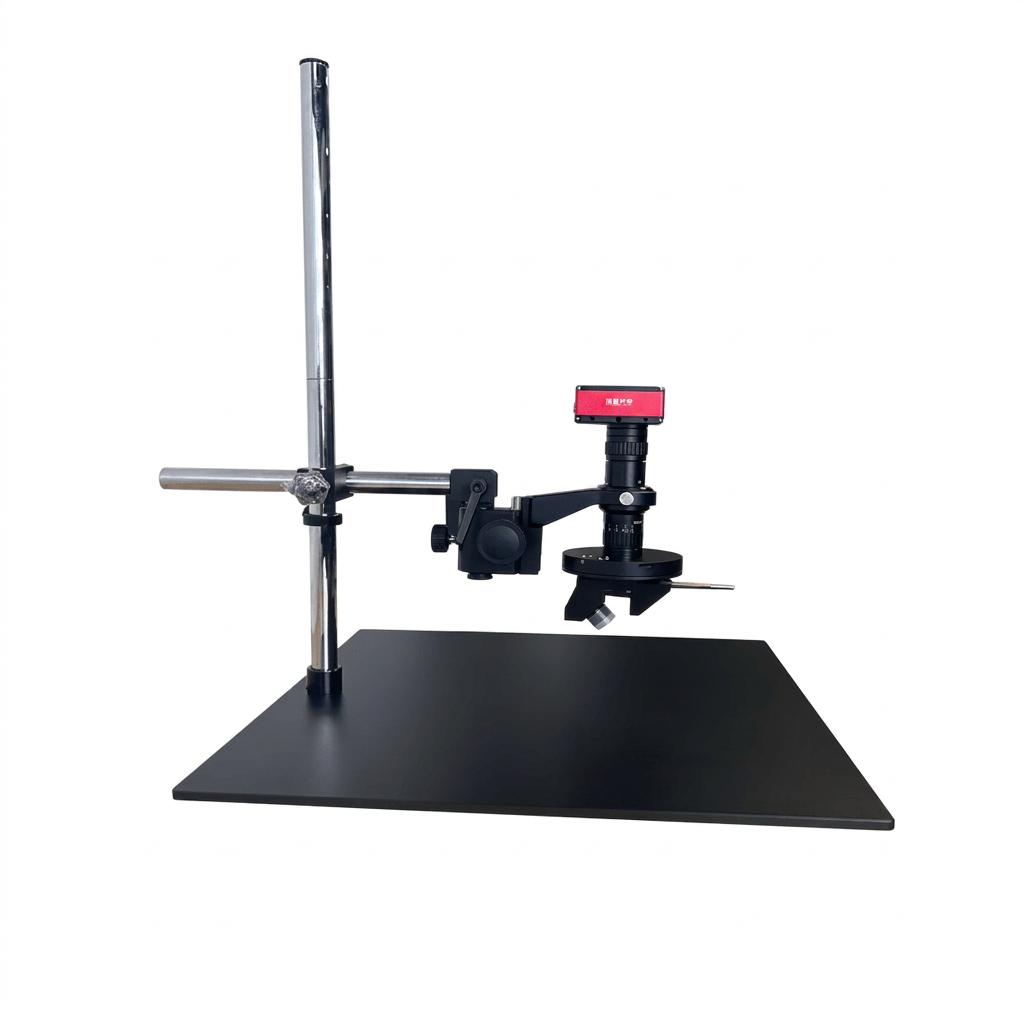} &
  \begin{tabular}[t]{@{}c@{}}
    \includegraphics[width=0.95\linewidth,height=1.48cm,keepaspectratio]{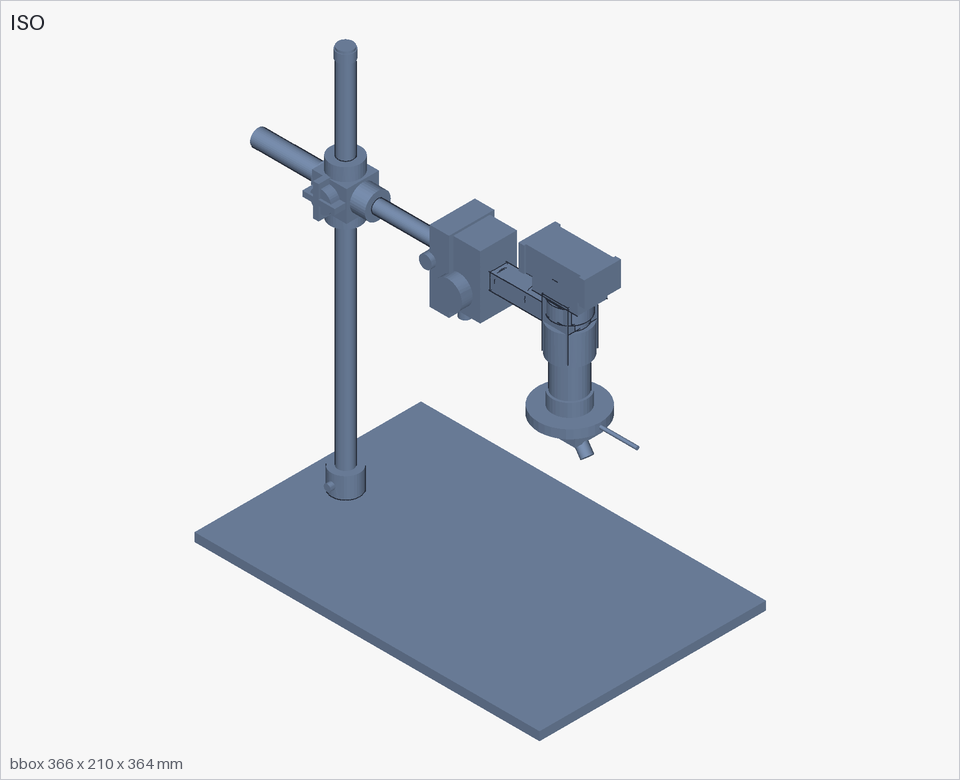} \\
    {\tiny\ttfamily\makecell[c]{Q 80.6 · F 86.3 · D 83.8 \\ 10.3 min · 20.9k}}
  \end{tabular} &
  \begin{tabular}[t]{@{}c@{}}
    \includegraphics[width=0.95\linewidth,height=1.48cm,keepaspectratio]{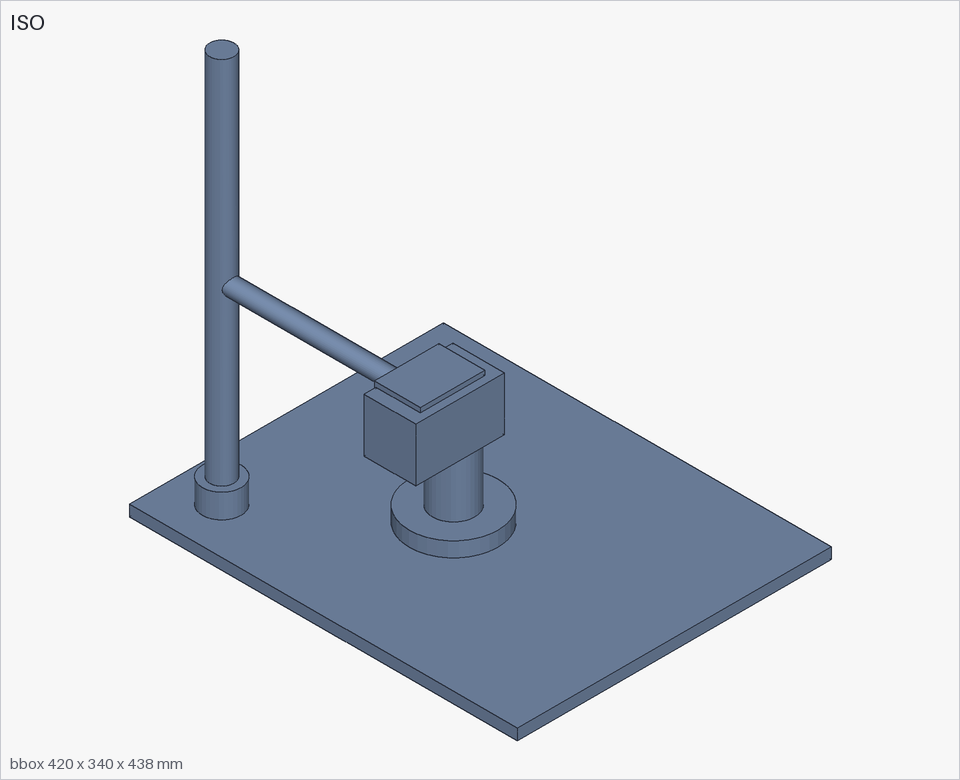} \\
    {\tiny\ttfamily\makecell[c]{Q 45.9 · F 70.7 · D 59.0 \\ 15.0 min · 21.5k}}
  \end{tabular} \\
\addlinespace[1pt]
\texttt{\scriptsize rcb\_000133697} &
  \includegraphics[width=0.95\linewidth,height=1.48cm,keepaspectratio]{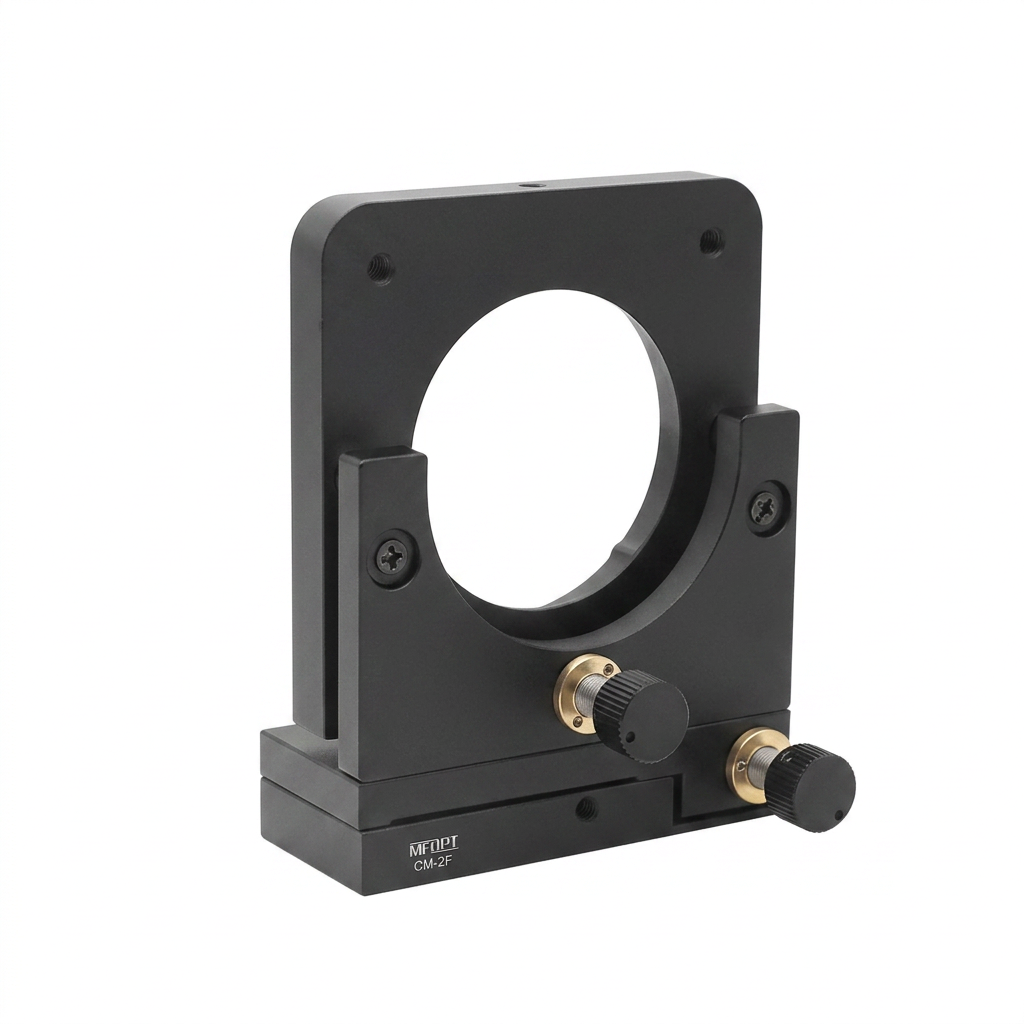} &
  \begin{tabular}[t]{@{}c@{}}
    \includegraphics[width=0.95\linewidth,height=1.48cm,keepaspectratio]{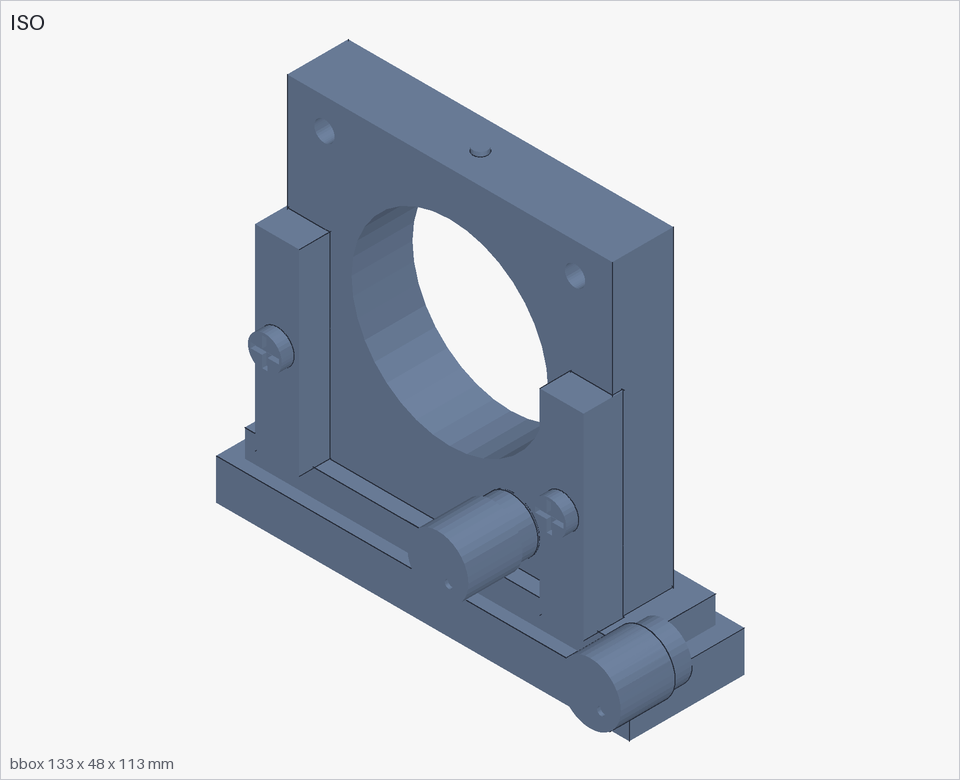} \\
    {\tiny\ttfamily\makecell[c]{Q 71.9 · F 81.6 · D 79.6 \\ 13.6 min · 25.2k}}
  \end{tabular} &
  \begin{tabular}[t]{@{}c@{}}
    \includegraphics[width=0.95\linewidth,height=1.48cm,keepaspectratio]{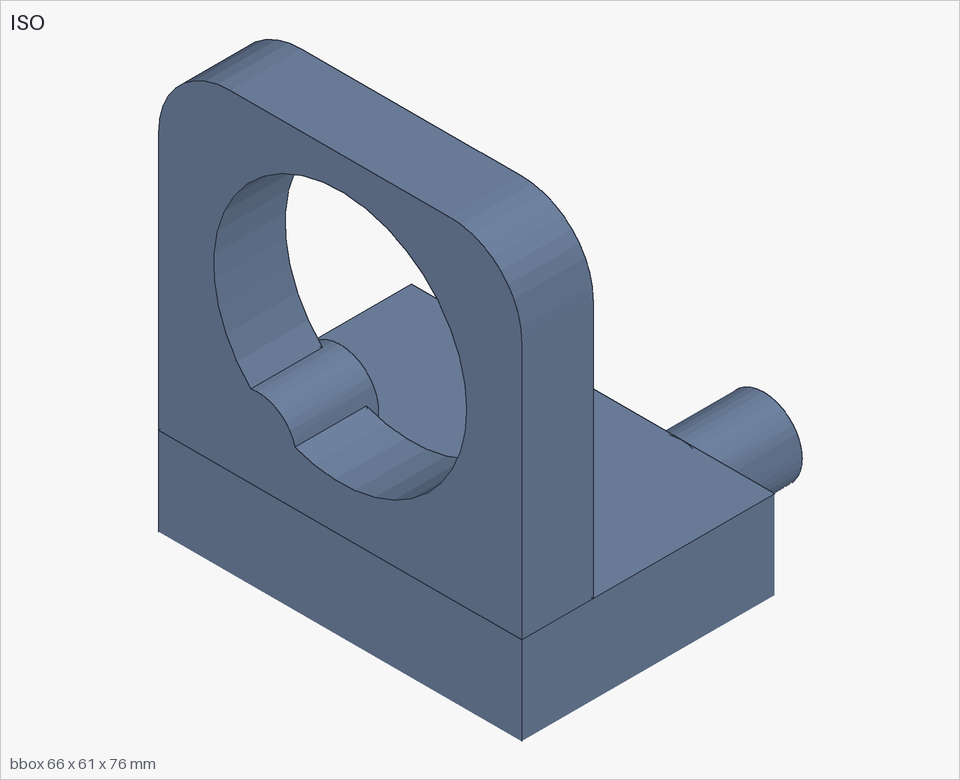} \\
    {\tiny\ttfamily\makecell[c]{Q 47.6 · F 64.2 · D 51.7 \\ 47.2 min · 58.5k}}
  \end{tabular} \\
\addlinespace[1pt]
\texttt{\scriptsize rcb\_000198365} &
  \includegraphics[width=0.95\linewidth,height=1.48cm,keepaspectratio]{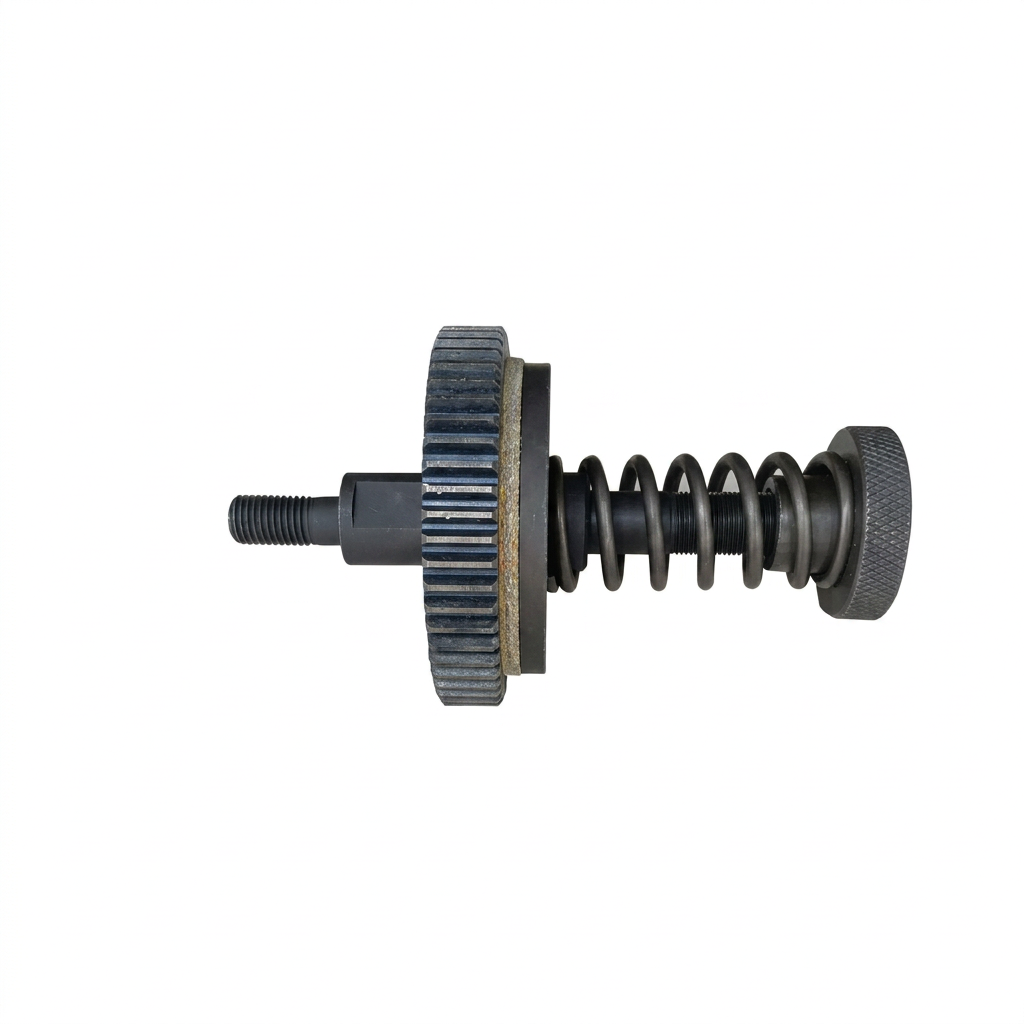} &
  \begin{tabular}[t]{@{}c@{}}
    \includegraphics[width=0.95\linewidth,height=1.48cm,keepaspectratio]{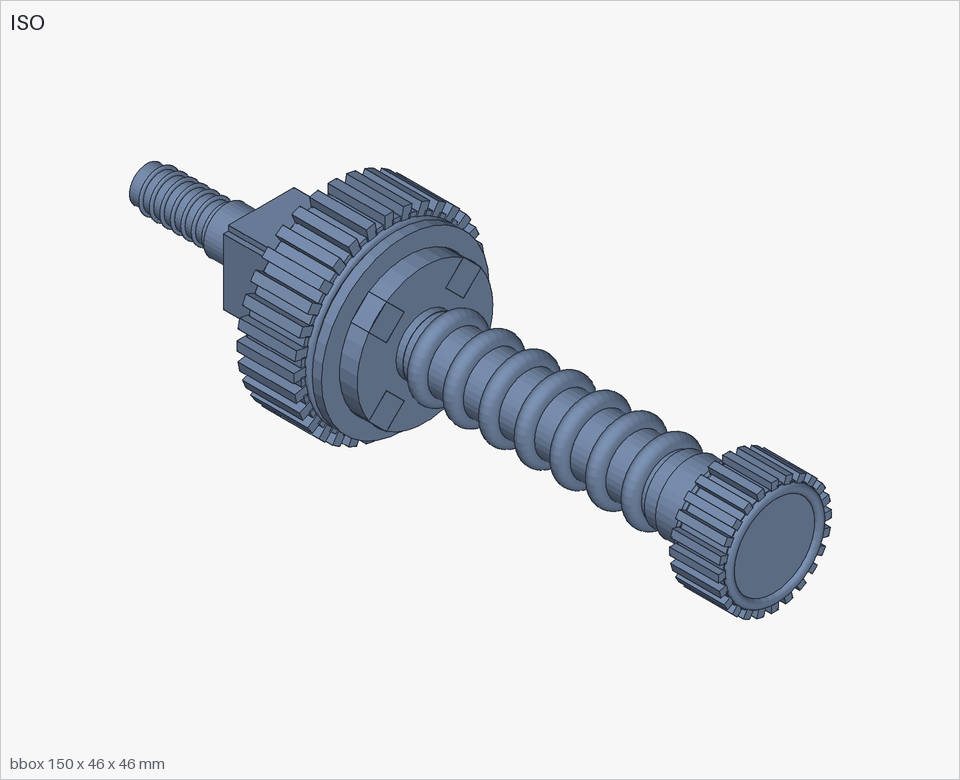} \\
    {\tiny\ttfamily\makecell[c]{Q 87.0 · F 91.0 · D 88.5 \\ 13.6 min · 20.6k}}
  \end{tabular} &
  \begin{tabular}[t]{@{}c@{}}
    \includegraphics[width=0.95\linewidth,height=1.48cm,keepaspectratio]{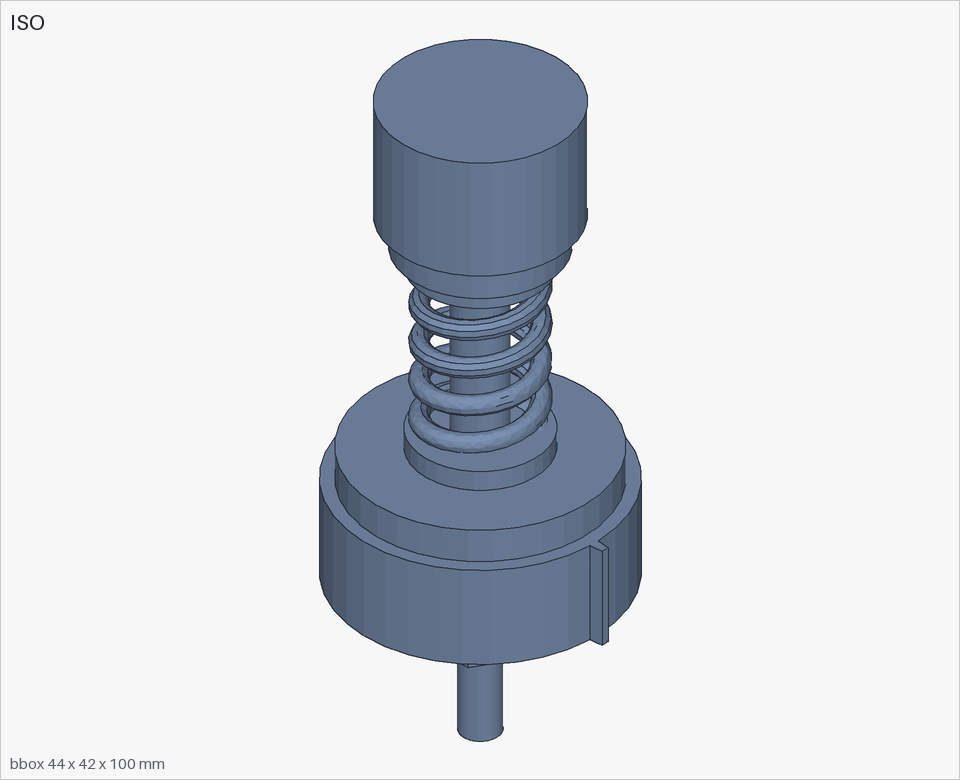} \\
    {\tiny\ttfamily\makecell[c]{Q 65.0 · F 68.5 · D 71.9 \\ 42.7 min · 41.5k}}
  \end{tabular} \\
\addlinespace[1pt]
\texttt{\scriptsize rcb\_000300010} &
  \includegraphics[width=0.95\linewidth,height=1.48cm,keepaspectratio]{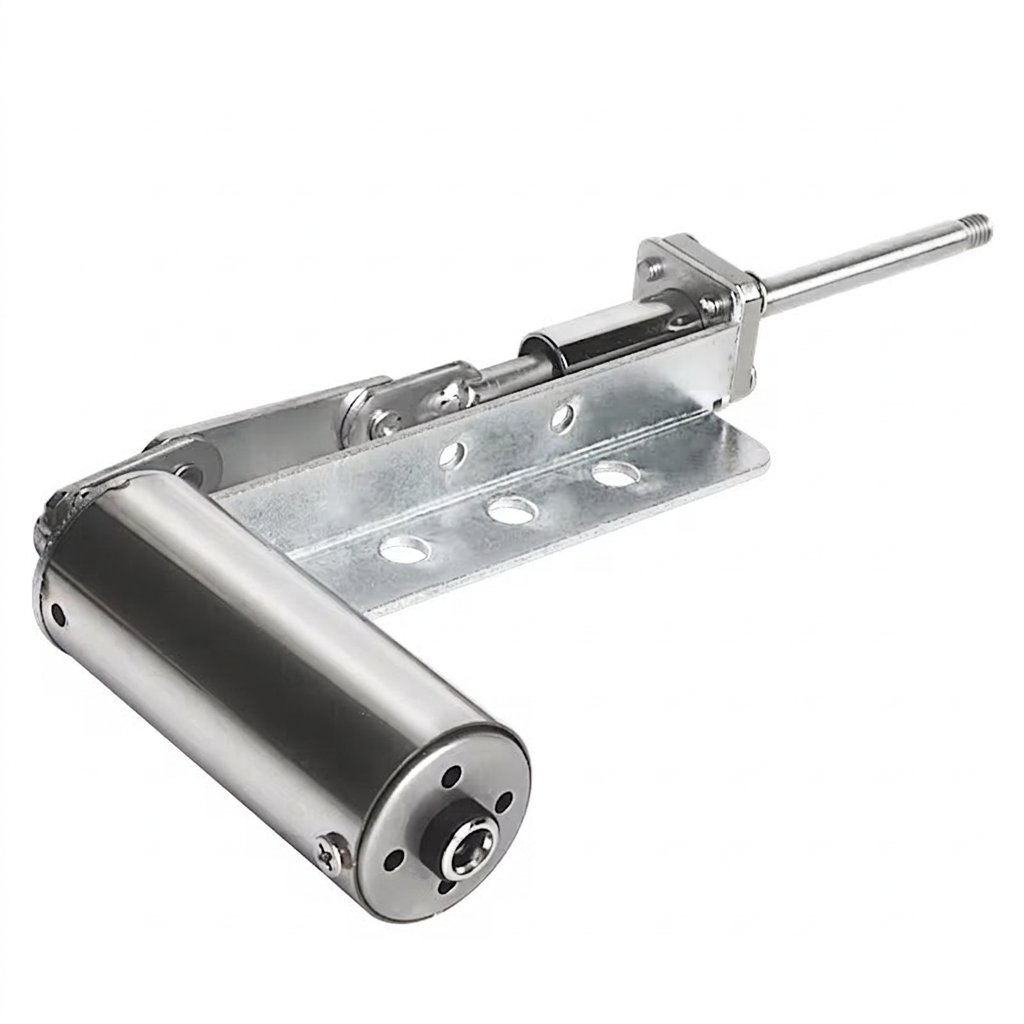} &
  \begin{tabular}[t]{@{}c@{}}
    {\scriptsize\makecell{no exported\\artifact}} \\
    {\tiny\ttfamily\makecell[c]{no exported artifact \\ 12.4 min · 24.3k}}
  \end{tabular} &
  \begin{tabular}[t]{@{}c@{}}
    \includegraphics[width=0.95\linewidth,height=1.48cm,keepaspectratio]{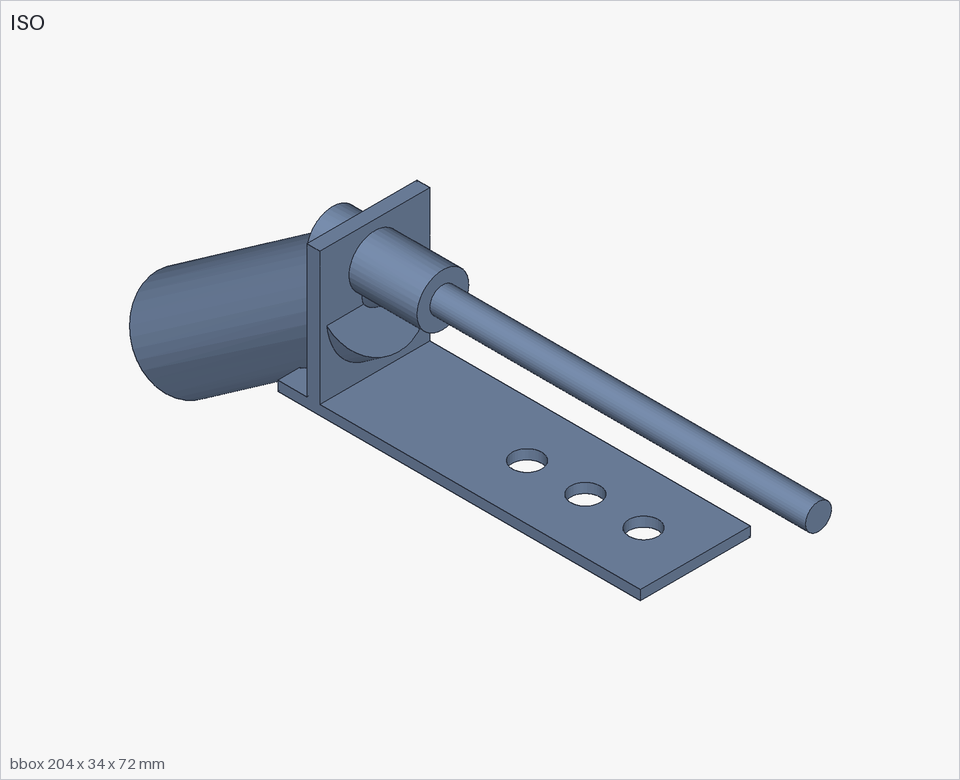} \\
    {\tiny\ttfamily\makecell[c]{Q 48.4 · F 62.1 · D 51.8 \\ 48.4 min · 67.3k}}
  \end{tabular} \\
\addlinespace[1pt]
\texttt{\scriptsize rcb\_000304481} &
  \includegraphics[width=0.95\linewidth,height=1.48cm,keepaspectratio]{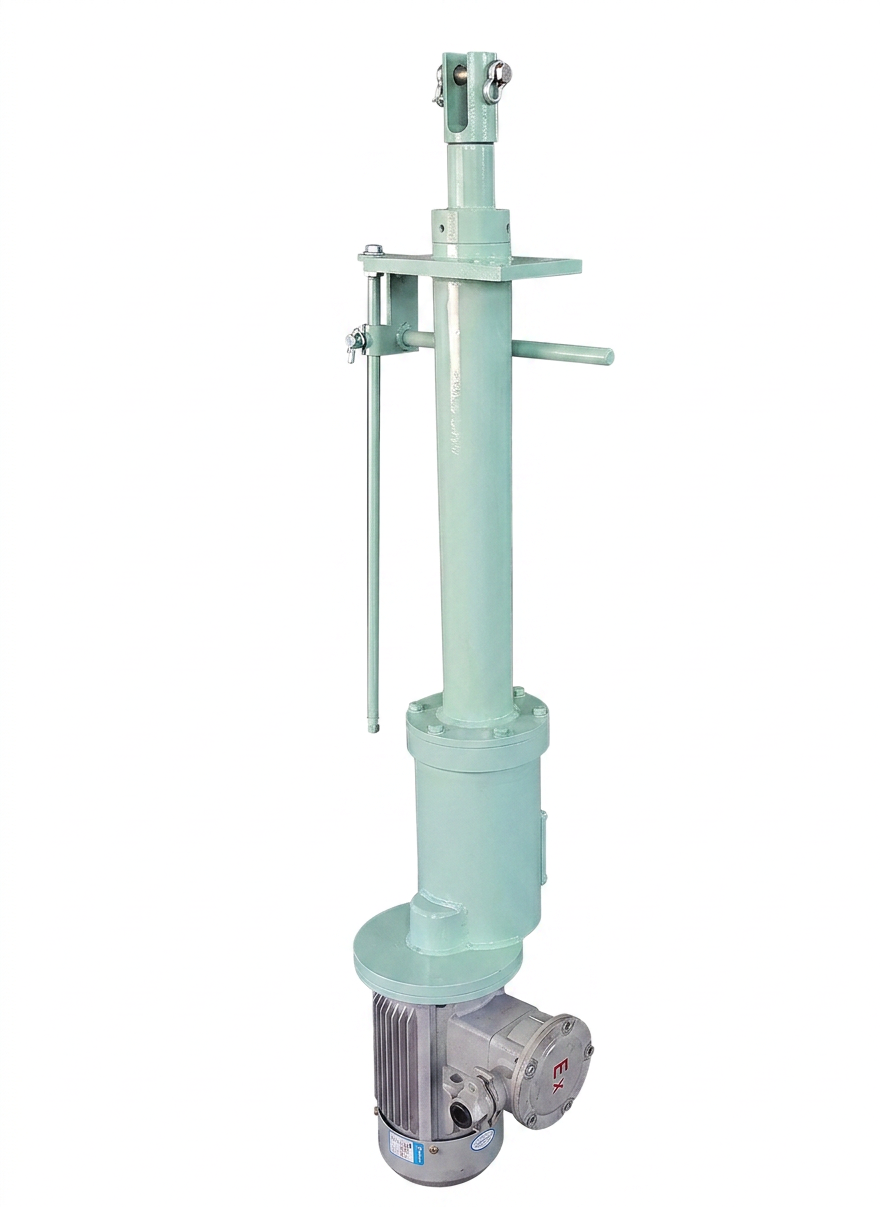} &
  \begin{tabular}[t]{@{}c@{}}
    \includegraphics[width=0.95\linewidth,height=1.48cm,keepaspectratio]{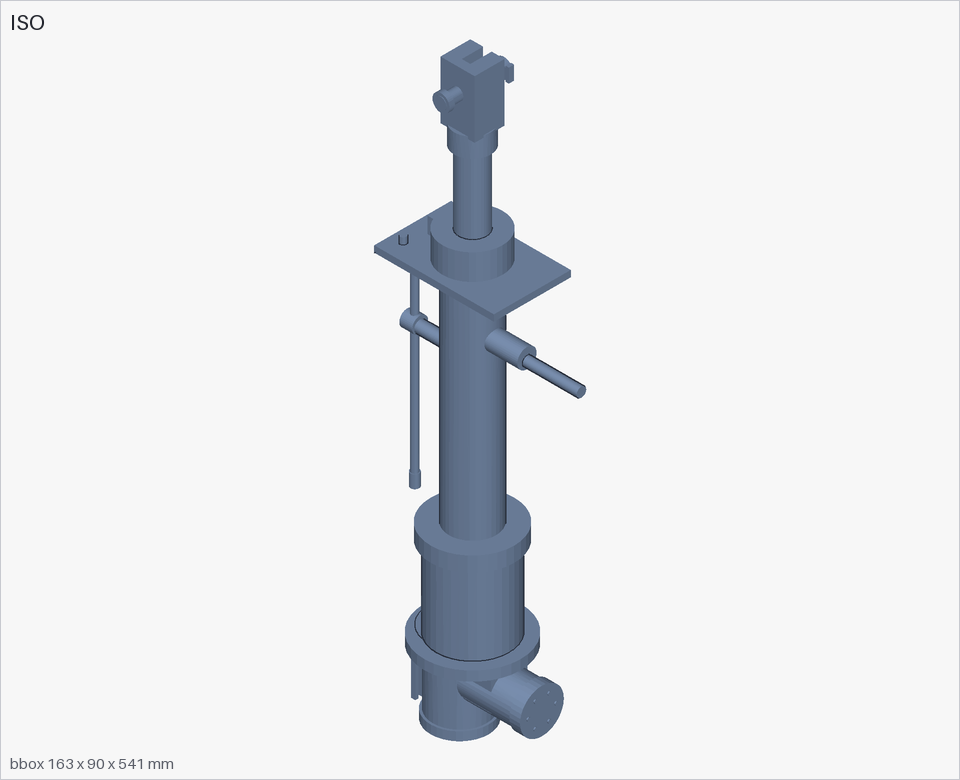} \\
    {\tiny\ttfamily\makecell[c]{Q 67.8 · F 79.2 · D 74.7 \\ 7.8 min · 17.3k}}
  \end{tabular} &
  \begin{tabular}[t]{@{}c@{}}
    \includegraphics[width=0.95\linewidth,height=1.48cm,keepaspectratio]{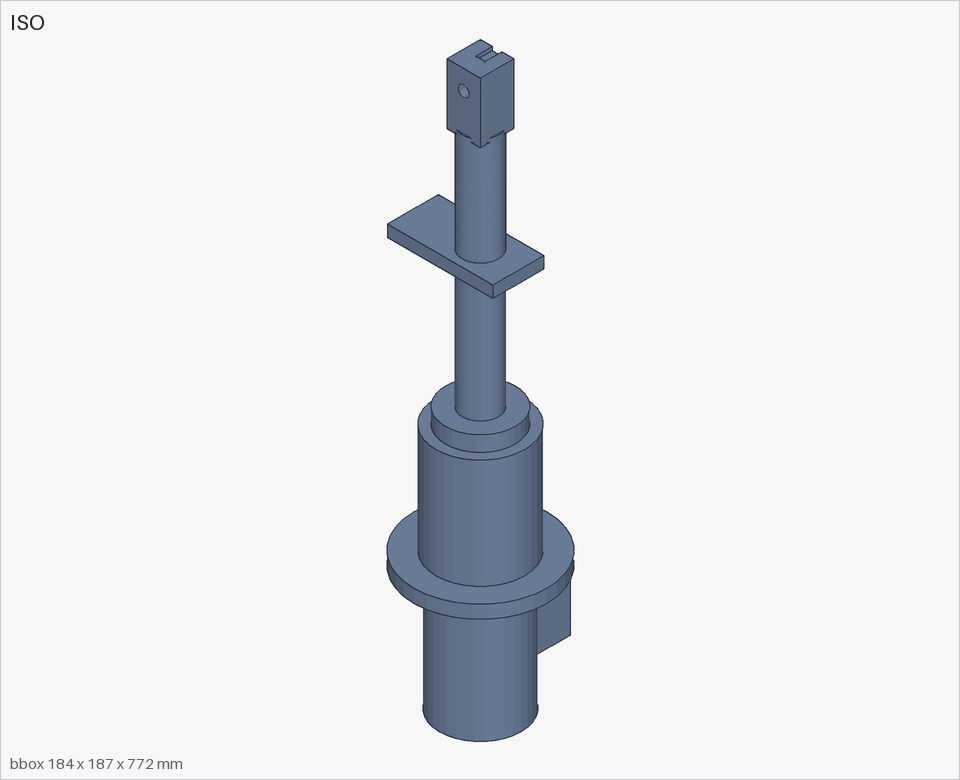} \\
    {\tiny\ttfamily\makecell[c]{Q 40.4 · F 56.0 · D 45.0 \\ 17.8 min · 50.3k}}
  \end{tabular} \\
\addlinespace[1pt]
\texttt{\scriptsize rcb\_000315240} &
  \includegraphics[width=0.95\linewidth,height=1.48cm,keepaspectratio]{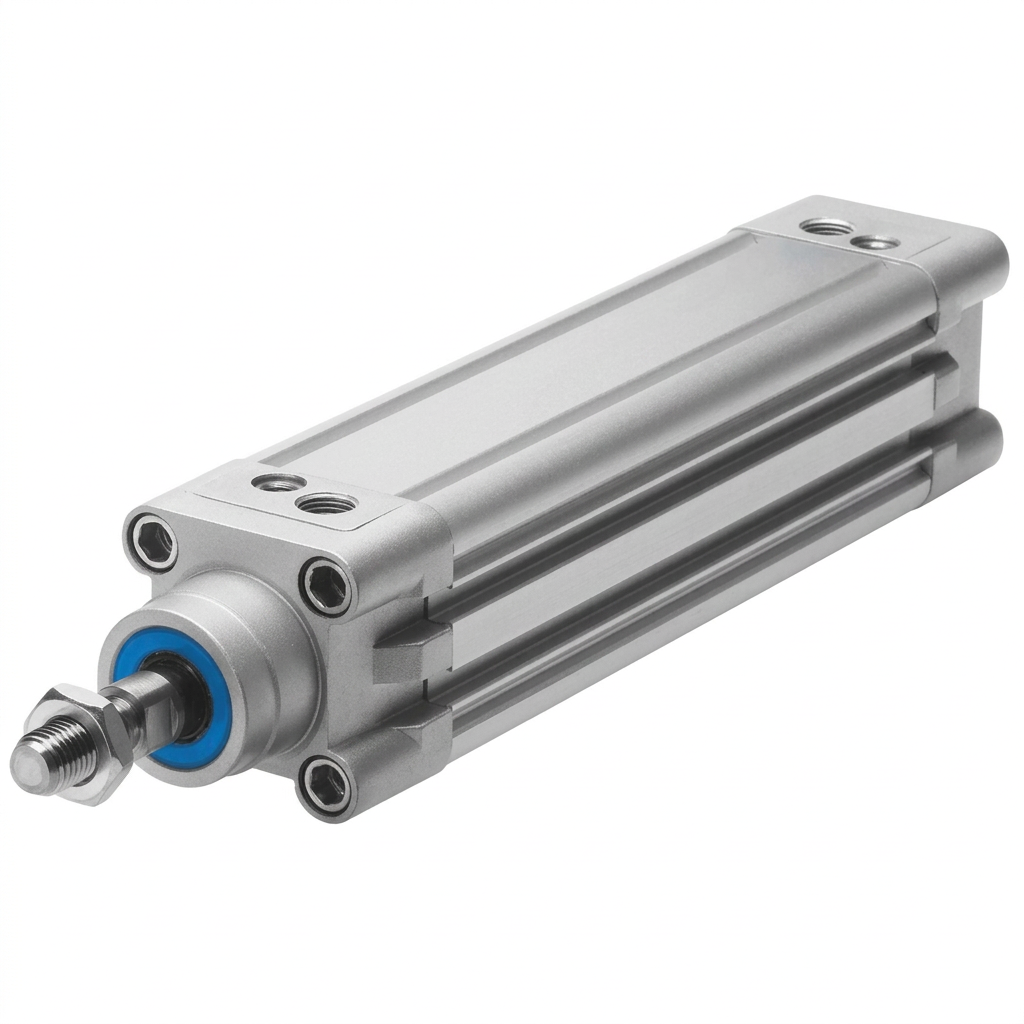} &
  \begin{tabular}[t]{@{}c@{}}
    \includegraphics[width=0.95\linewidth,height=1.48cm,keepaspectratio]{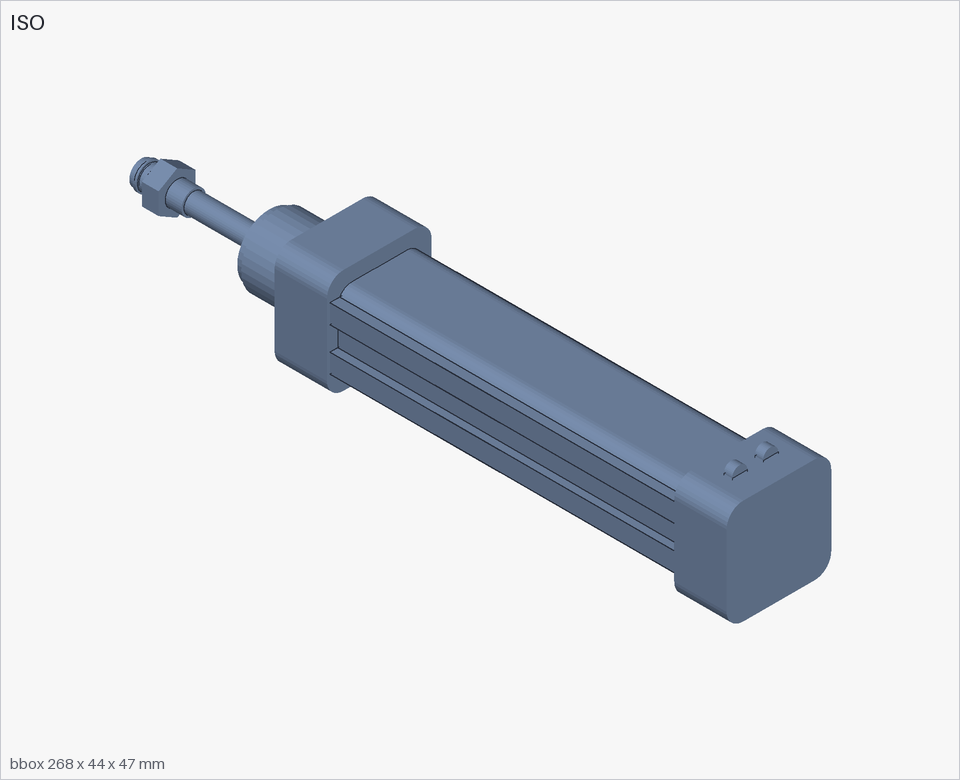} \\
    {\tiny\ttfamily\makecell[c]{Q 73.8 · F 83.2 · D 80.0 \\ 14.1 min · 20.9k}}
  \end{tabular} &
  \begin{tabular}[t]{@{}c@{}}
    \includegraphics[width=0.95\linewidth,height=1.48cm,keepaspectratio]{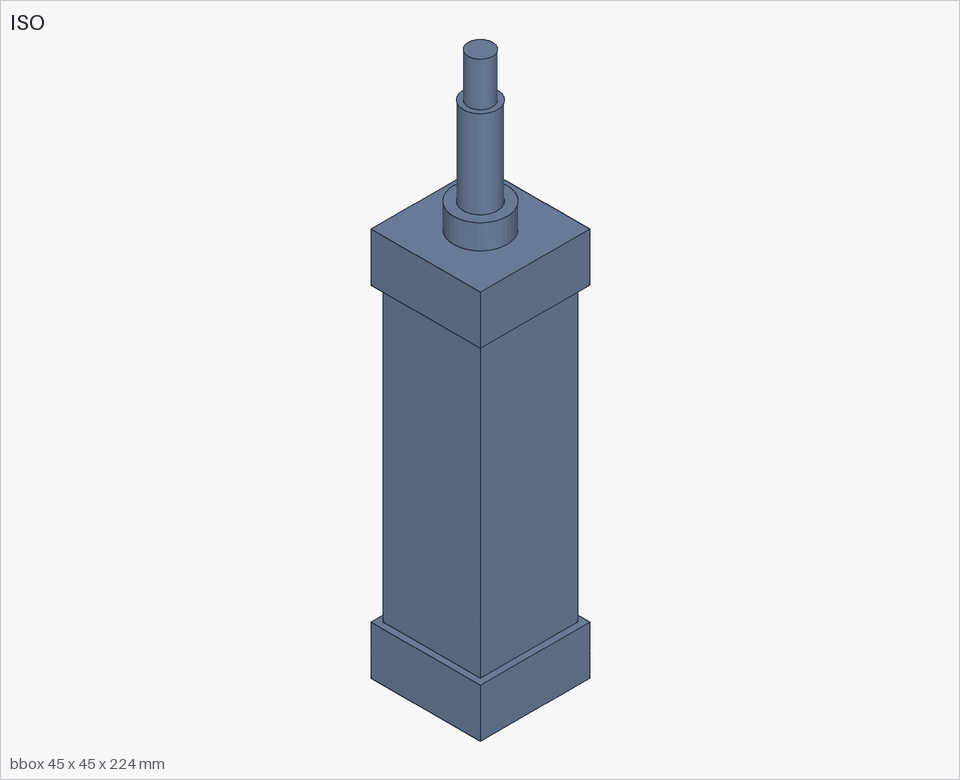} \\
    {\tiny\ttfamily\makecell[c]{Q 42.2 · F 65.5 · D 56.8 \\ 27.5 min · 37.7k}}
  \end{tabular} \\
\addlinespace[1pt]
\texttt{\scriptsize rcb\_000330472} &
  \includegraphics[width=0.95\linewidth,height=1.48cm,keepaspectratio]{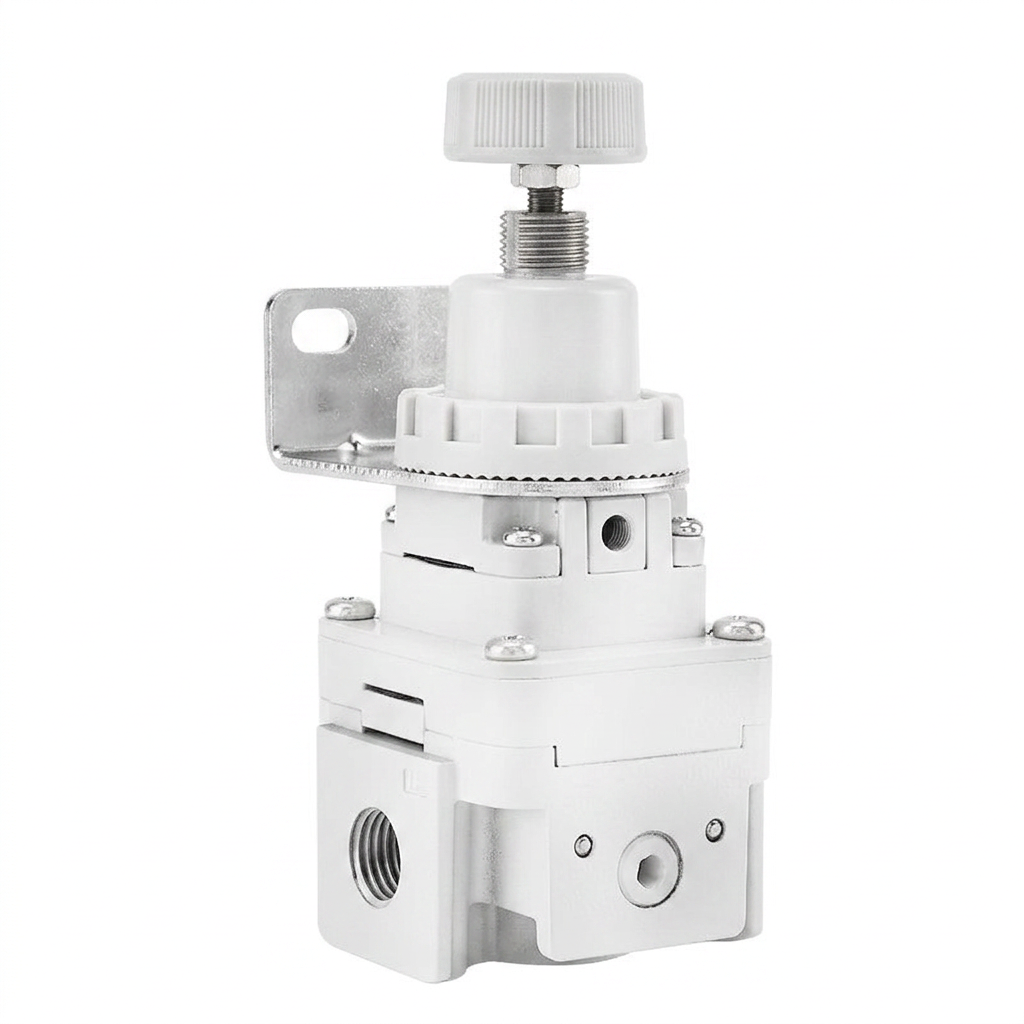} &
  \begin{tabular}[t]{@{}c@{}}
    \includegraphics[width=0.95\linewidth,height=1.48cm,keepaspectratio]{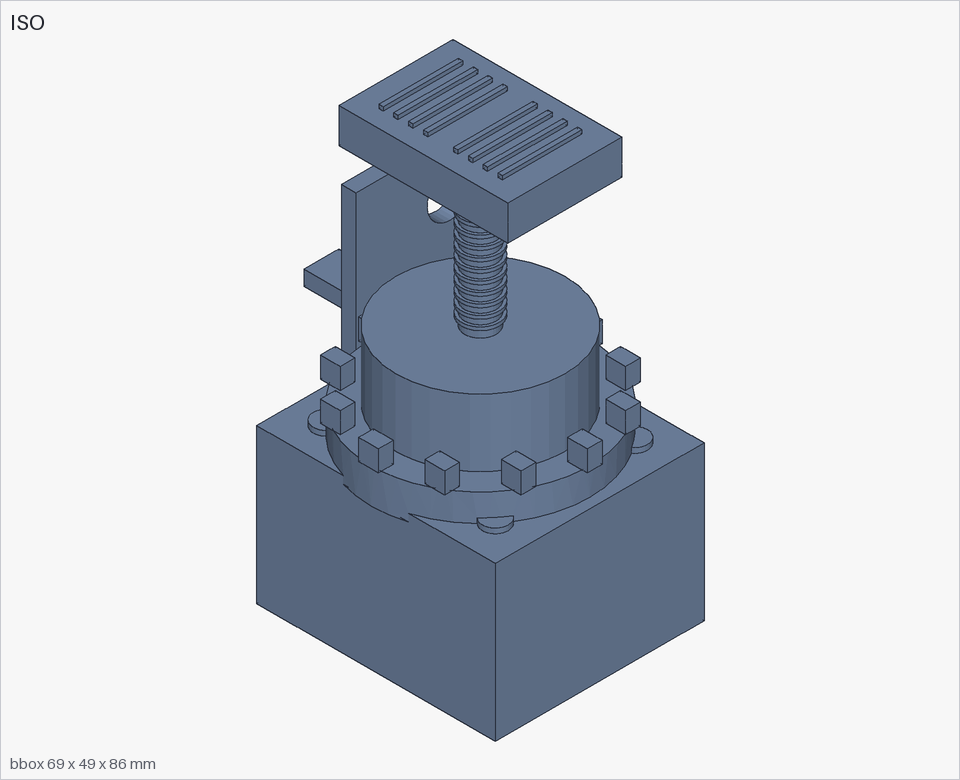} \\
    {\tiny\ttfamily\makecell[c]{Q 58.9 · F 73.4 · D 59.0 \\ 11.2 min · 23.1k}}
  \end{tabular} &
  \begin{tabular}[t]{@{}c@{}}
    \includegraphics[width=0.95\linewidth,height=1.48cm,keepaspectratio]{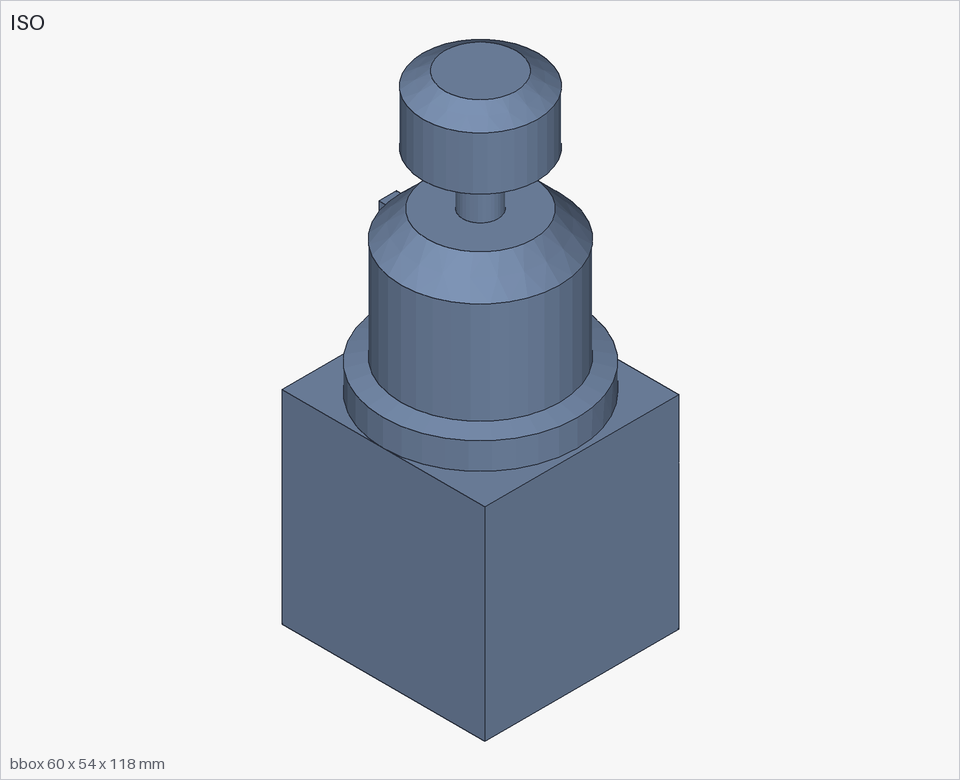} \\
    {\tiny\ttfamily\makecell[c]{Q 41.1 · F 61.1 · D 50.8 \\ 33.9 min · 39.5k}}
  \end{tabular} \\
\addlinespace[1pt]
\texttt{\scriptsize rcb\_000342241} &
  \includegraphics[width=0.95\linewidth,height=1.48cm,keepaspectratio]{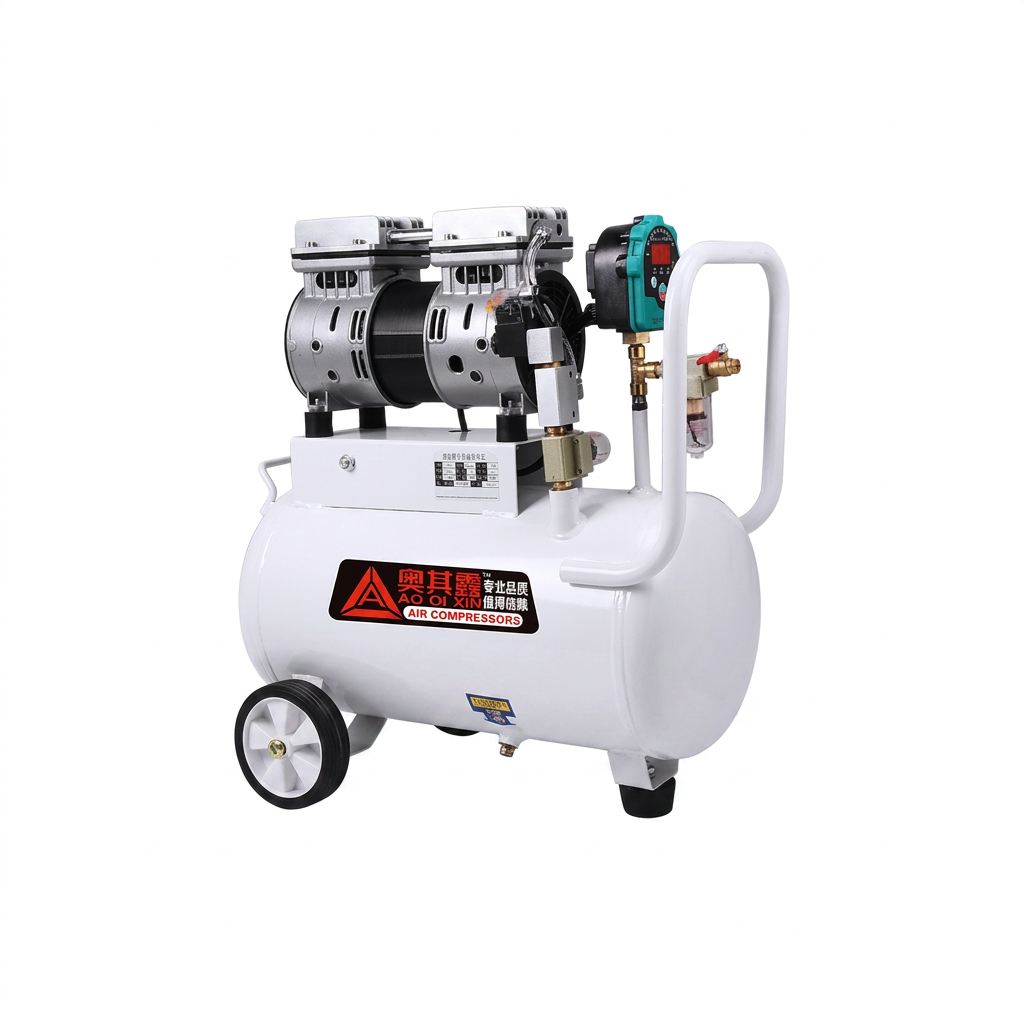} &
  \begin{tabular}[t]{@{}c@{}}
    \includegraphics[width=0.95\linewidth,height=1.48cm,keepaspectratio]{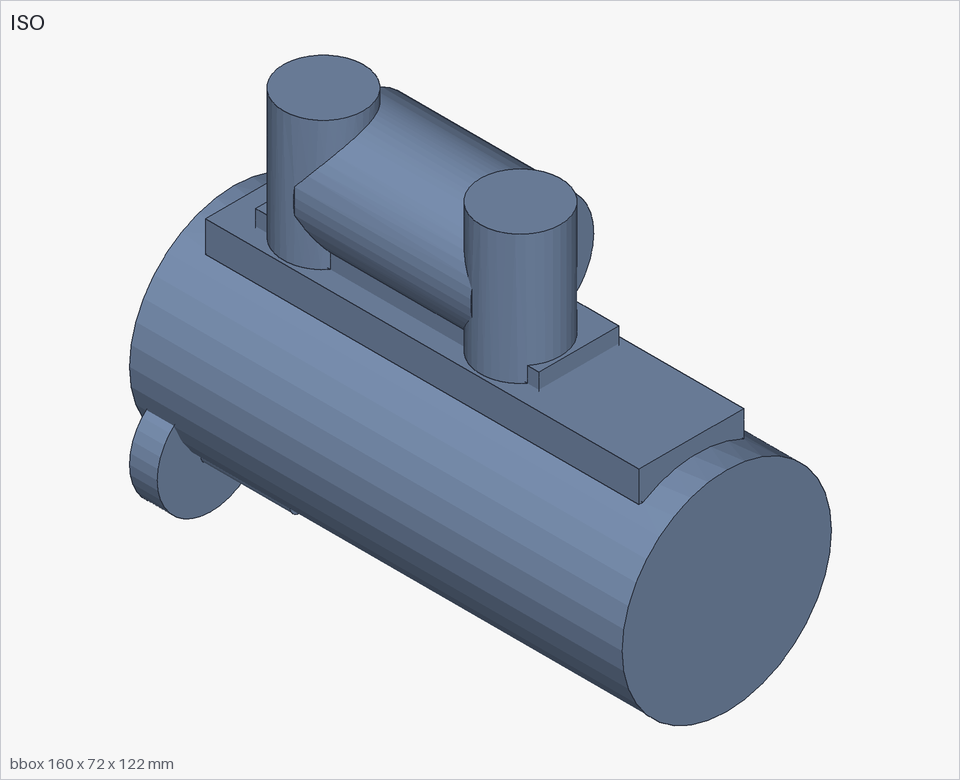} \\
    {\tiny\ttfamily\makecell[c]{Q 43.5 · F 53.0 · D 44.7 \\ 13.5 min · 26.0k}}
  \end{tabular} &
  \begin{tabular}[t]{@{}c@{}}
    \includegraphics[width=0.95\linewidth,height=1.48cm,keepaspectratio]{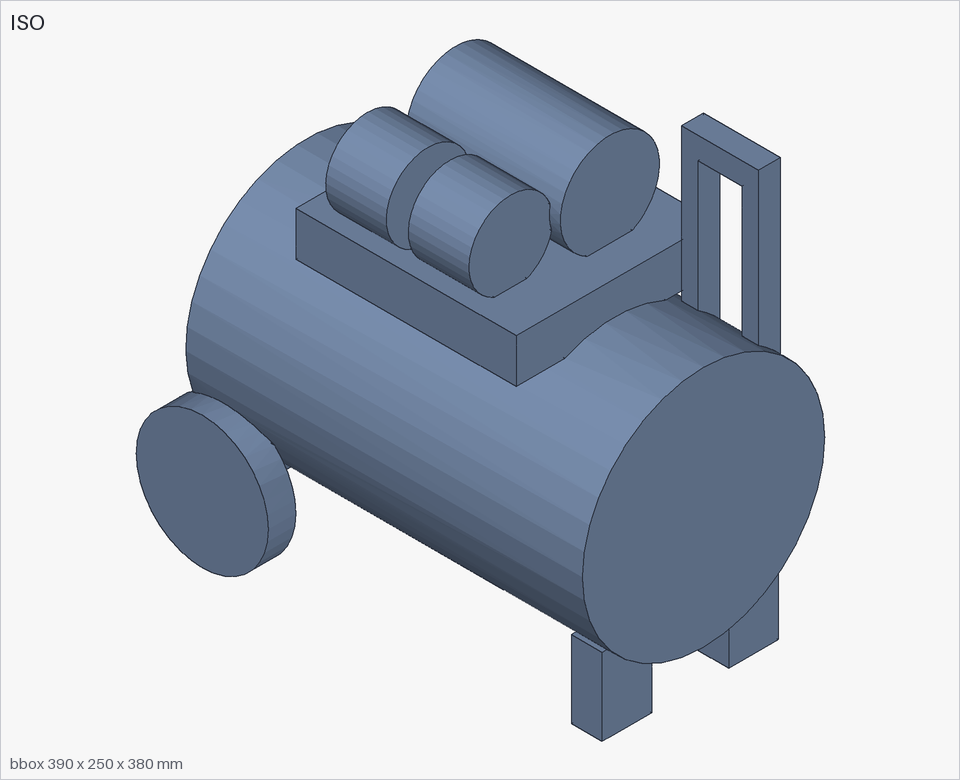} \\
    {\tiny\ttfamily\makecell[c]{Q 51.3 · F 68.6 · D 60.2 \\ 36.9 min · 53.3k}}
  \end{tabular} \\
\addlinespace[1pt]
\texttt{\scriptsize rcb\_000348219} &
  \includegraphics[width=0.95\linewidth,height=1.48cm,keepaspectratio]{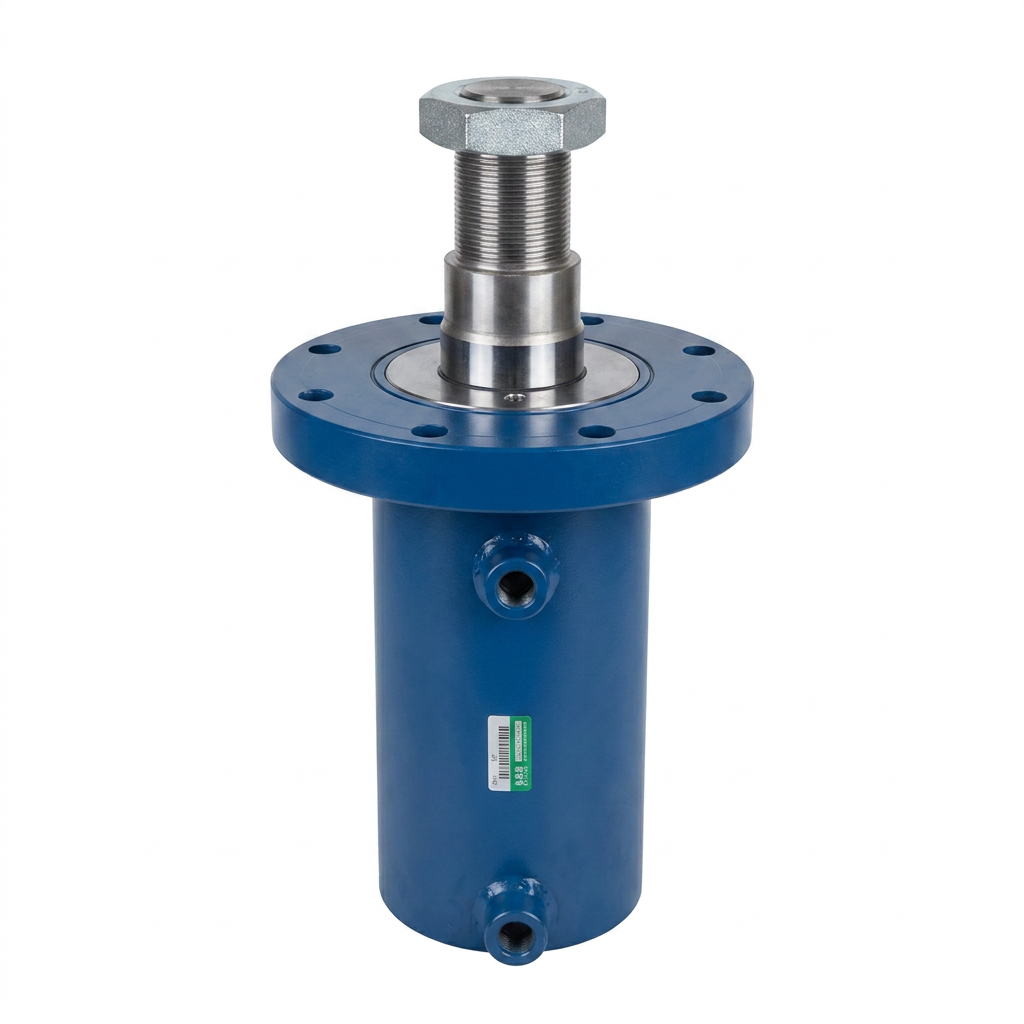} &
  \begin{tabular}[t]{@{}c@{}}
    \includegraphics[width=0.95\linewidth,height=1.48cm,keepaspectratio]{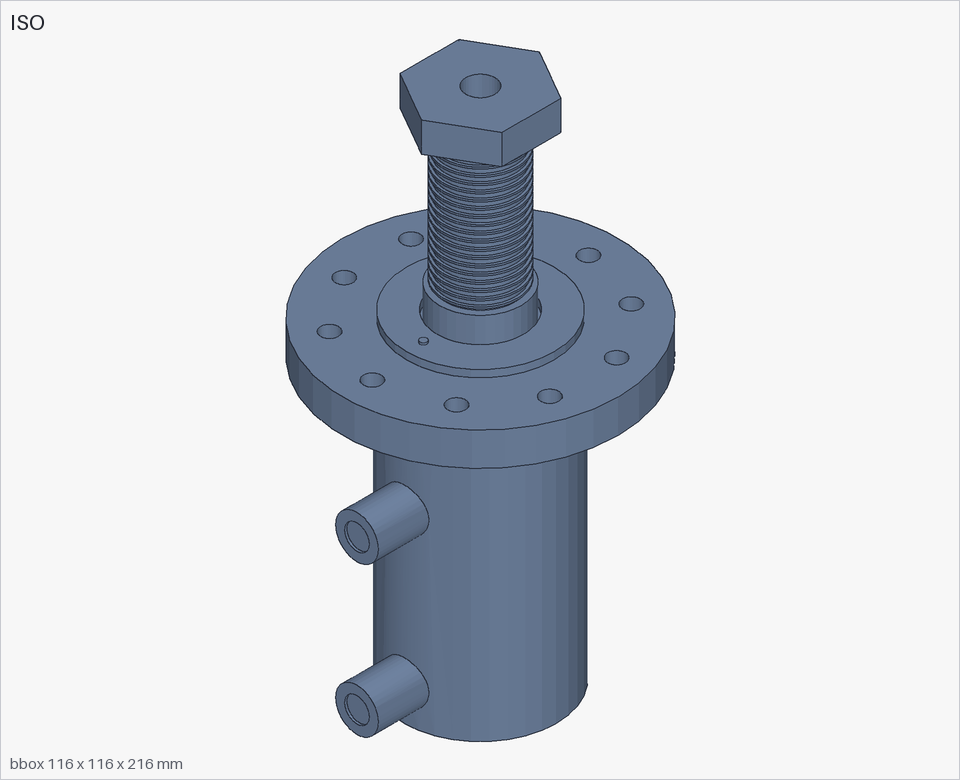} \\
    {\tiny\ttfamily\makecell[c]{Q 86.7 · F 90.3 · D 87.5 \\ 11.7 min · 24.8k}}
  \end{tabular} &
  \begin{tabular}[t]{@{}c@{}}
    \includegraphics[width=0.95\linewidth,height=1.48cm,keepaspectratio]{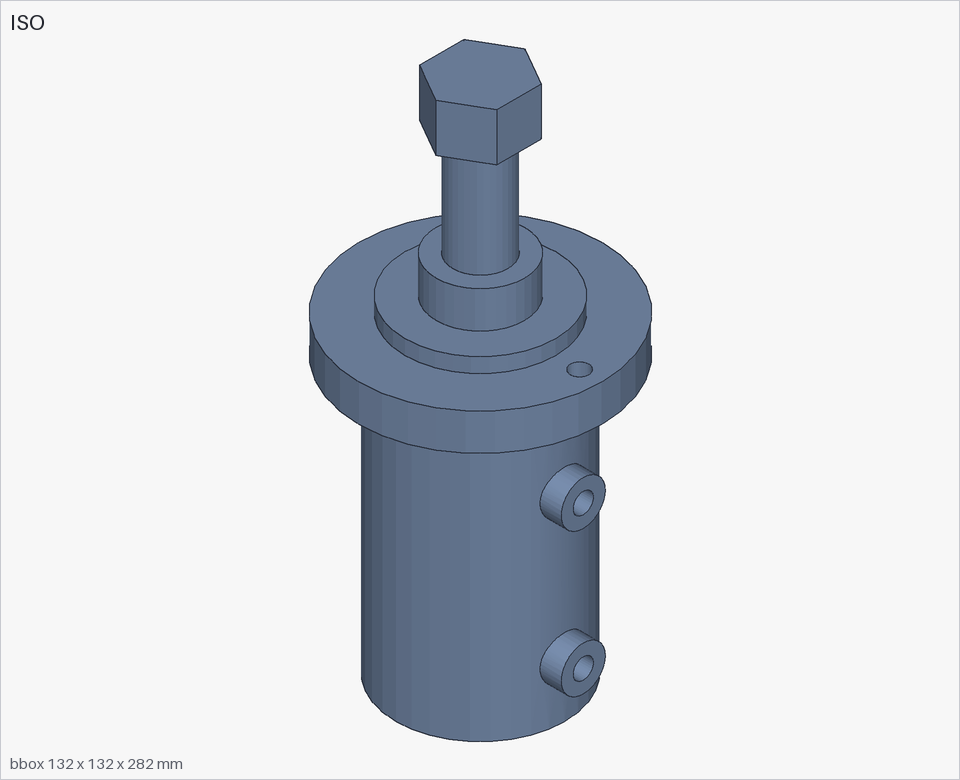} \\
    {\tiny\ttfamily\makecell[c]{Q 69.0 · F 82.7 · D 73.0 \\ 46.6 min · 52.7k}}
  \end{tabular} \\
\addlinespace[1pt]
\texttt{\scriptsize rcb\_000352539} &
  \includegraphics[width=0.95\linewidth,height=1.48cm,keepaspectratio]{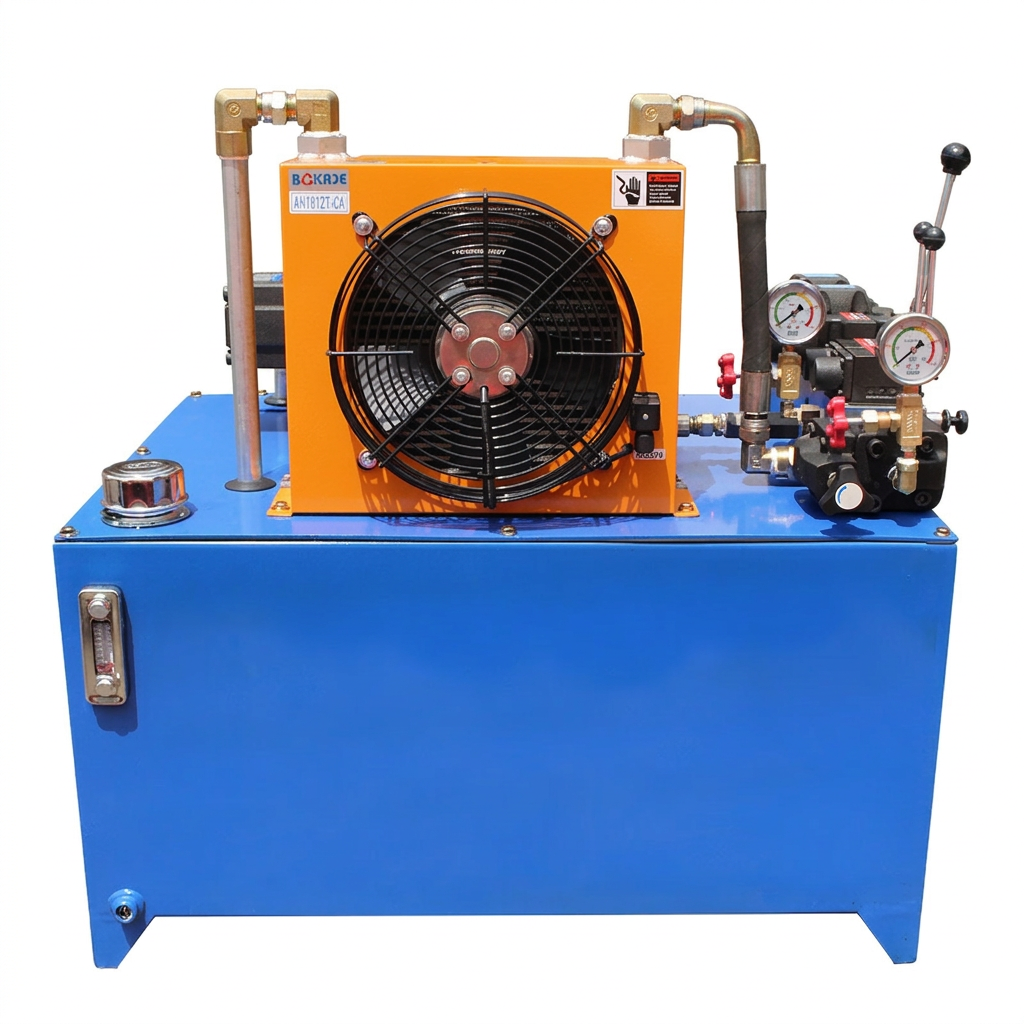} &
  \begin{tabular}[t]{@{}c@{}}
    \includegraphics[width=0.95\linewidth,height=1.48cm,keepaspectratio]{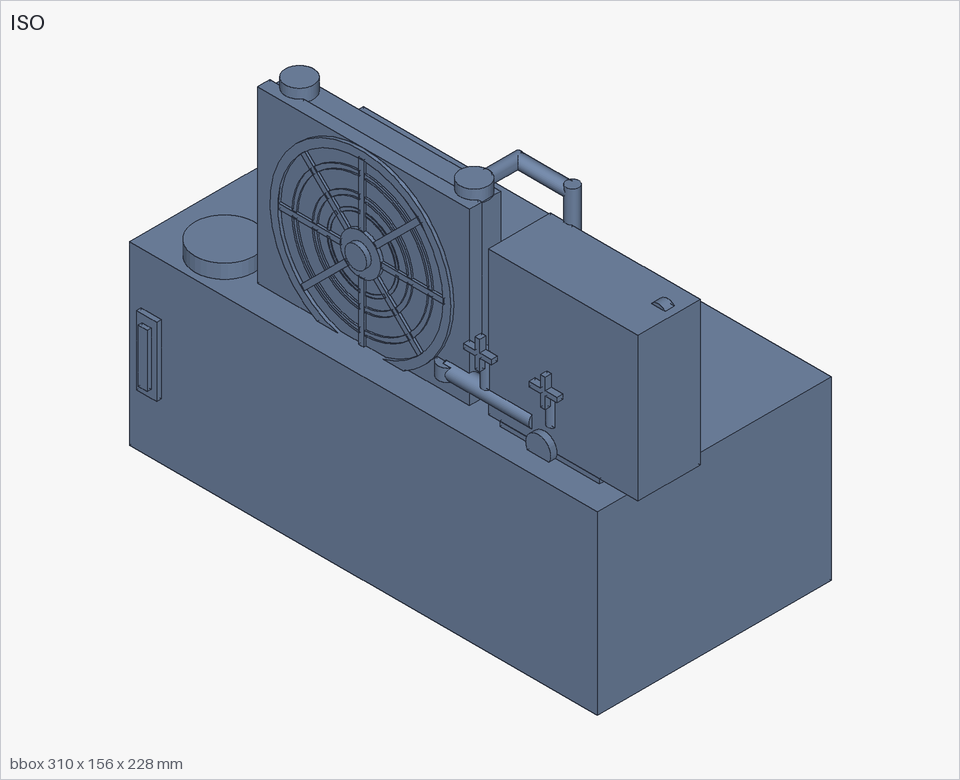} \\
    {\tiny\ttfamily\makecell[c]{Q 61.5 · F 70.2 · D 61.2 \\ 11.4 min · 26.2k}}
  \end{tabular} &
  \begin{tabular}[t]{@{}c@{}}
    \includegraphics[width=0.95\linewidth,height=1.48cm,keepaspectratio]{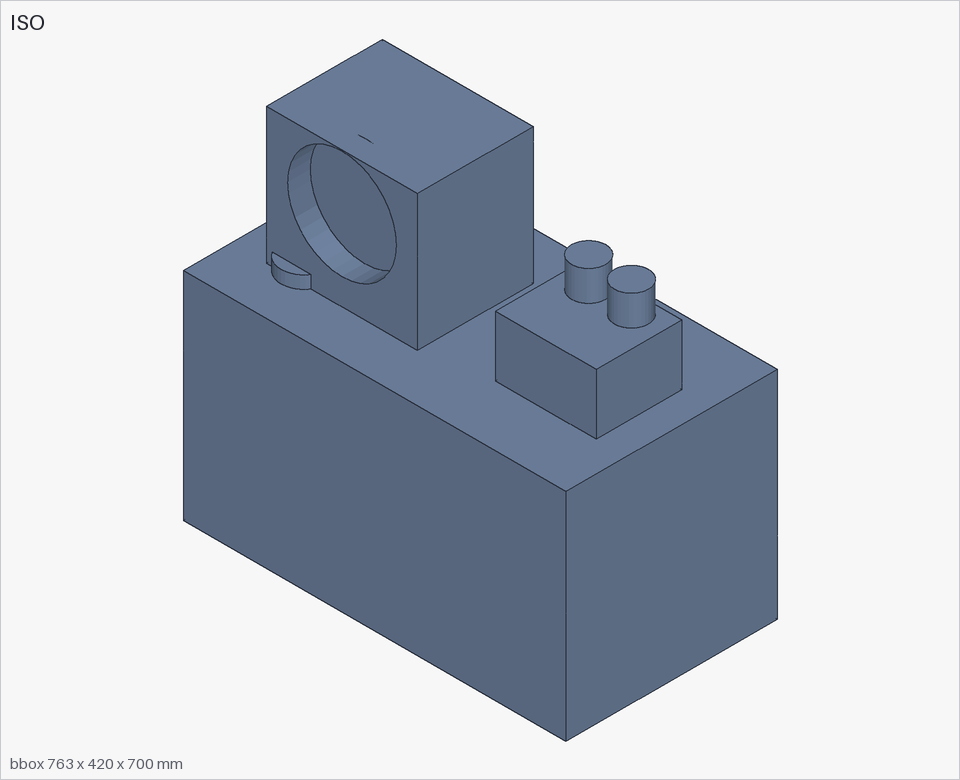} \\
    {\tiny\ttfamily\makecell[c]{Q 37.7 · F 57.3 · D 40.2 \\ --- · ---}}
  \end{tabular} \\
\addlinespace[1pt]
\texttt{\scriptsize rcb\_000362639} &
  \includegraphics[width=0.95\linewidth,height=1.48cm,keepaspectratio]{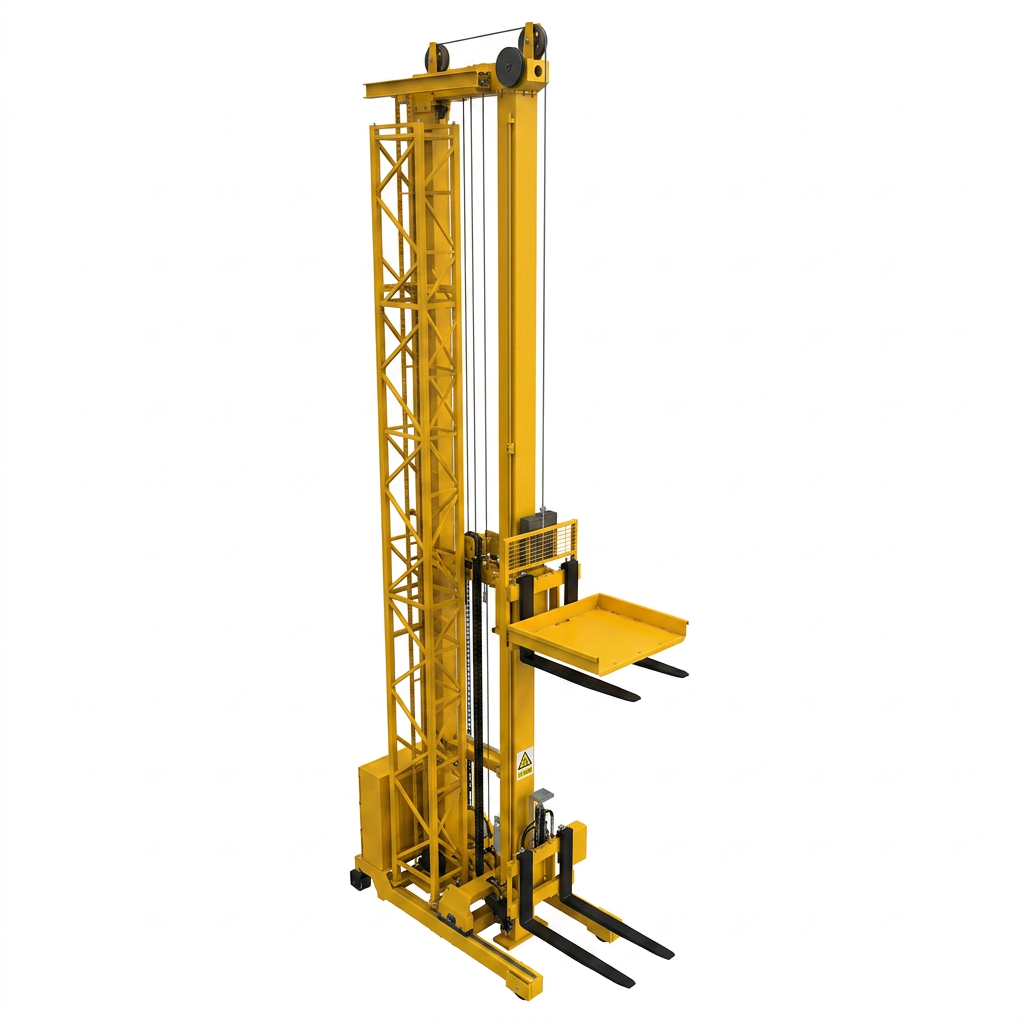} &
  \begin{tabular}[t]{@{}c@{}}
    \includegraphics[width=0.95\linewidth,height=1.48cm,keepaspectratio]{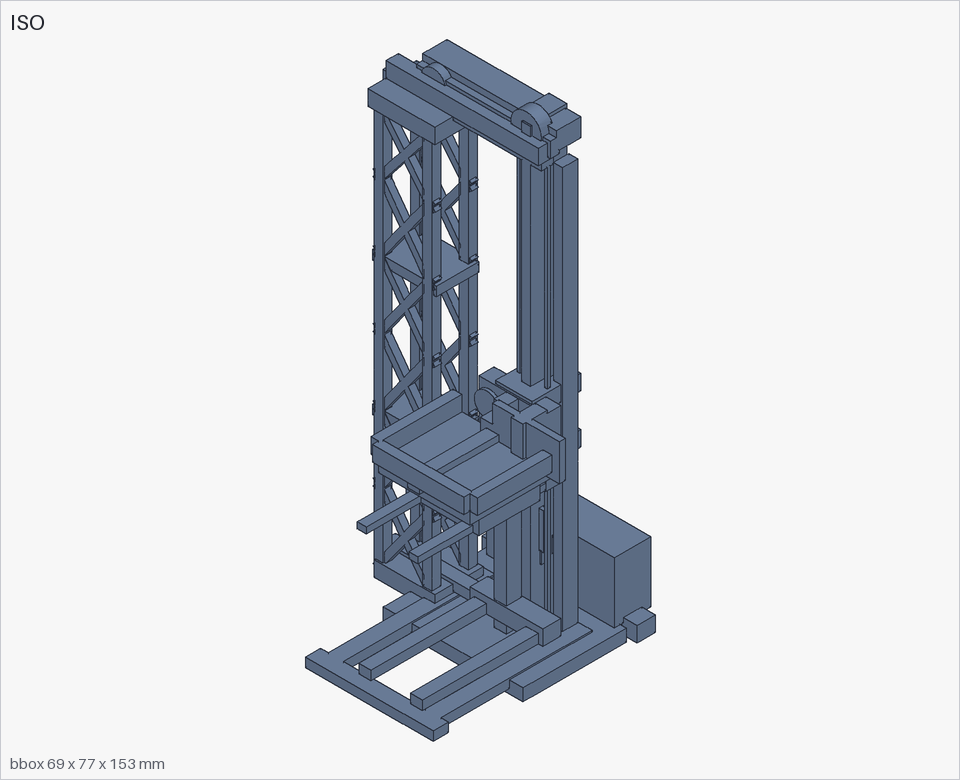} \\
    {\tiny\ttfamily\makecell[c]{Q 76.2 · F 83.2 · D 78.0 \\ 12.7 min · 23.3k}}
  \end{tabular} &
  \begin{tabular}[t]{@{}c@{}}
    \includegraphics[width=0.95\linewidth,height=1.48cm,keepaspectratio]{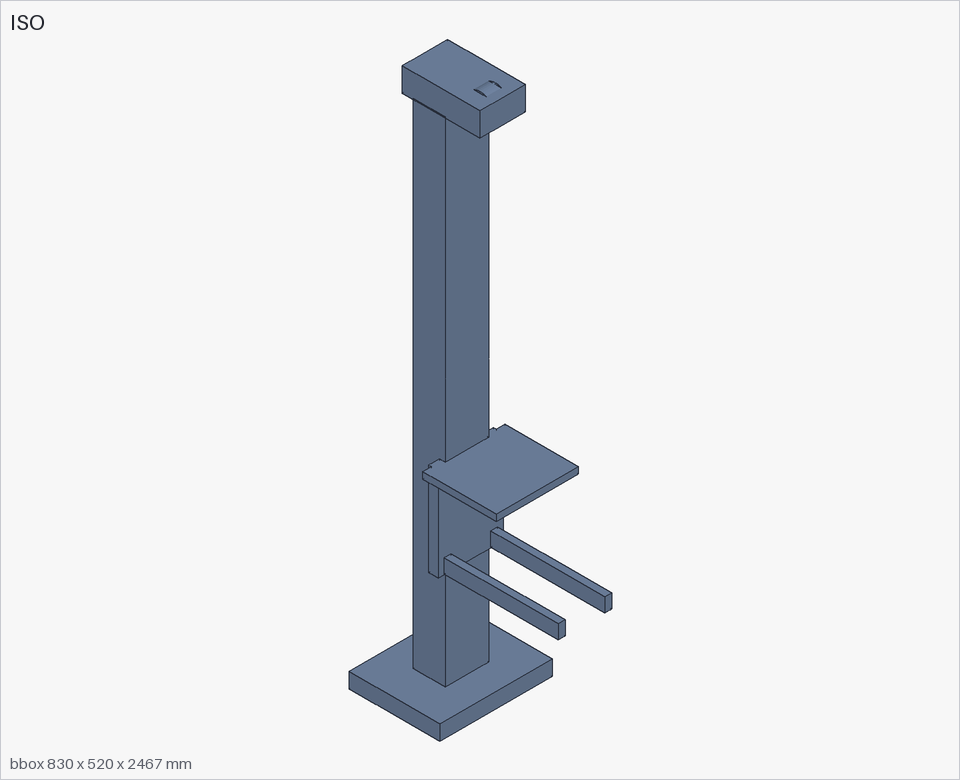} \\
    {\tiny\ttfamily\makecell[c]{Q 35.9 · F 51.4 · D 35.6 \\ 28.9 min · 44.3k}}
  \end{tabular} \\
\addlinespace[1pt]
\texttt{\scriptsize rcb\_000371197} &
  \includegraphics[width=0.95\linewidth,height=1.48cm,keepaspectratio]{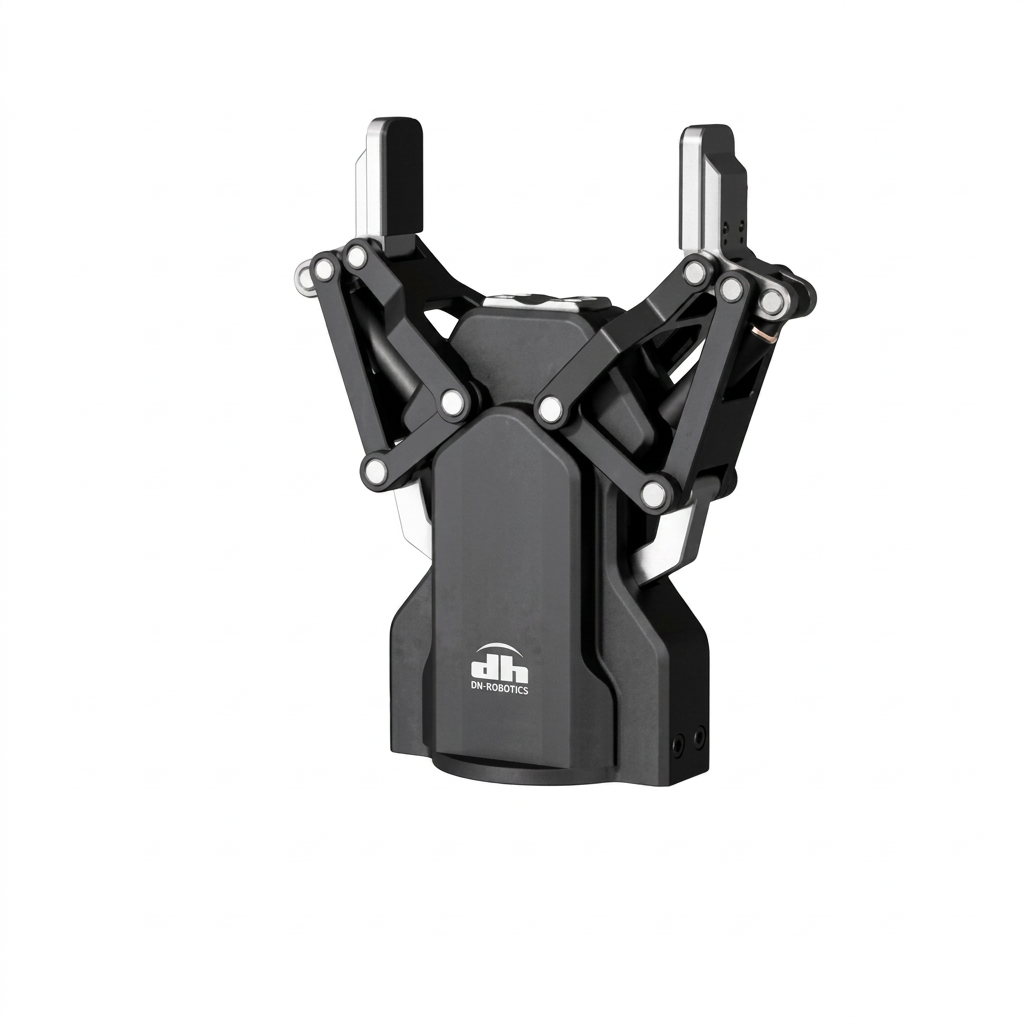} &
  \begin{tabular}[t]{@{}c@{}}
    \includegraphics[width=0.95\linewidth,height=1.48cm,keepaspectratio]{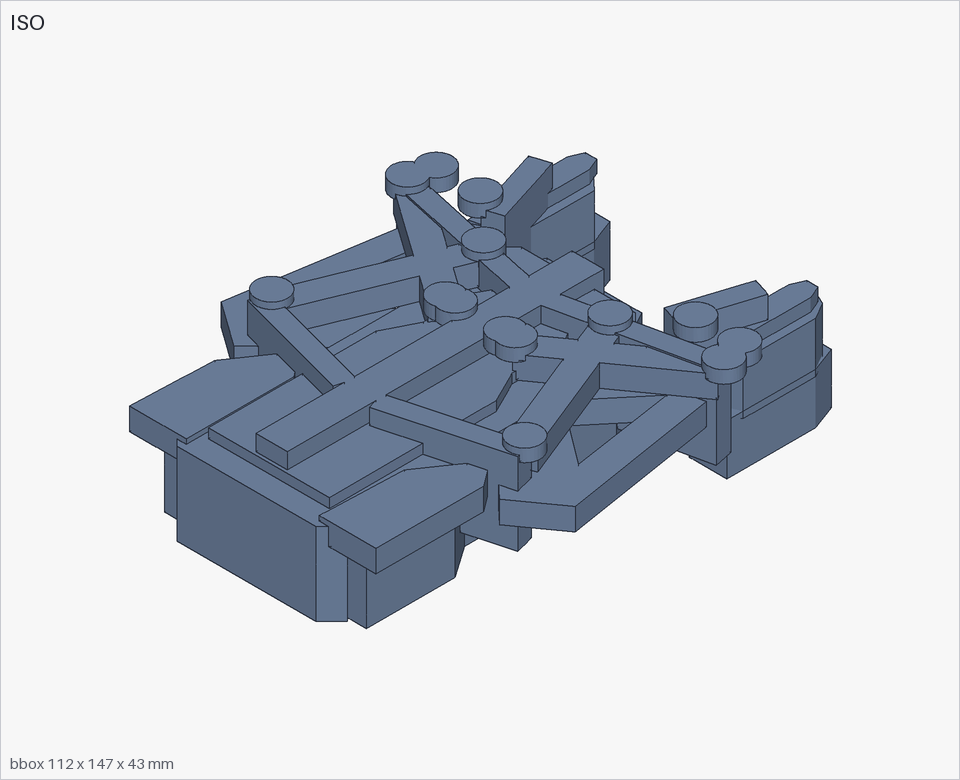} \\
    {\tiny\ttfamily\makecell[c]{Q 70.3 · F 73.7 · D 73.5 \\ 13.5 min · 26.5k}}
  \end{tabular} &
  \begin{tabular}[t]{@{}c@{}}
    \includegraphics[width=0.95\linewidth,height=1.48cm,keepaspectratio]{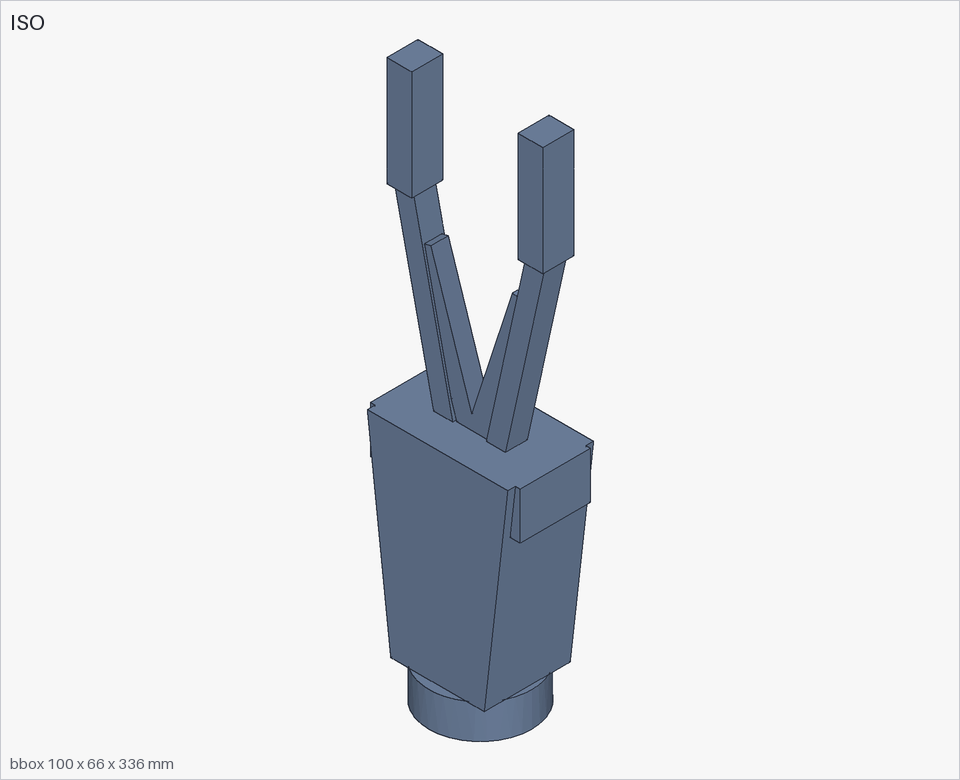} \\
    {\tiny\ttfamily\makecell[c]{Q 38.8 · F 56.2 · D 41.5 \\ 25.6 min · 40.6k}}
  \end{tabular} \\
\addlinespace[1pt]
\texttt{\scriptsize rcb\_000398895} &
  \includegraphics[width=0.95\linewidth,height=1.48cm,keepaspectratio]{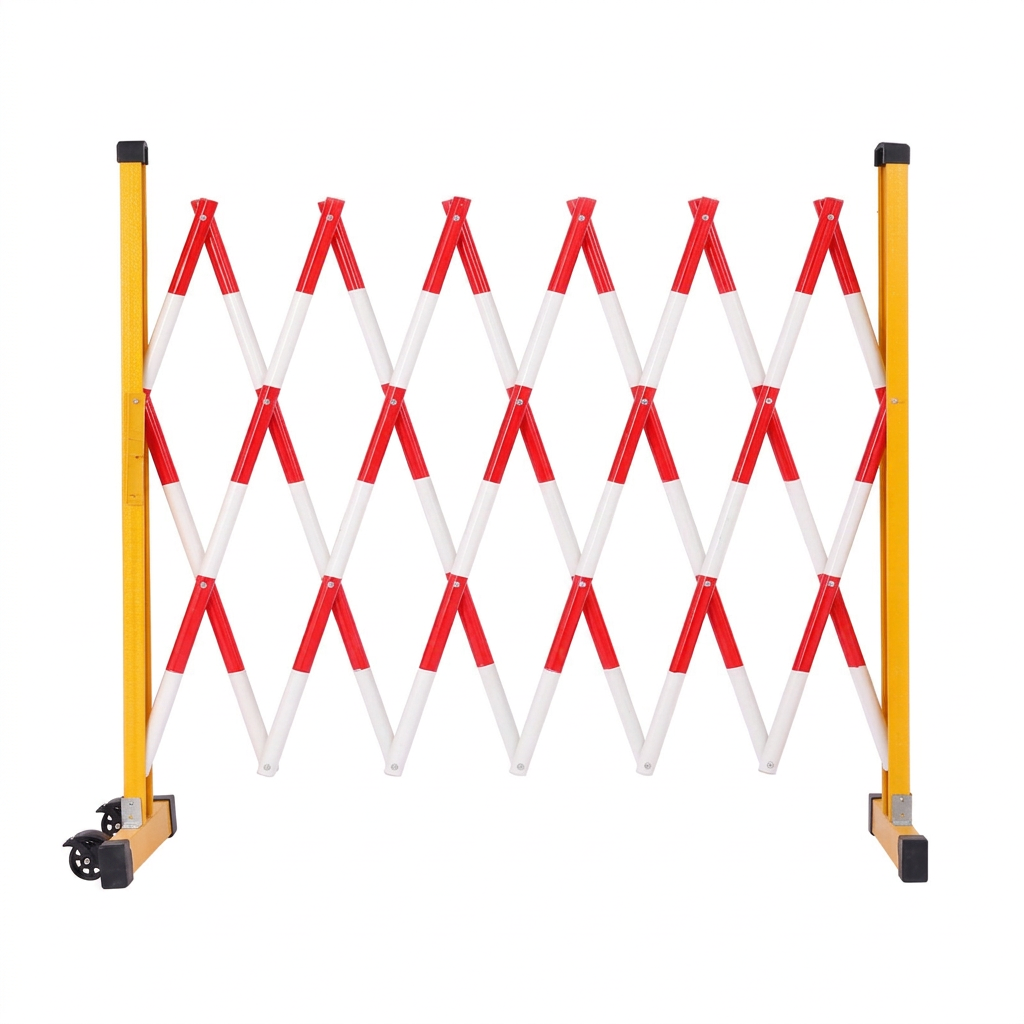} &
  \begin{tabular}[t]{@{}c@{}}
    \includegraphics[width=0.95\linewidth,height=1.48cm,keepaspectratio]{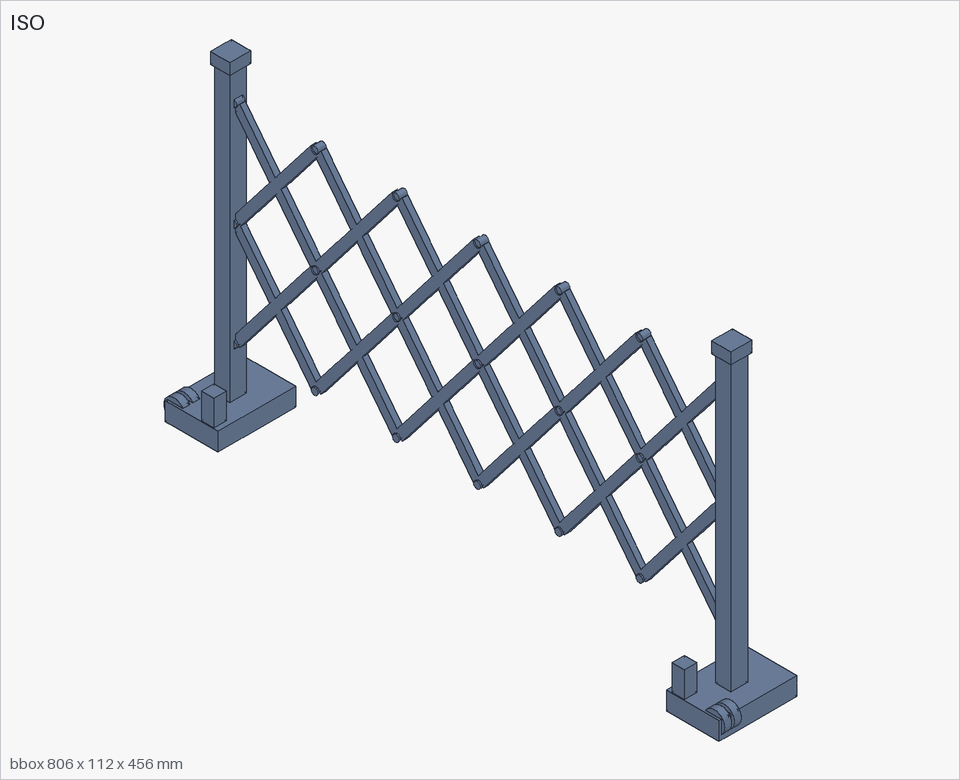} \\
    {\tiny\ttfamily\makecell[c]{Q 82.0 · F 85.7 · D 83.5 \\ 6.2 min · 17.0k}}
  \end{tabular} &
  \begin{tabular}[t]{@{}c@{}}
    \includegraphics[width=0.95\linewidth,height=1.48cm,keepaspectratio]{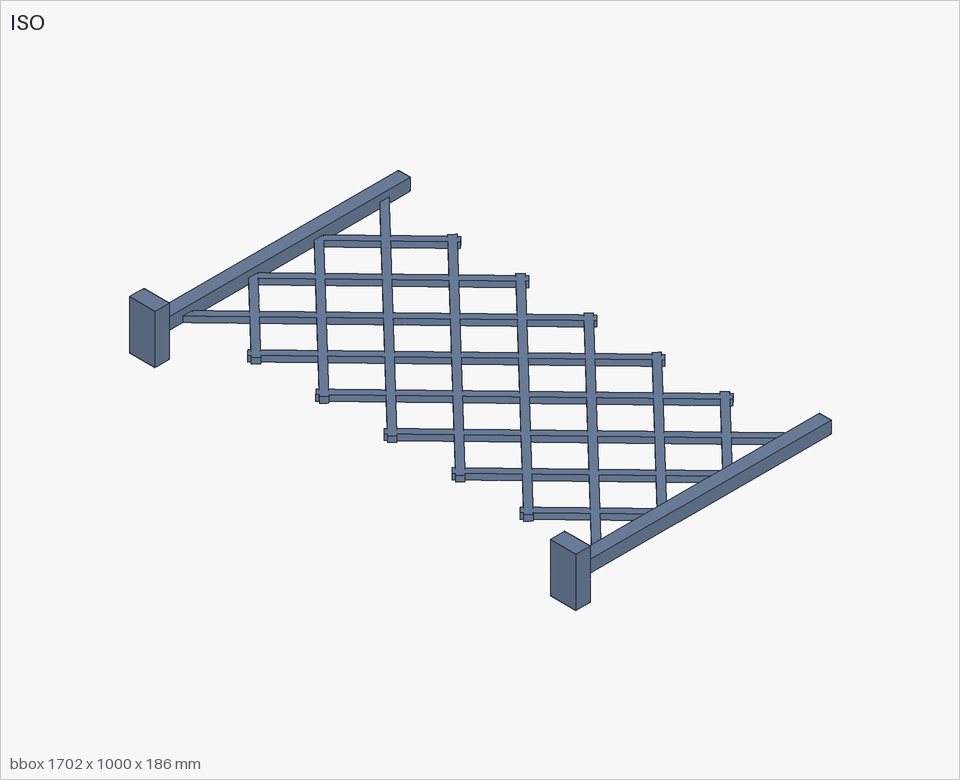} \\
    {\tiny\ttfamily\makecell[c]{Q 59.5 · F 52.0 · D 56.0 \\ 29.7 min · 55.0k}}
  \end{tabular} \\
\addlinespace[1pt]
\texttt{\scriptsize rcb\_000398996} &
  \includegraphics[width=0.95\linewidth,height=1.48cm,keepaspectratio]{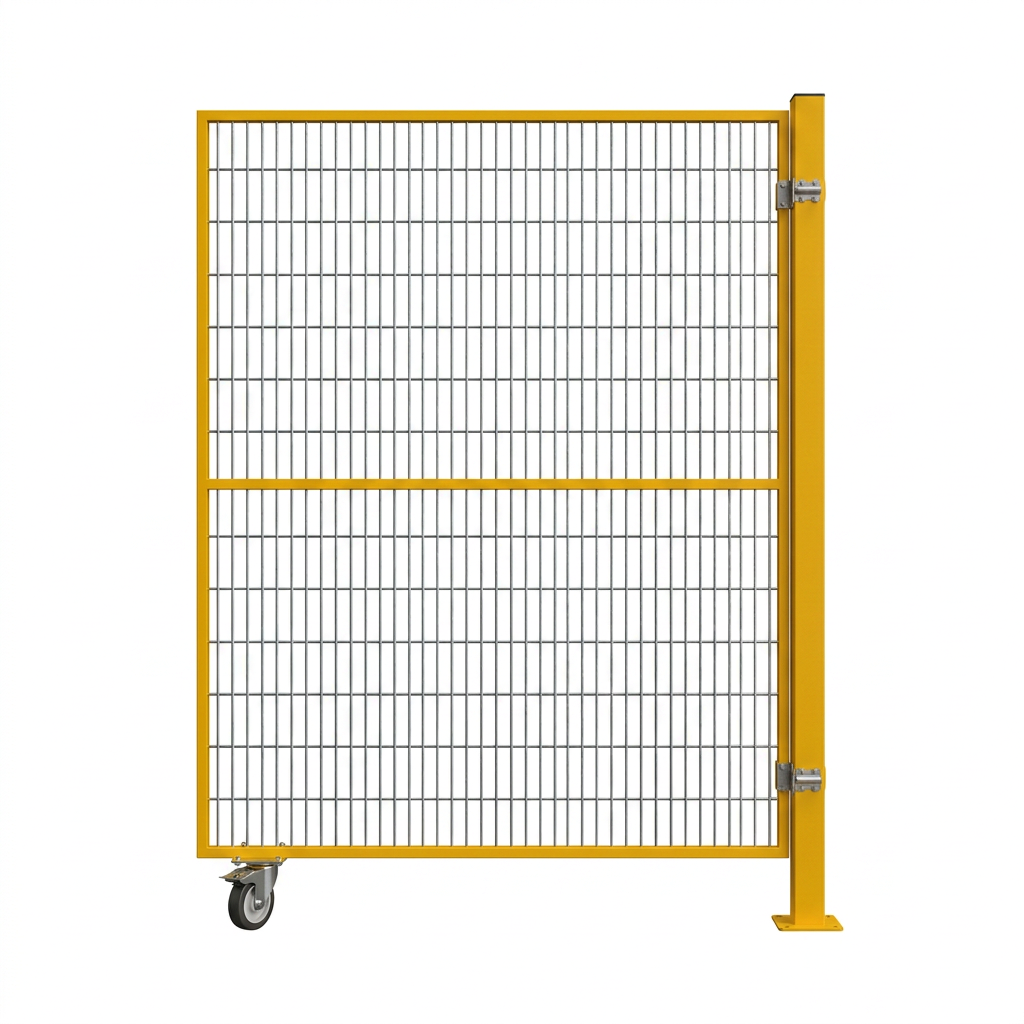} &
  \begin{tabular}[t]{@{}c@{}}
    \includegraphics[width=0.95\linewidth,height=1.48cm,keepaspectratio]{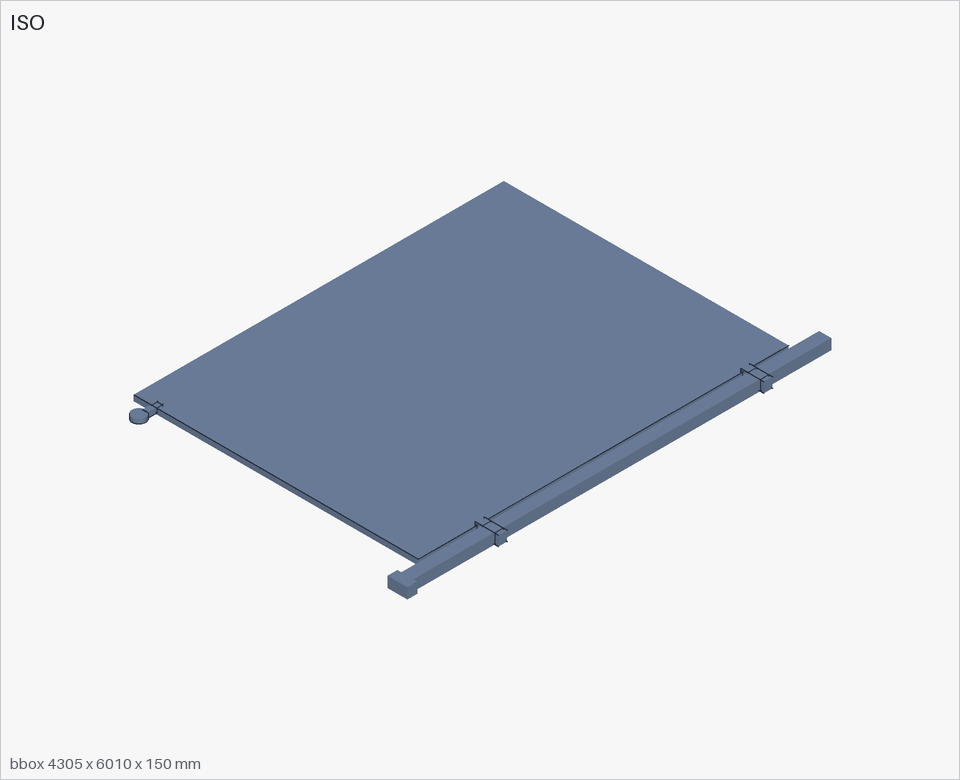} \\
    {\tiny\ttfamily\makecell[c]{Q 39.5 · F 48.5 · D 49.5 \\ 9.3 min · 24.1k}}
  \end{tabular} &
  \begin{tabular}[t]{@{}c@{}}
    \includegraphics[width=0.95\linewidth,height=1.48cm,keepaspectratio]{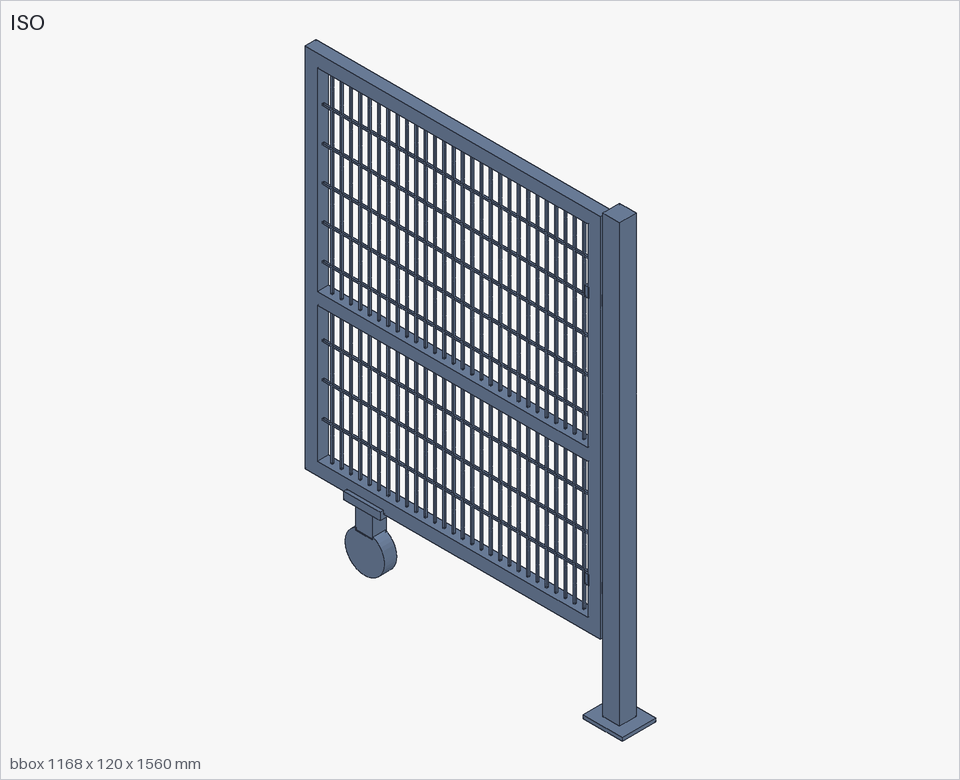} \\
    {\tiny\ttfamily\makecell[c]{Q 85.0 · F 87.4 · D 85.3 \\ 28.6 min · 39.7k}}
  \end{tabular} \\
\addlinespace[1pt]
\texttt{\scriptsize rcb\_000514100} &
  \includegraphics[width=0.95\linewidth,height=1.48cm,keepaspectratio]{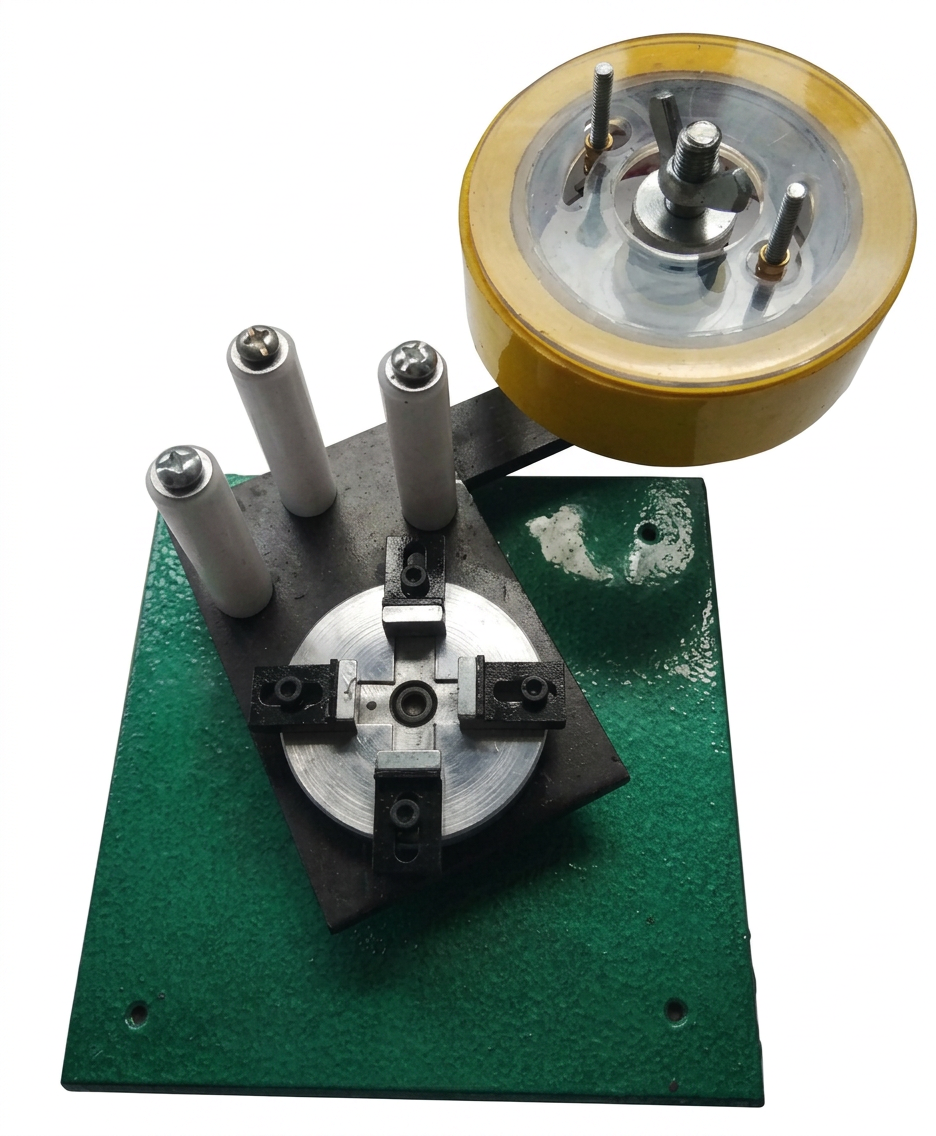} &
  \begin{tabular}[t]{@{}c@{}}
    \includegraphics[width=0.95\linewidth,height=1.48cm,keepaspectratio]{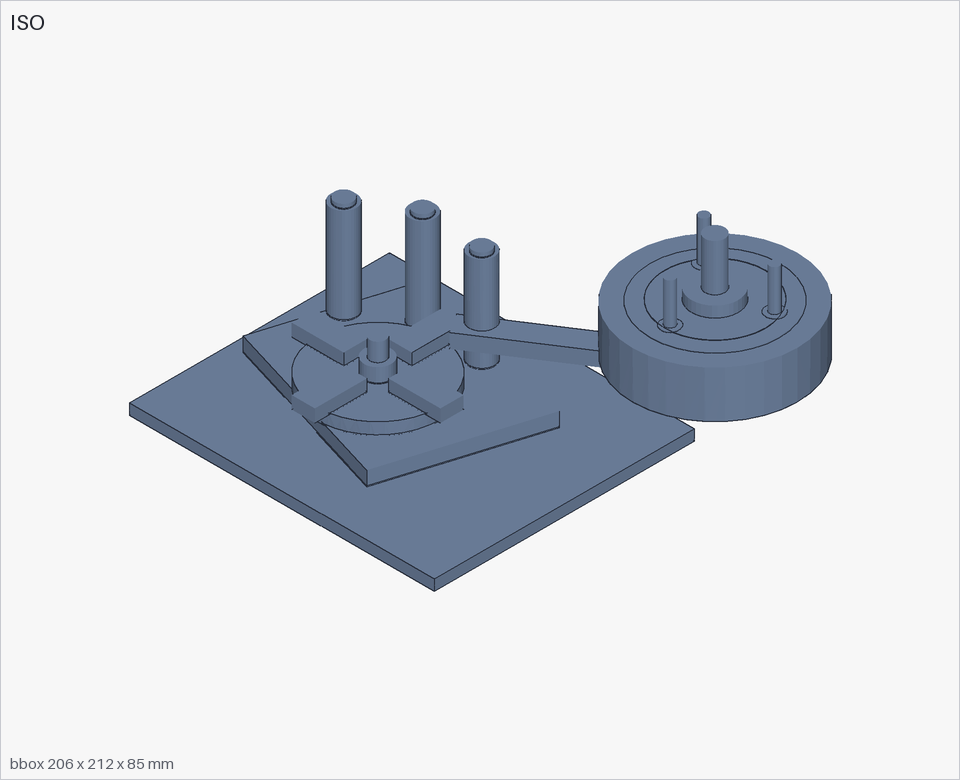} \\
    {\tiny\ttfamily\makecell[c]{Q 73.8 · F 81.5 · D 76.0 \\ 10.2 min · 20.1k}}
  \end{tabular} &
  \begin{tabular}[t]{@{}c@{}}
    \includegraphics[width=0.95\linewidth,height=1.48cm,keepaspectratio]{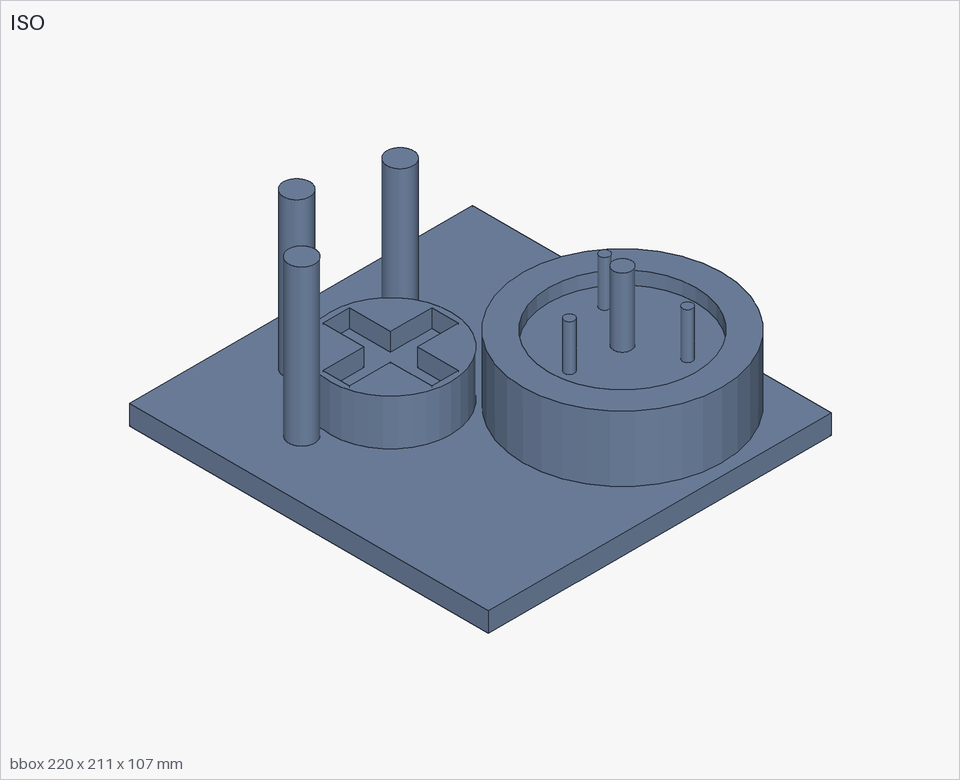} \\
    {\tiny\ttfamily\makecell[c]{Q 52.0 · F 54.4 · D 51.8 \\ --- · ---}}
  \end{tabular} \\
\addlinespace[1pt]
\texttt{\scriptsize rcb\_000529447} &
  \includegraphics[width=0.95\linewidth,height=1.48cm,keepaspectratio]{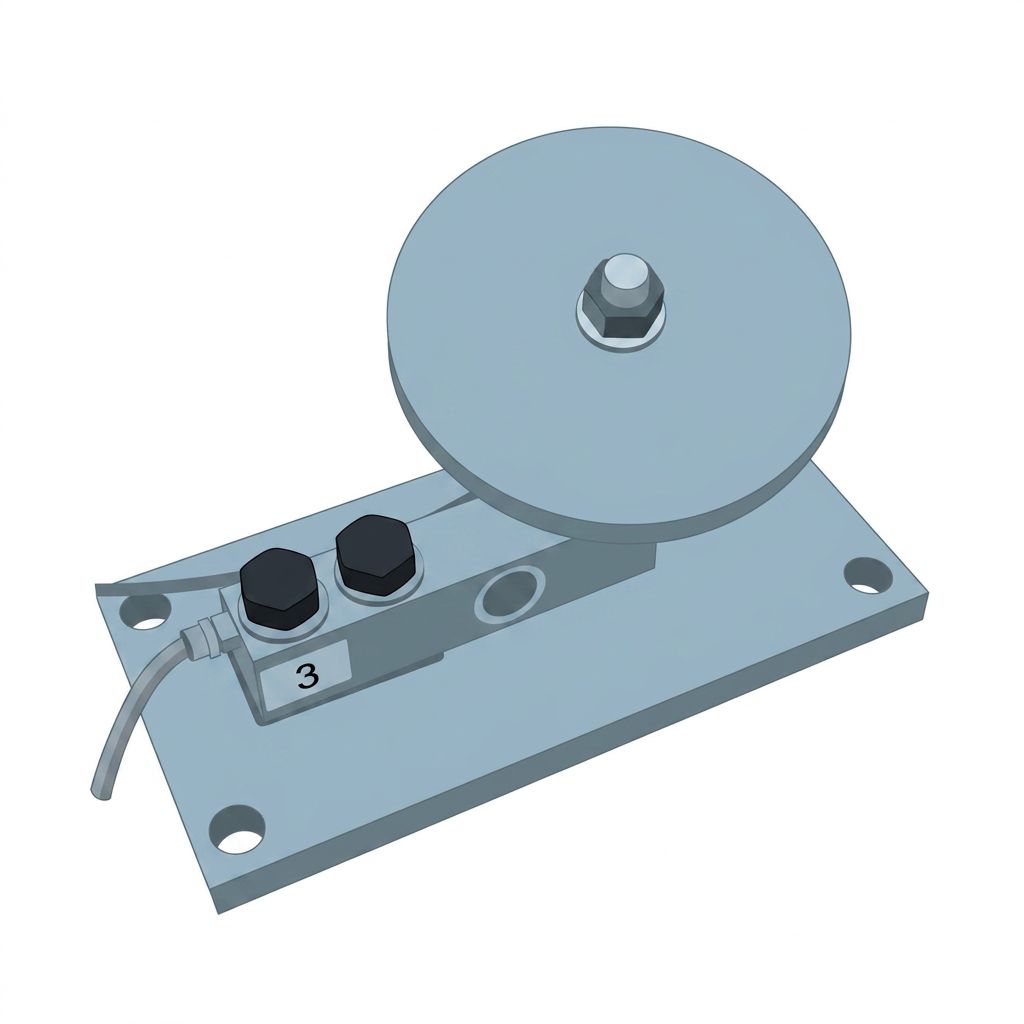} &
  \begin{tabular}[t]{@{}c@{}}
    \includegraphics[width=0.95\linewidth,height=1.48cm,keepaspectratio]{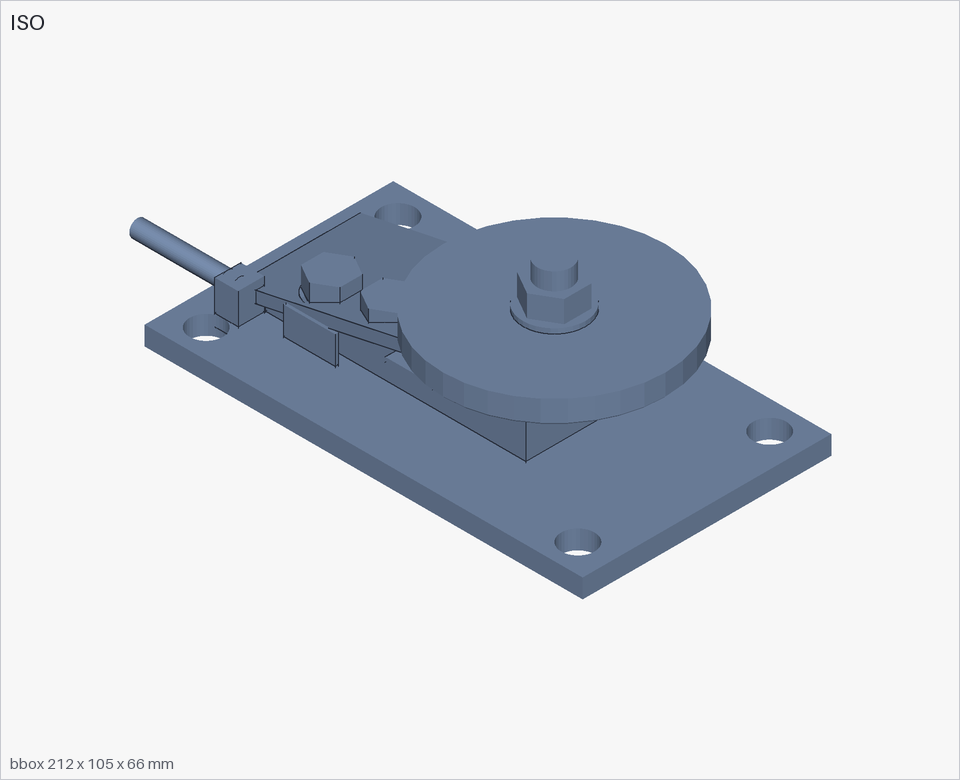} \\
    {\tiny\ttfamily\makecell[c]{Q 79.5 · F 81.7 · D 80.2 \\ 14.4 min · 29.1k}}
  \end{tabular} &
  \begin{tabular}[t]{@{}c@{}}
    \includegraphics[width=0.95\linewidth,height=1.48cm,keepaspectratio]{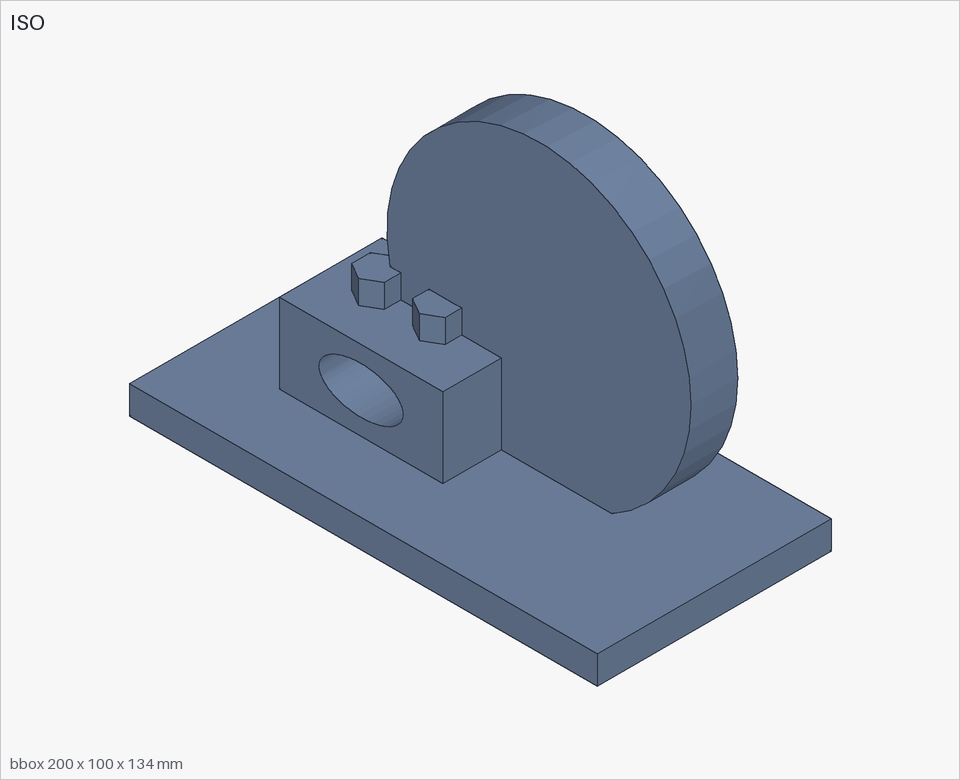} \\
    {\tiny\ttfamily\makecell[c]{Q 57.1 · F 49.7 · D 50.3 \\ 27.4 min · 41.2k}}
  \end{tabular} \\
\addlinespace[1pt]
\texttt{\scriptsize rcb\_000539368} &
  \includegraphics[width=0.95\linewidth,height=1.48cm,keepaspectratio]{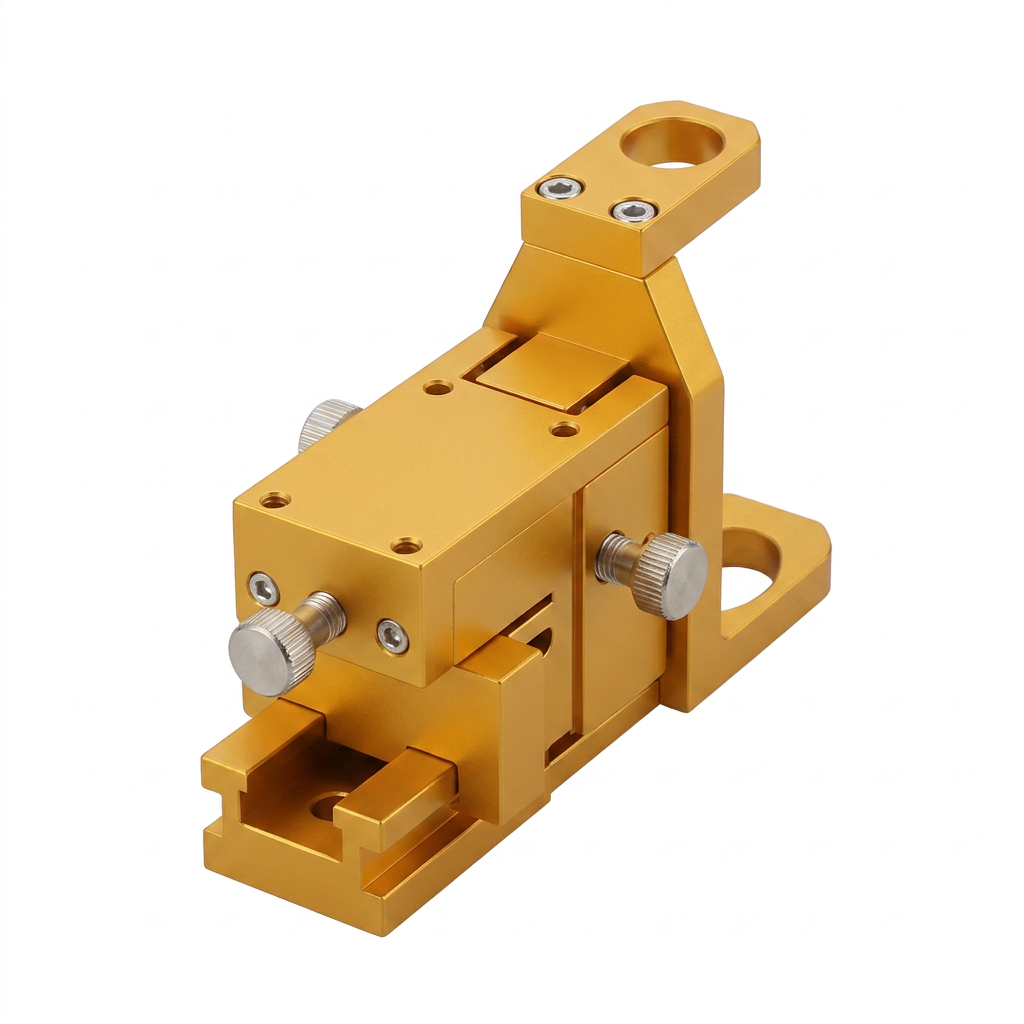} &
  \begin{tabular}[t]{@{}c@{}}
    \includegraphics[width=0.95\linewidth,height=1.48cm,keepaspectratio]{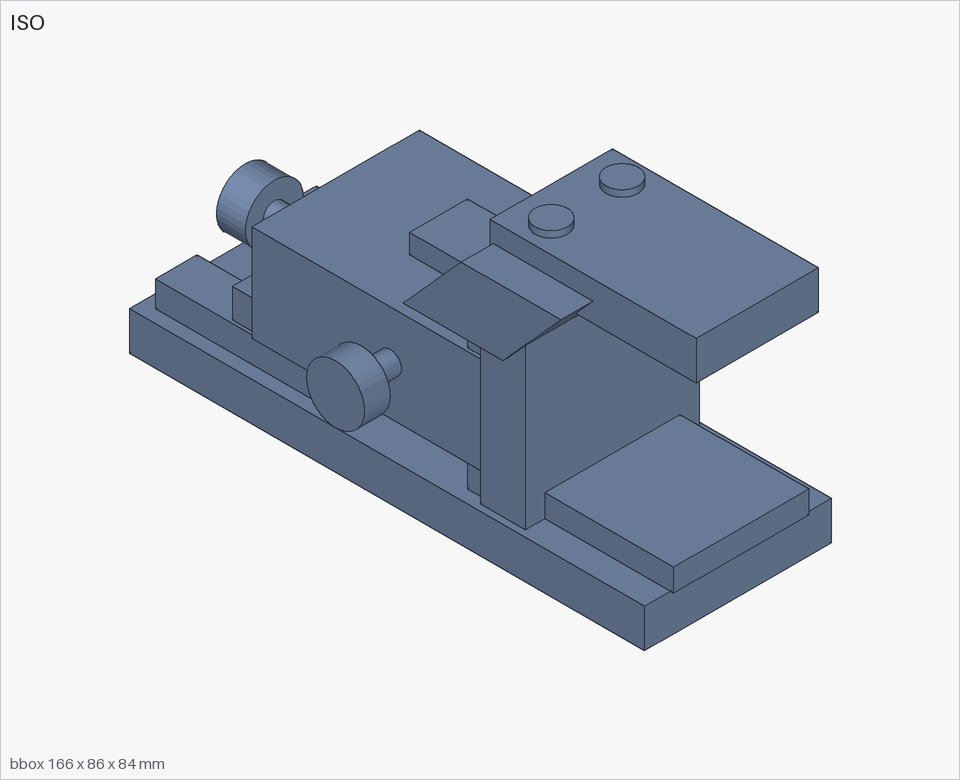} \\
    {\tiny\ttfamily\makecell[c]{Q 54.4 · F 68.2 · D 58.3 \\ 10.5 min · 22.7k}}
  \end{tabular} &
  \begin{tabular}[t]{@{}c@{}}
    \includegraphics[width=0.95\linewidth,height=1.48cm,keepaspectratio]{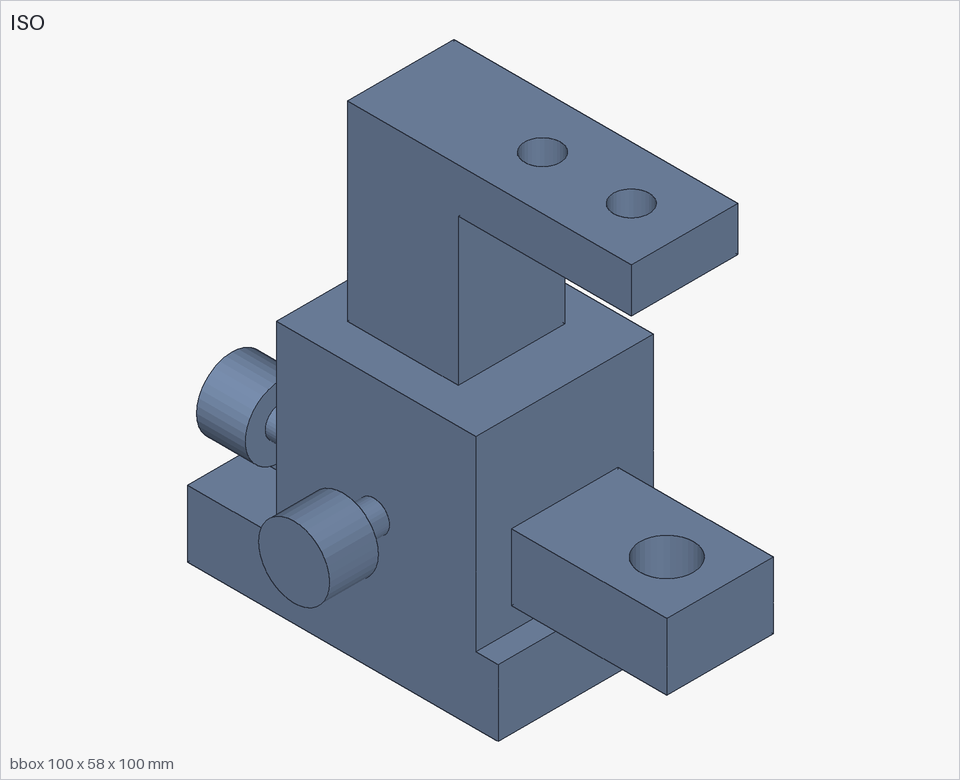} \\
    {\tiny\ttfamily\makecell[c]{Q 51.4 · F 68.9 · D 61.3 \\ 20.6 min · 42.2k}}
  \end{tabular} \\
\addlinespace[1pt]
\texttt{\scriptsize rcb\_000540054} &
  \includegraphics[width=0.95\linewidth,height=1.48cm,keepaspectratio]{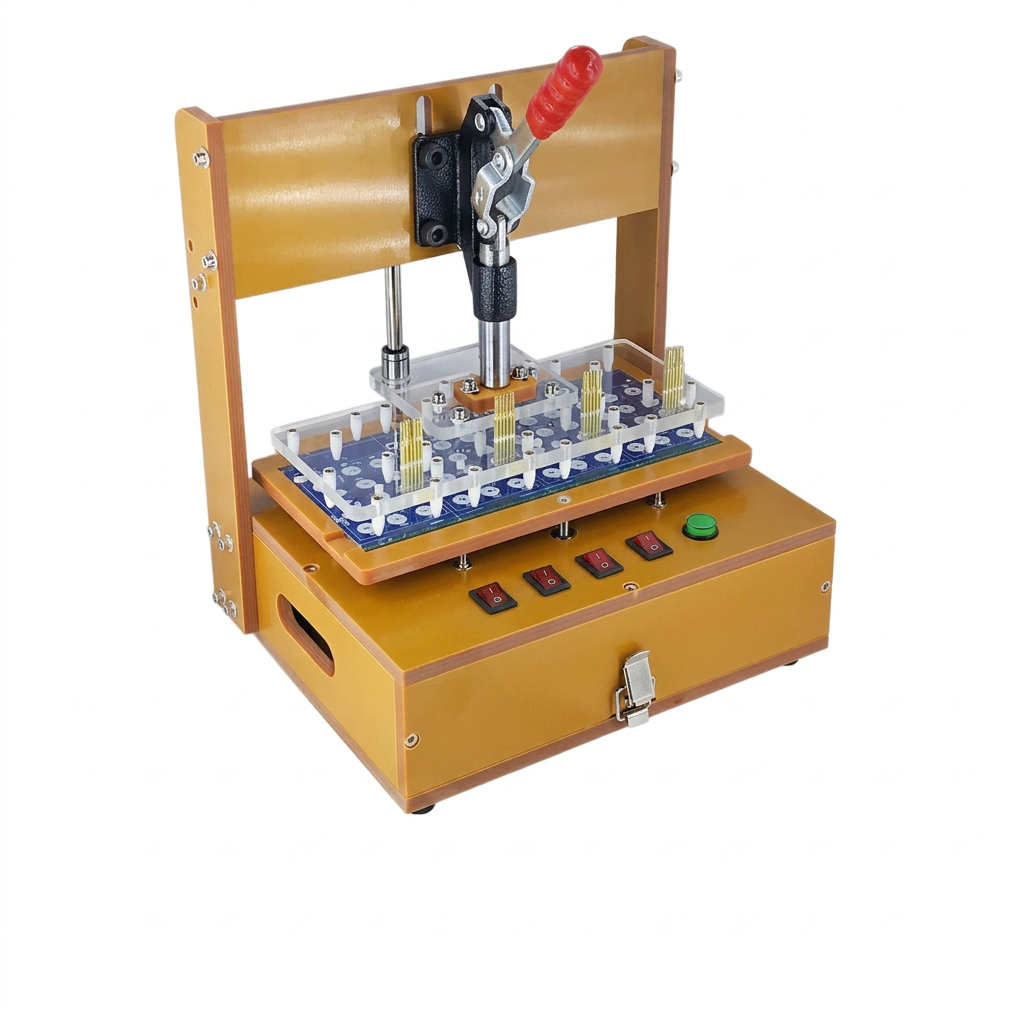} &
  \begin{tabular}[t]{@{}c@{}}
    \includegraphics[width=0.95\linewidth,height=1.48cm,keepaspectratio]{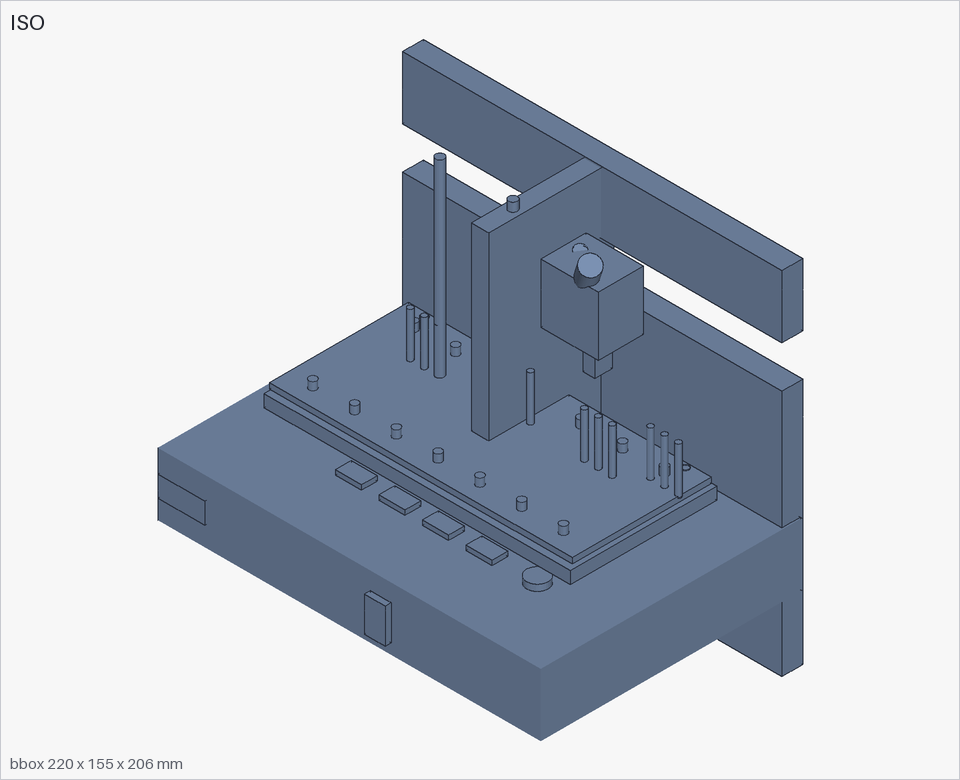} \\
    {\tiny\ttfamily\makecell[c]{Q 76.5 · F 82.2 · D 80.2 \\ 14.3 min · 26.7k}}
  \end{tabular} &
  \begin{tabular}[t]{@{}c@{}}
    \includegraphics[width=0.95\linewidth,height=1.48cm,keepaspectratio]{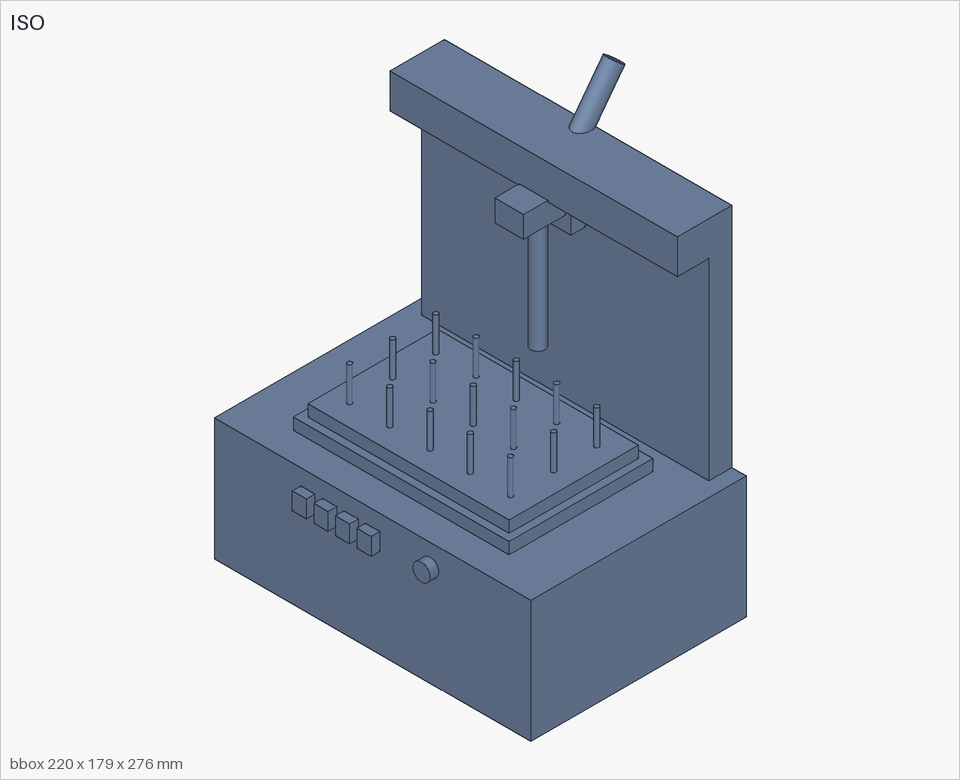} \\
    {\tiny\ttfamily\makecell[c]{Q 67.8 · F 76.9 · D 73.2 \\ 22.8 min · 51.8k}}
  \end{tabular} \\
\addlinespace[1pt]
\texttt{\scriptsize rcb\_000575344} &
  \includegraphics[width=0.95\linewidth,height=1.48cm,keepaspectratio]{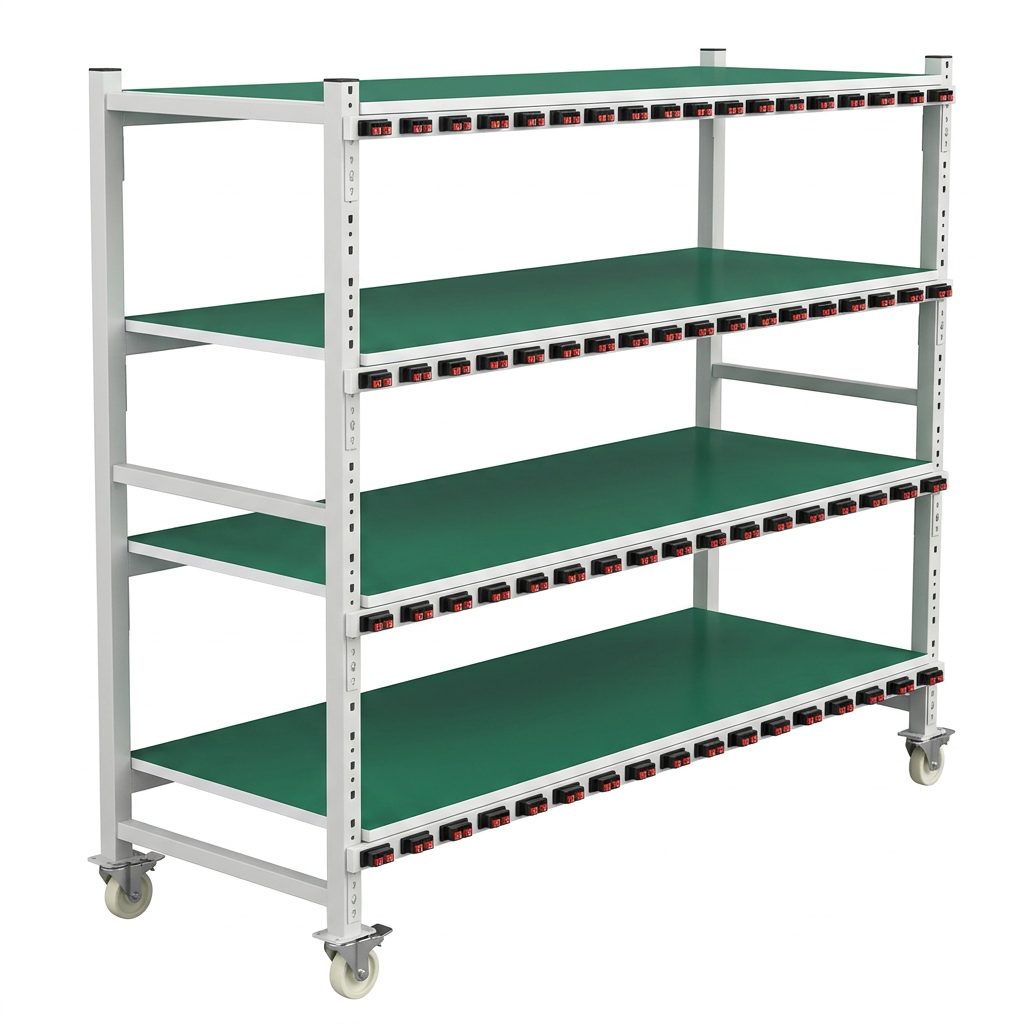} &
  \begin{tabular}[t]{@{}c@{}}
    \includegraphics[width=0.95\linewidth,height=1.48cm,keepaspectratio]{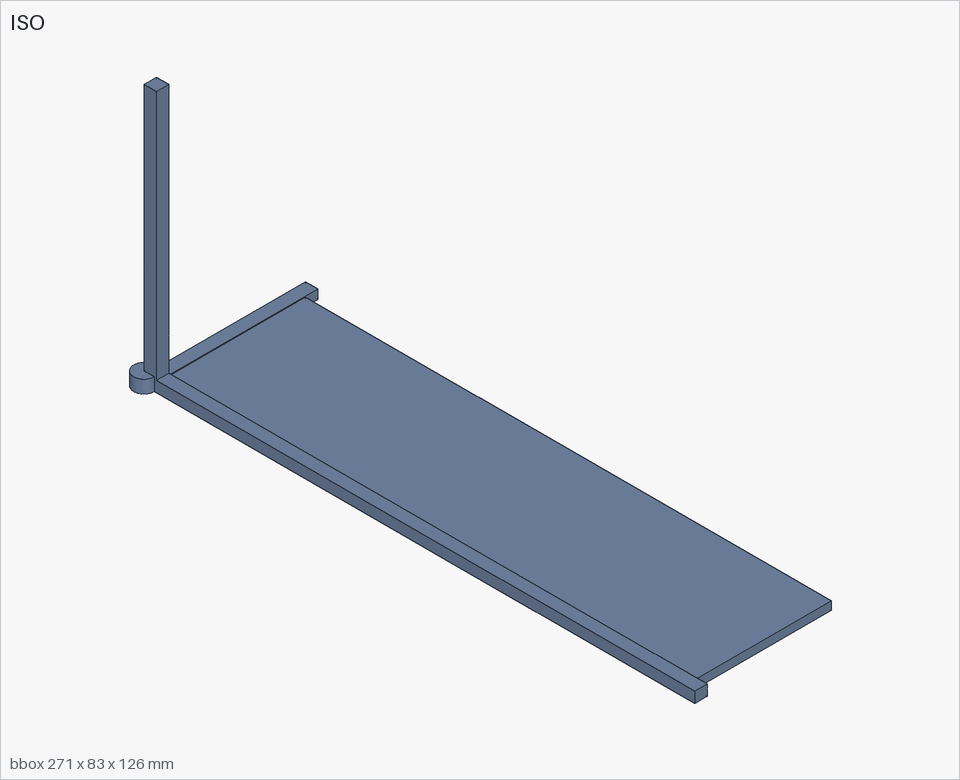} \\
    {\tiny\ttfamily\makecell[c]{Q 22.0 · F 23.2 · D 12.5 \\ 12.3 min · 27.8k}}
  \end{tabular} &
  \begin{tabular}[t]{@{}c@{}}
    \includegraphics[width=0.95\linewidth,height=1.48cm,keepaspectratio]{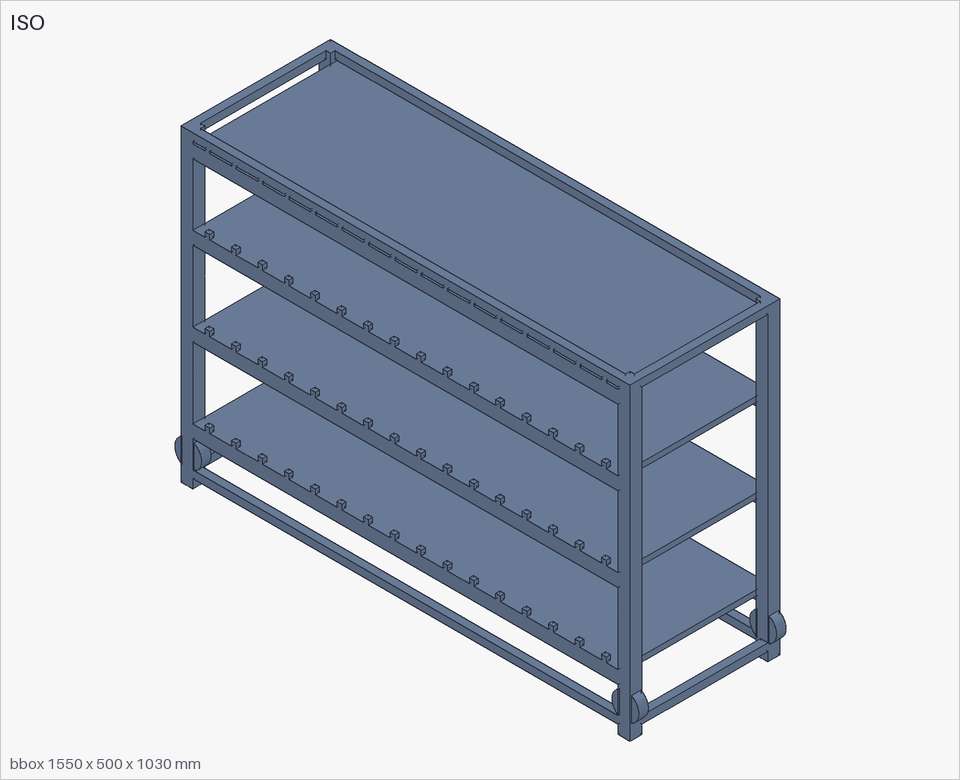} \\
    {\tiny\ttfamily\makecell[c]{Q 68.7 · F 82.8 · D 74.5 \\ 41.7 min · 56.2k}}
  \end{tabular} \\
\addlinespace[1pt]
\texttt{\scriptsize rcb\_000580852} &
  \includegraphics[width=0.95\linewidth,height=1.48cm,keepaspectratio]{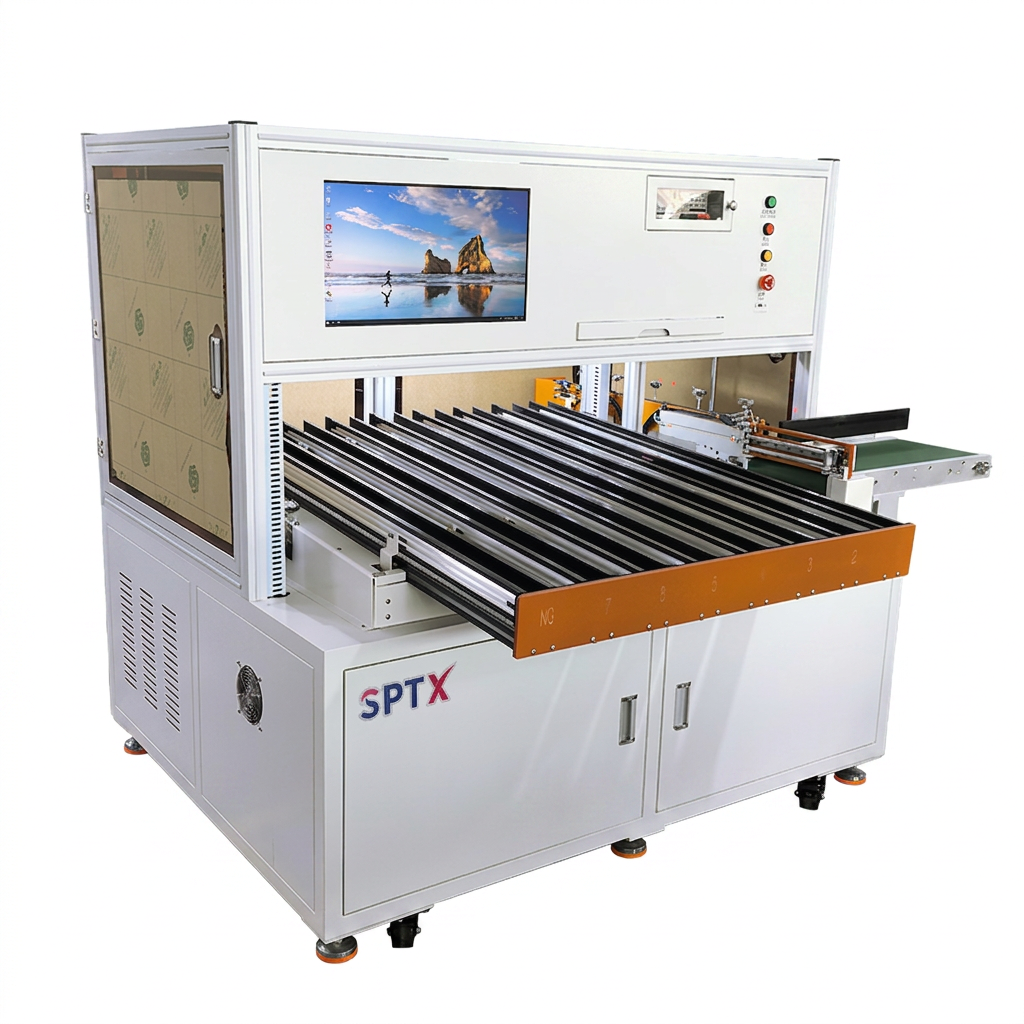} &
  \begin{tabular}[t]{@{}c@{}}
    \includegraphics[width=0.95\linewidth,height=1.48cm,keepaspectratio]{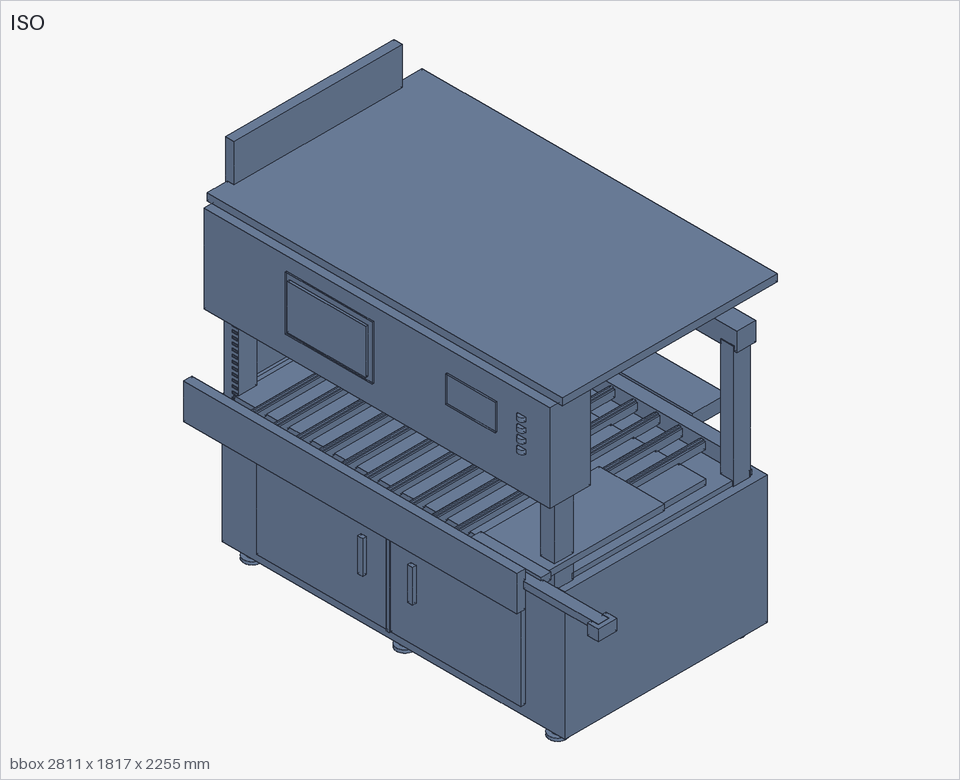} \\
    {\tiny\ttfamily\makecell[c]{Q 70.5 · F 78.0 · D 75.0 \\ 9.2 min · 22.2k}}
  \end{tabular} &
  \begin{tabular}[t]{@{}c@{}}
    \includegraphics[width=0.95\linewidth,height=1.48cm,keepaspectratio]{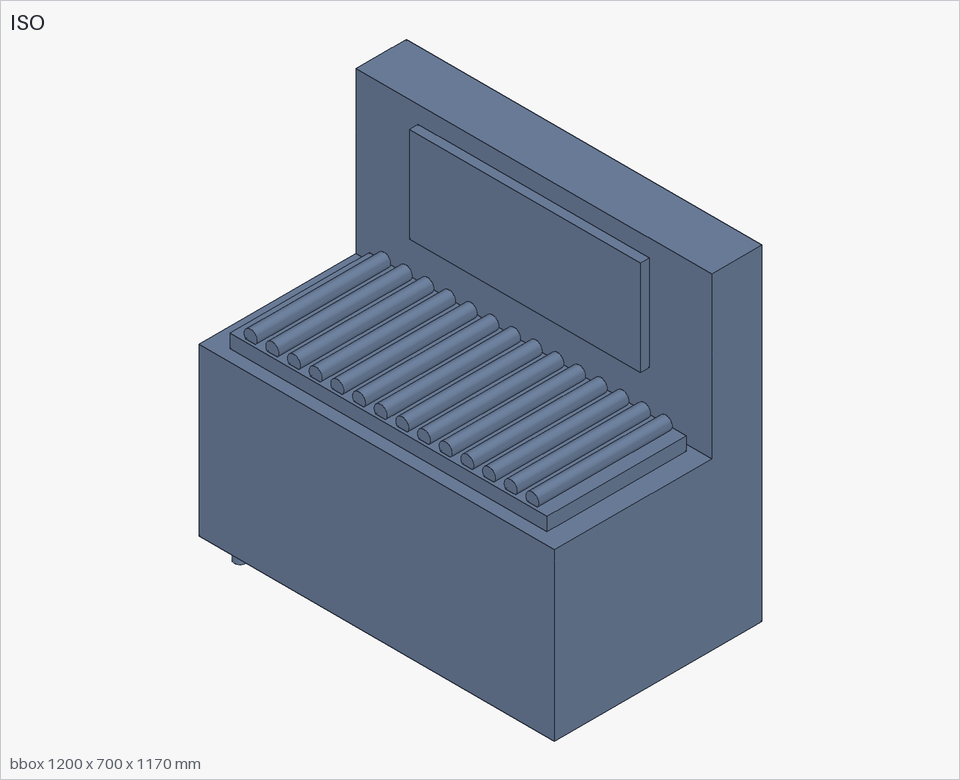} \\
    {\tiny\ttfamily\makecell[c]{Q 37.4 · F 50.3 · D 36.7 \\ 32.0 min · 72.3k}}
  \end{tabular} \\
\addlinespace[1pt]
\texttt{\scriptsize rcb\_000613854} &
  \includegraphics[width=0.95\linewidth,height=1.48cm,keepaspectratio]{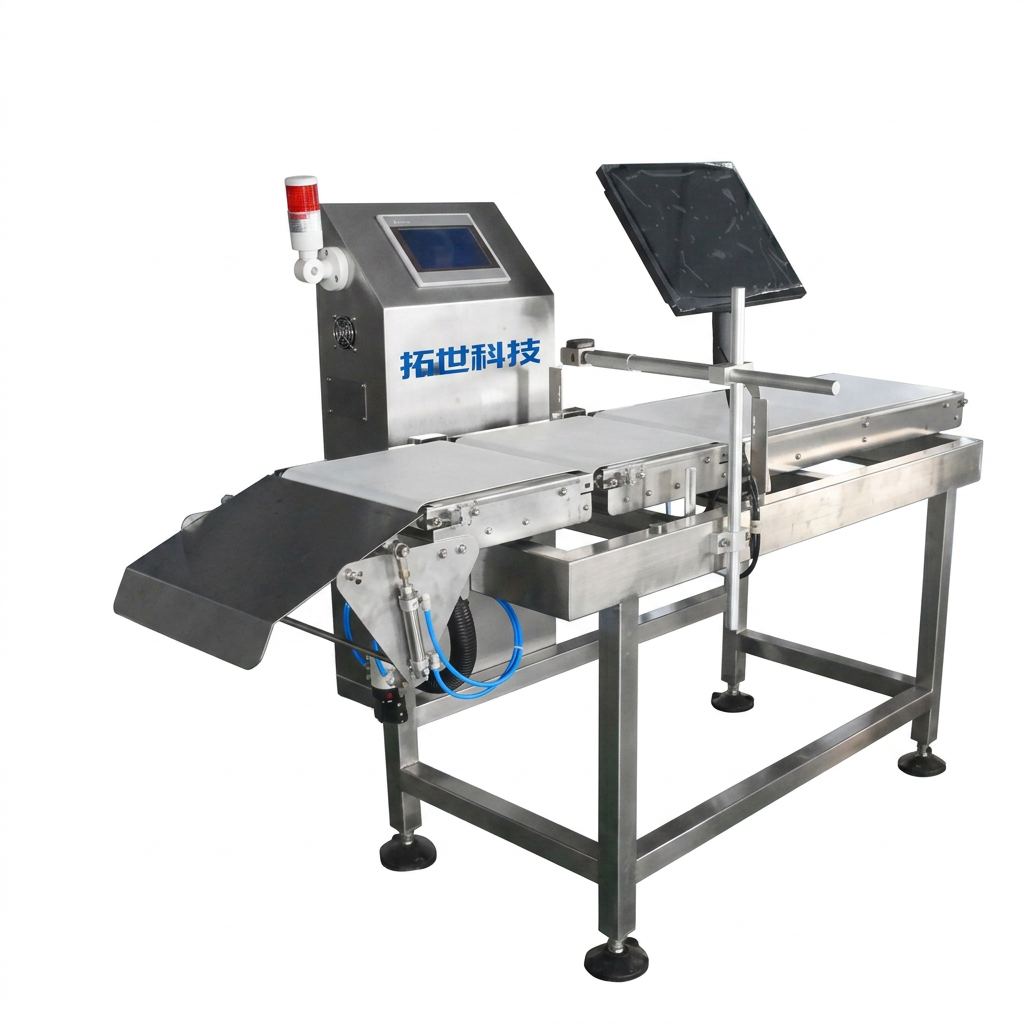} &
  \begin{tabular}[t]{@{}c@{}}
    \includegraphics[width=0.95\linewidth,height=1.48cm,keepaspectratio]{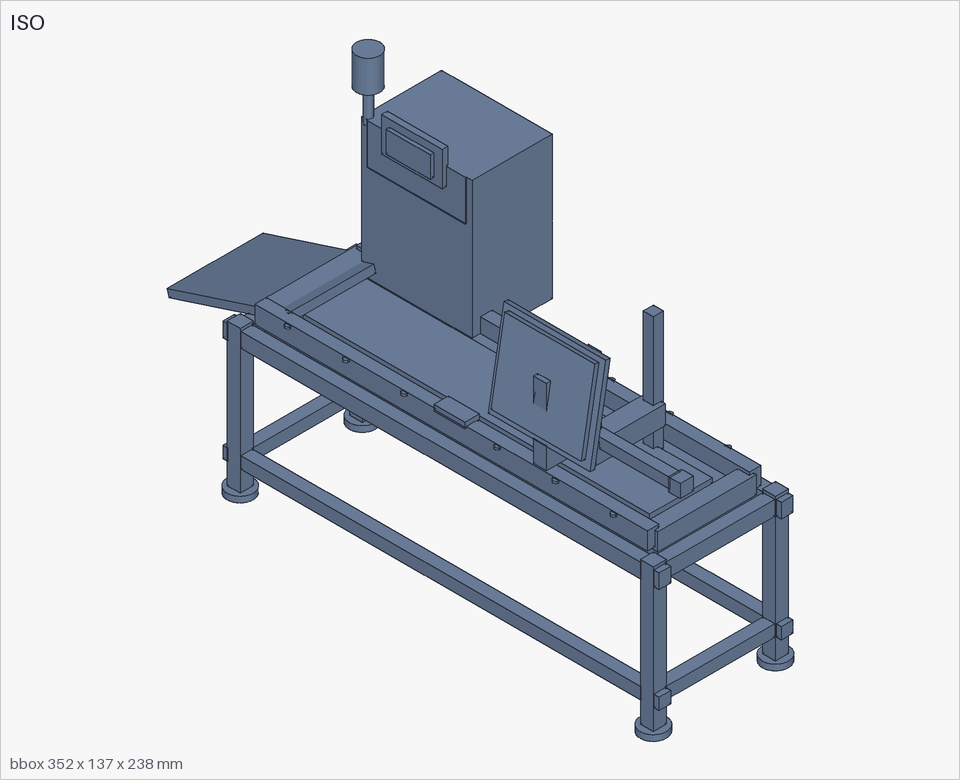} \\
    {\tiny\ttfamily\makecell[c]{Q 72.5 · F 78.8 · D 74.5 \\ 13.8 min · 29.0k}}
  \end{tabular} &
  \begin{tabular}[t]{@{}c@{}}
    \includegraphics[width=0.95\linewidth,height=1.48cm,keepaspectratio]{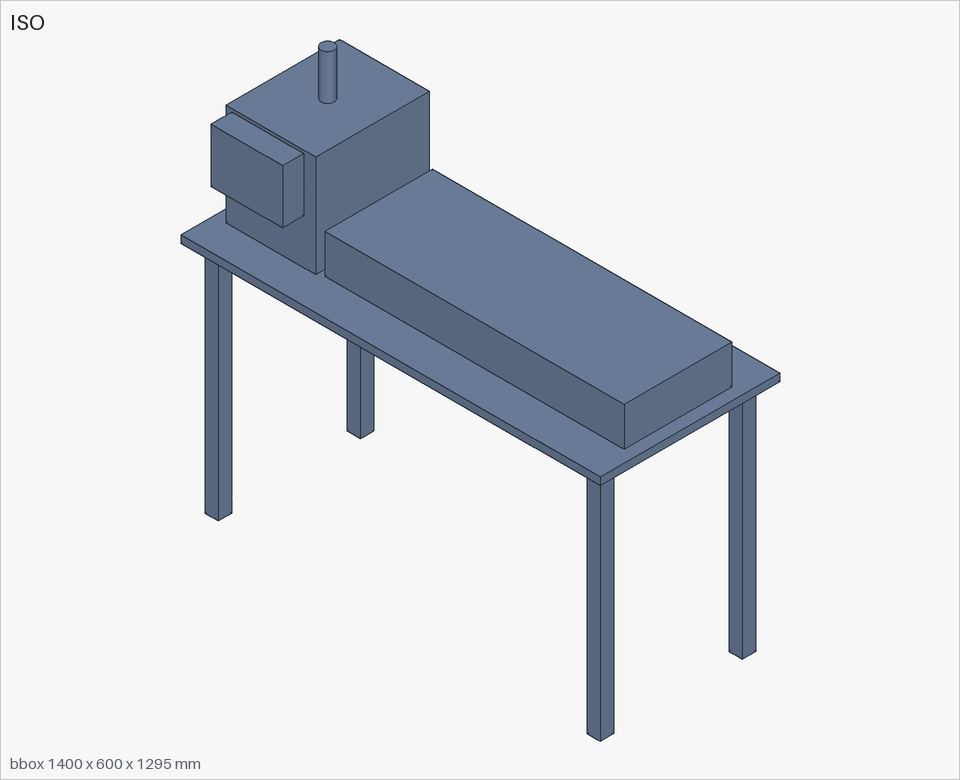} \\
    {\tiny\ttfamily\makecell[c]{Q 31.6 · F 46.0 · D 37.2 \\ 15.7 min · 26.8k}}
  \end{tabular} \\
\addlinespace[1pt]
\texttt{\scriptsize rcb\_000627090} &
  \includegraphics[width=0.95\linewidth,height=1.48cm,keepaspectratio]{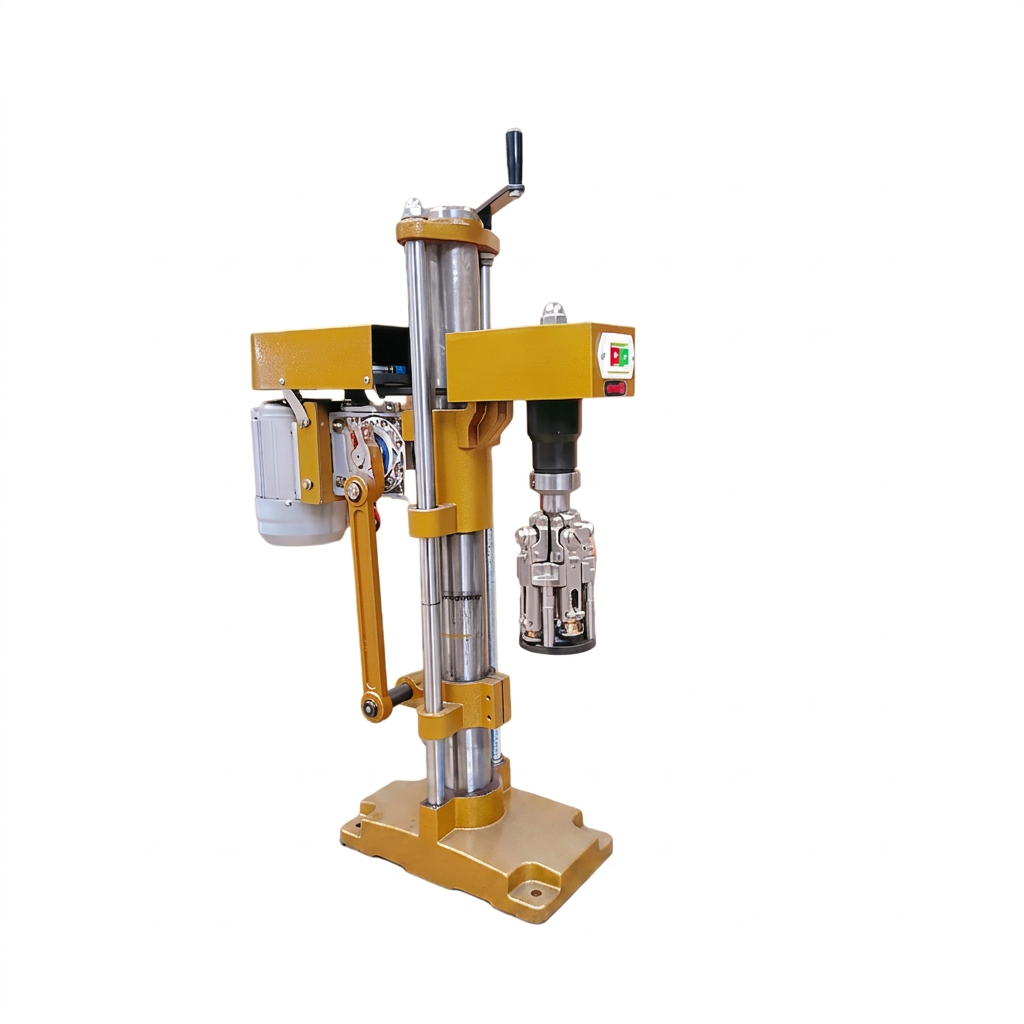} &
  \begin{tabular}[t]{@{}c@{}}
    \includegraphics[width=0.95\linewidth,height=1.48cm,keepaspectratio]{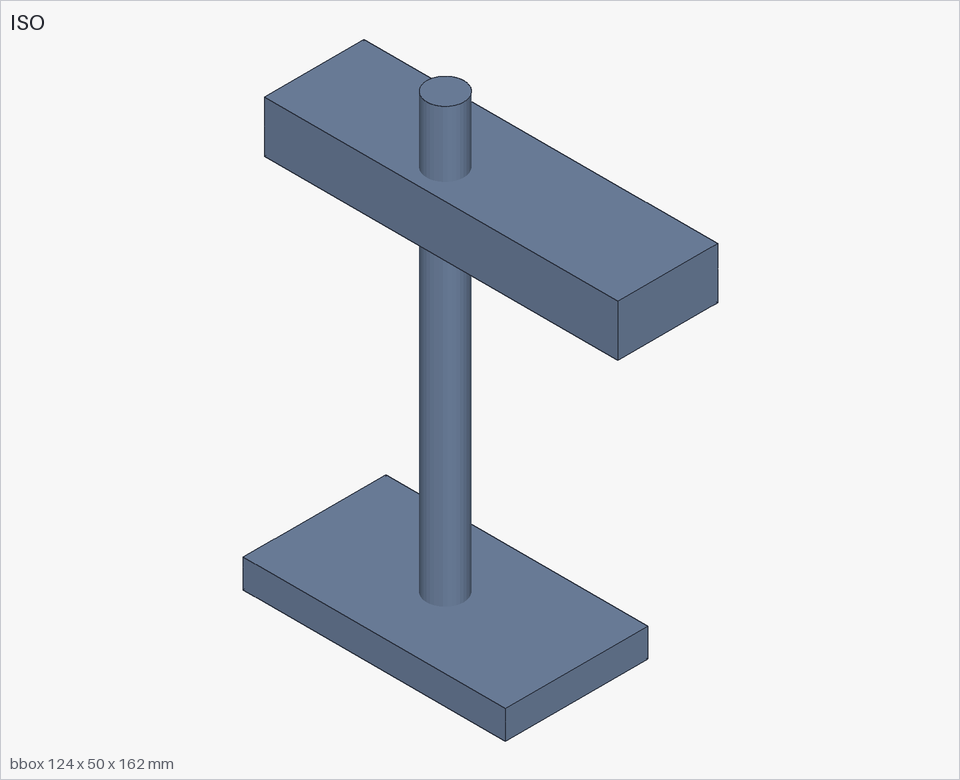} \\
    {\tiny\ttfamily\makecell[c]{Q 20.2 · F 39.8 · D 22.8 \\ 13.7 min · 29.9k}}
  \end{tabular} &
  \begin{tabular}[t]{@{}c@{}}
    \includegraphics[width=0.95\linewidth,height=1.48cm,keepaspectratio]{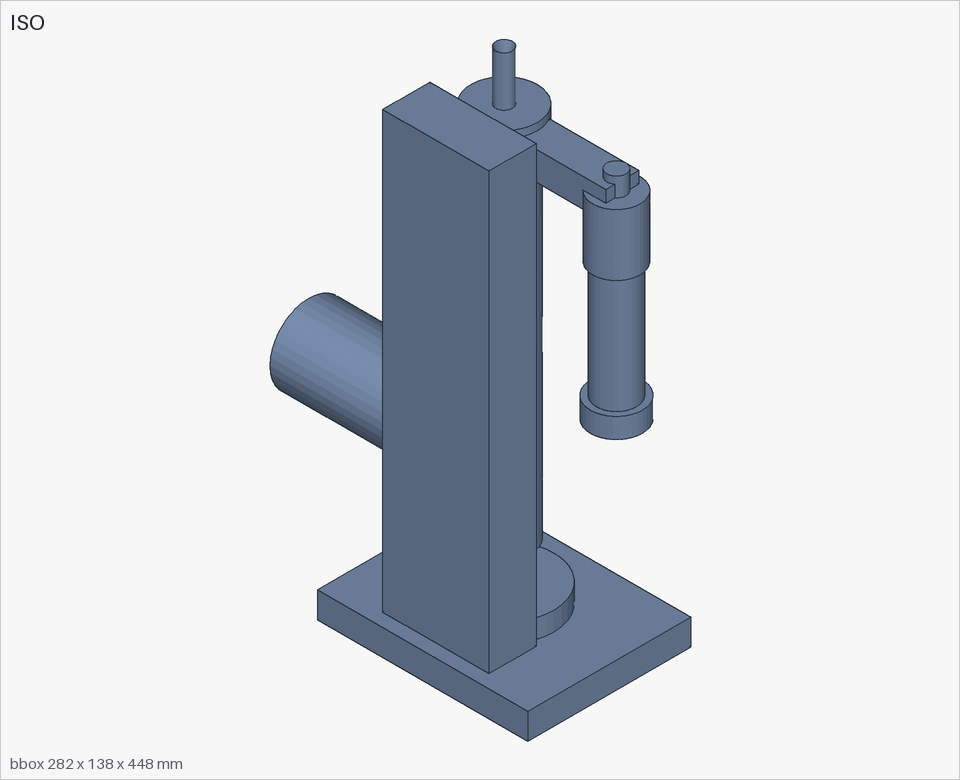} \\
    {\tiny\ttfamily\makecell[c]{Q 31.6 · F 56.3 · D 48.5 \\ 36.4 min · 48.7k}}
  \end{tabular} \\
\addlinespace[1pt]
\texttt{\scriptsize rcb\_000631731} &
  \includegraphics[width=0.95\linewidth,height=1.48cm,keepaspectratio]{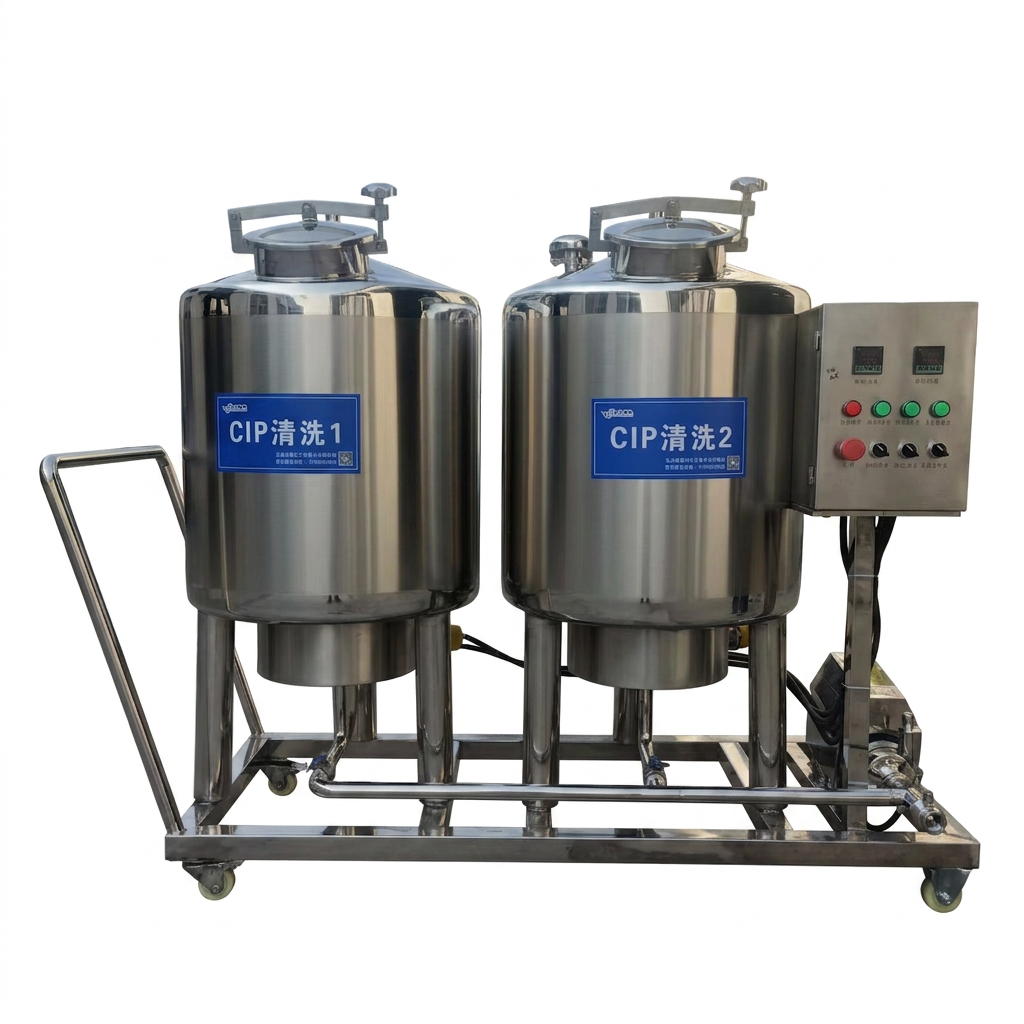} &
  \begin{tabular}[t]{@{}c@{}}
    \includegraphics[width=0.95\linewidth,height=1.48cm,keepaspectratio]{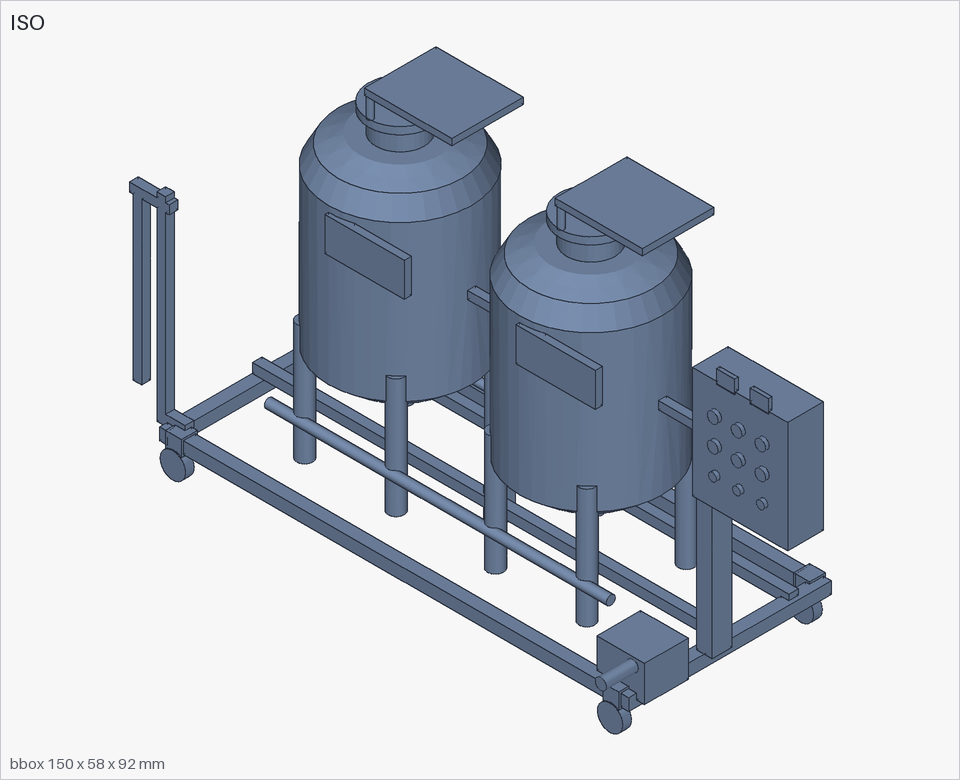} \\
    {\tiny\ttfamily\makecell[c]{Q 68.2 · F 76.2 · D 72.5 \\ 14.7 min · 30.4k}}
  \end{tabular} &
  \begin{tabular}[t]{@{}c@{}}
    \includegraphics[width=0.95\linewidth,height=1.48cm,keepaspectratio]{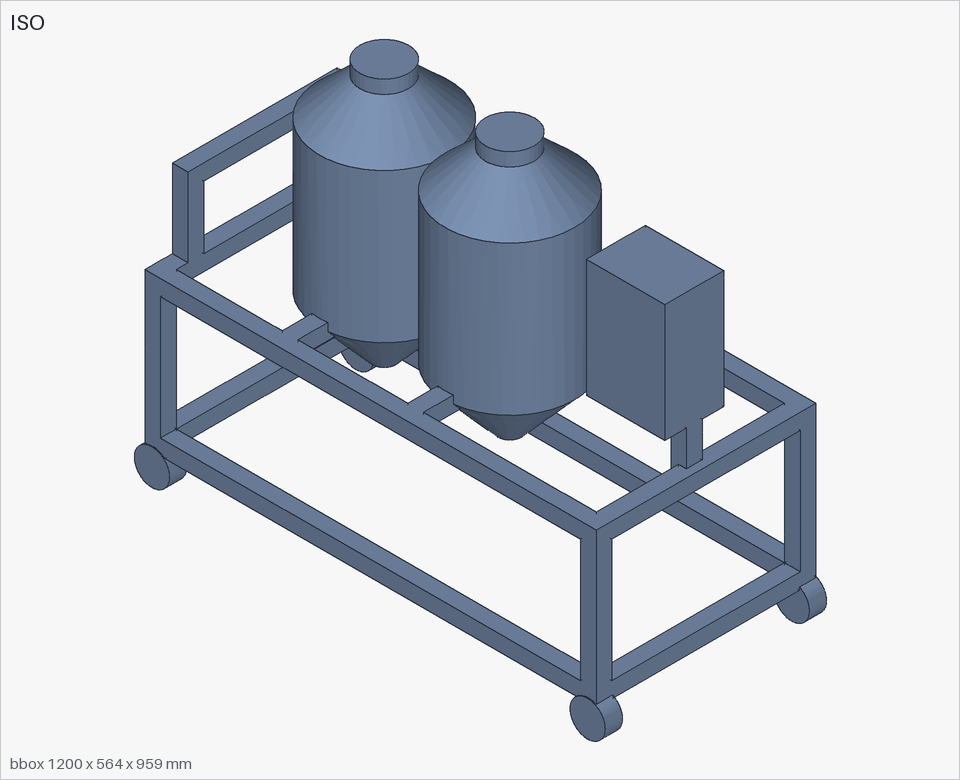} \\
    {\tiny\ttfamily\makecell[c]{Q 43.5 · F 61.5 · D 46.0 \\ 22.6 min · 63.5k}}
  \end{tabular} \\
\addlinespace[1pt]
\texttt{\scriptsize rcb\_000633471} &
  \includegraphics[width=0.95\linewidth,height=1.48cm,keepaspectratio]{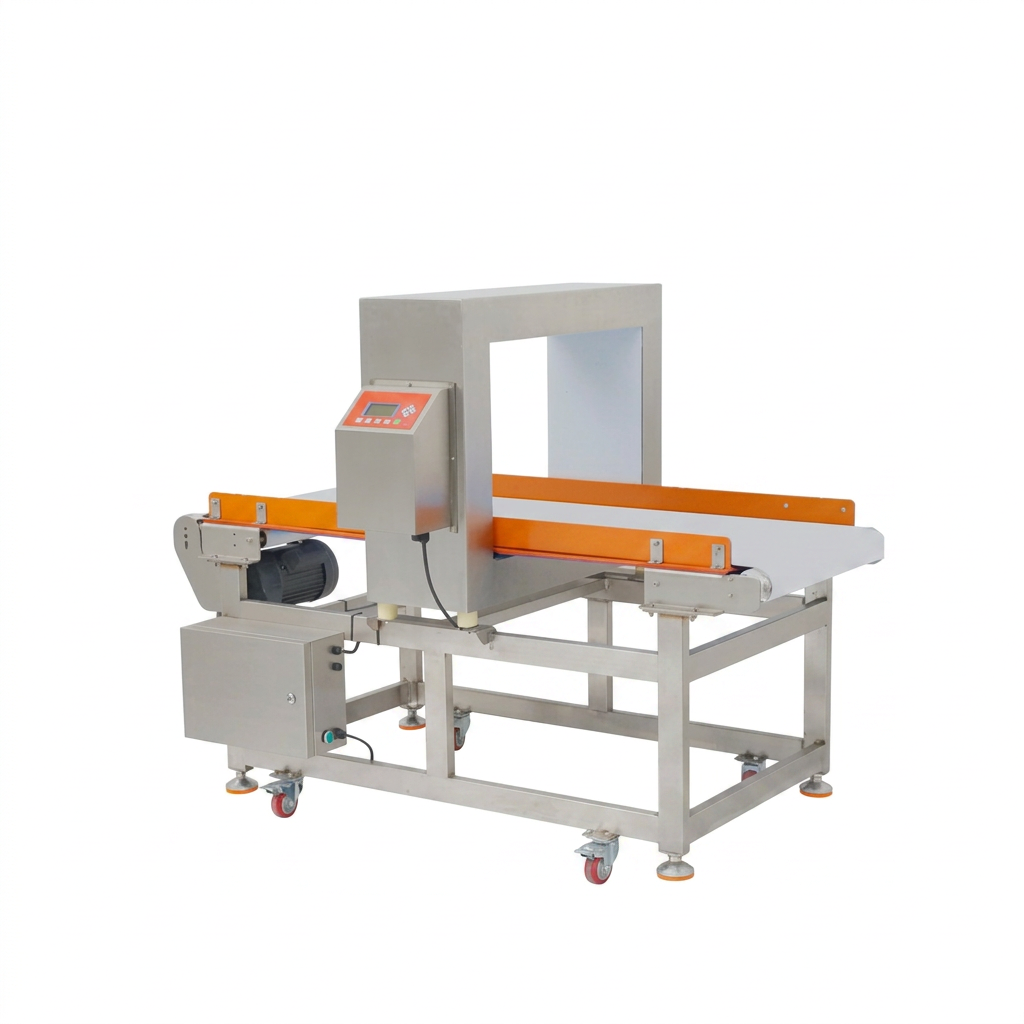} &
  \begin{tabular}[t]{@{}c@{}}
    \includegraphics[width=0.95\linewidth,height=1.48cm,keepaspectratio]{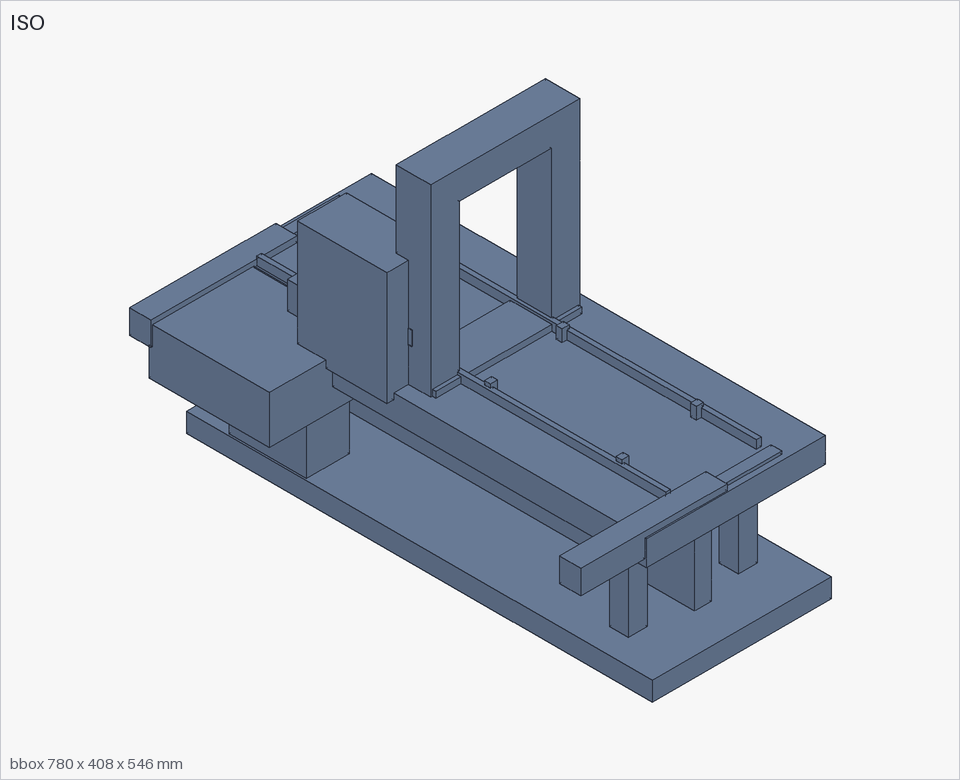} \\
    {\tiny\ttfamily\makecell[c]{Q 65.0 · F 77.2 · D 70.0 \\ 14.8 min · 30.2k}}
  \end{tabular} &
  \begin{tabular}[t]{@{}c@{}}
    {\scriptsize\makecell{no exported\\artifact}} \\
    {\tiny\ttfamily\makecell[c]{no exported artifact \\ 18.6 min · 21.8k}}
  \end{tabular} \\

\end{longtable}
\renewcommand{\arraystretch}{1}

\end{document}